\documentclass{article}

\usepackage[preprint,nonatbib]{neurips_2023}

\usepackage[utf8]{inputenc} % allow utf-8 input
\usepackage[T1]{fontenc}    % use 8-bit T1 fonts
\usepackage{hyperref}       % hyperlinks
\usepackage{url}            % simple URL typesetting
\usepackage{booktabs}       % professional-quality tables
\usepackage{amsfonts}       % blackboard math symbols
\usepackage{nicefrac}       % compact symbols for 1/2, etc.
\usepackage{microtype}      % microtypography
\usepackage[dvipsnames]{xcolor}         % colors

\usepackage{adjustbox}
\usepackage{amsmath}
\usepackage{amssymb}
\usepackage[natbib]{biblatex}
\usepackage{graphicx}
\usepackage[font=small]{caption}
\usepackage{multirow}
\usepackage{forloop}
\usepackage{pgfplots}
\usepackage{rotating}
\usepackage{subcaption}
\usepackage{tikz}
\usepackage{tcolorbox}
\usepackage{fancyvrb}

\pgfplotsset{compat=1.18}
\usepgfplotslibrary{fillbetween}
\usetikzlibrary{shapes.misc, positioning, shapes, shadows, patterns}
\tcbuselibrary{skins}

\title{Large Language Models Are Not Strong Abstract Reasoners}

\author{%
  Ga\"{e}l Gendron \\
  University of Auckland \\
  \texttt{ggen187@aucklanduni.ac.nz} \\
  \And
  Qiming Bao \\
  University of Auckland \\
  \texttt{qbao775@aucklanduni.ac.nz} \\
  \And
  Michael Witbrock \\
  University of Auckland\\
  \texttt{m.witbrock@auckland.ac.nz} \\
  \And
  Gillian Dobbie \\
  University of Auckland \\
  \texttt{g.dobbie@auckland.ac.nz} \\
}

\begin{document}

\maketitle

\begin{abstract}

Large Language Models have shown tremendous performance on a large variety of natural language processing tasks, ranging from text comprehension to common sense reasoning. 
However, the mechanisms responsible for this success remain opaque, and it is unclear whether LLMs can achieve human-like cognitive capabilities or whether these models are still fundamentally circumscribed.
Abstract reasoning is a fundamental task for cognition, consisting of finding and applying a general pattern from few data. Evaluating deep neural architectures on this task could give insight into their potential limitations regarding reasoning and their broad generalisation abilities, yet this is currently an under-explored area.
In this paper, we introduce a new benchmark for evaluating language models beyond memorization on abstract reasoning tasks. We perform extensive evaluations of state-of-the-art LLMs, showing that they currently achieve very limited performance in contrast with other natural language tasks, even when applying techniques that have been shown to improve performance on other NLP tasks.
We argue that guiding LLM generation to follow causal paths could help improve the generalisation and reasoning abilities of LLMs.

\end{abstract}

\section{Introduction}

Large Language Models (LLMs) have recently achieved impressive performance on a large variety of Natural Language Processing (NLP) tasks, including text comprehension \citep{DBLP:conf/naacl/DevlinCLT19, radford2019language}, commonsense reasoning \citep{DBLP:conf/nips/TalmorTCGB20}, translation \citep{DBLP:journals/jmlr/RaffelSRLNMZLL20}, and code generation \citep{DBLP:journals/corr/abs-2107-03374, DBLP:journals/corr/abs-2303-12712}, and have shown promising results for out-of-distribution generalisation \citep{DBLP:conf/nips/BrownMRSKDNSSAA20, DBLP:journals/corr/abs-2303-12712}.
The most recent and larger language models also perform well on mathematical problems, which had been out of reach for transformers for a long time \citep{DBLP:journals/corr/abs-2211-12588, DBLP:journals/corr/abs-2210-12023}. 
While empirical testing of LLMs trained on large corpora of data yields signs of high comprehension of presented problems, there is little theoretical evidence regarding why and how this performance has been achieved and whether these models are simply memorising the training data, extrapolating it, or some combination \citep{DBLP:conf/nips/TirumalaMZA22, DBLP:journals/corr/abs-2011-15091}. A notable limitation of these models is a lack of control mechanisms, or possible misalignment \citep{DBLP:conf/nips/Ouyang0JAWMZASR22}, for which the absence of a world model or causal representation have been advanced as explanations \citep{DBLP:conf/fat/BenderGMS21, DBLP:journals/corr/abs-2206-10591}. More recently, early experiments on GPT-4 showed signs of limitations on reasoning tasks requiring planning and backtracking \citep{DBLP:journals/corr/abs-2303-12712}. Despite these early limitations, the question of whether or not LLMs can perform human-like reasoning remains open, as measuring the intelligence, or more broadly, the competence, of a system is a challenging task \citep{DBLP:journals/corr/abs-1911-01547}.

Abstract reasoning is a potential task for effective measurement of the cognitive abilities of neural models \citep{DBLP:conf/icml/SantoroHBML18, DBLP:journals/corr/abs-1911-01547}. Abstract reasoning problems consist of identifying generic structures over a small set of examples and applying them to unseen cases. They aim to evaluate the ability of a system to integrate a new skill or process from limited data. The abstract nature of these problems helps avoid spurious correlations that could lie in the data and may create potential bias in the results. In particular, this task is well-suited for evaluating the broad or strong generalisation capacity of a system, i.e. its ability to handle a large category of tasks and environments without human intervention, including situations that may not have been foreseen when the system was created \citep{DBLP:journals/corr/abs-1911-01547}. This is a well-studied class of task in the field of program induction \citep{DBLP:journals/corr/abs-2006-08381, lake2015human}. However, the problem of abstract reasoning has long remained outside the scope of evaluation of language models, and there currently exist no extensive evaluations of the performance of LLMs in this domain.

In this paper, we seek to bridge this gap by investigating the abstract reasoning abilities of LLMs and by providing insight into the following question: Do LLMs contain sufficient building blocks for broad generalisation, or do they lack fundamental capabities?
We evaluate state-of-the-art LLMs on abstract reasoning tasks, applying recent fine-tuning and prompt design techniques that have been shown to improve performance on other NLP tasks. To this end, we create a benchmark based on existing datasets and novel datasets transposed from vision tasks and adapted to text-based models. We then perform extensive experiments on this benchmark. We also build and train a language model for abstract reasoning and compare its performance with the other models. Our results indicate that Large Language Models do not yet have the ability to perform sound abstract reasoning. All of the tested models exhibit poor performance, and the tuning techniques that improved LLM reasoning abilities do not provide significant help for abstract reasoning. 
We release our code and data at: \url{https://github.com/Strong-AI-Lab/Logical-and-abstract-reasoning}.
Our contributions can be summarised as follows:
\begin{itemize}
    \item We evaluate Large Language Models on abstract reasoning tasks.
    \item We show that existing training and tuning techniques do not help increase the performance of LLMs in abstract reasoning, and investigate the reasons and leads for improvement.
    \item We create a benchmark for the evaluation of language models for abstract reasoning.
\end{itemize}

\section{Related Work}

The abilities of Language Models have been thoroughly studied on a wide range of problems. In particular, their reasoning capacities are the focus of a great deal of recent work. Some of this \citep{DBLP:conf/nips/Wei0SBIXCLZ22, DBLP:journals/corr/abs-2206-02336, DBLP:journals/corr/abs-2211-12588} has explored prompt techniques to improve mathematical reasoning in LLMs; \citet{DBLP:journals/corr/abs-2210-12023} propose a framework based on causality theory to evaluate language models on this kind of task. Recently, GPT-4 has been shown to perform well on mathematical problems, outperforming PaLM and LLaMA \citep{DBLP:journals/corr/abs-2204-02311, DBLP:journals/corr/abs-2302-13971}, although it still produces calculation mistakes \citep{DBLP:journals/corr/abs-2303-12712}. 
In the domain of logical reasoning, several methods and benchmarks exists for evaluating language models. Notable benchmarks include DEER \citep{DBLP:journals/corr/abs-2212-10923}, ParaRules \citep{DBLP:conf/ijcai/ClarkTR20}, PARARULE-Plus \citep{DBLP:conf/nesy/BaoPHTDWL22}, ReClor \citep{DBLP:conf/iclr/YuJDF20}, LogiQA \citep{DBLP:conf/ijcai/LiuCLHWZ20}, and AbductionRules \citep{DBLP:conf/acl/Young0BW22}. Models such as LReasoner \citep{DBLP:conf/acl/WangZTWFJ0D22}, MERIt \citep{DBLP:conf/acl/JiaoGSN22}, and AMR-LE \citep{DBLP:journals/corr/abs-2305-12599} attempt to induce logical reasoning abilities in language models, but the performance of the most recent LLMs is yet to be evaluated.
Similarly, the CLRS dataset benchmark for evaluating algorithmic reasoning has not yet been applied to language models \citep{DBLP:conf/icml/VelickovicBBPBD22}.
Causal structure discovery and causal inference are other domains where LLMs have shown mixed results \citep{DBLP:journals/corr/abs-2206-10591, DBLP:journals/corr/abs-2305-00050}. These tasks are distinct from commonsense causal reasoning, where LLMs perform well \citep{DBLP:conf/semeval/GordonKR12, DBLP:conf/icml/0001ZSR22, DBLP:journals/corr/abs-2305-00050}.
Early experiments with GPT-4 \citep{DBLP:journals/corr/abs-2303-12712} showed that, despite presenting systematically better performance than its previous versions, it still has some innate limitations. The authors introduce several examples indicating that the autoregressive nature of LLMs may prevent them from planning and backtracking, two abilities necessary for complex reasoning \citep{DBLP:journals/corr/abs-2303-12712}. GPT-4 also does not always reason in a consistent manner. Although it produces consistent results more often than GPT-3, there are no guarantees that the process leading to the result is always correct. The scope of cognitive abilities of the system remain incompletely characterised, especially for precise reasoning \citep{DBLP:journals/corr/abs-2303-12712}.

The evaluations described above do not, of course, provide a measure of the intelligence or global cognitive abilities of those models;  measuring the level of intelligence of LLMs and other AI systems is challenging as there is no clear widely accepted definition \citep{DBLP:conf/aaai/BoochFHKLLLMMRS21, DBLP:journals/corr/abs-2011-15091}. \citet{DBLP:journals/corr/abs-1911-01547} defines the intelligence of a system as "a measure of its skill-acquisition efficiency over a scope of tasks, with respect to priors, experience, and generalization difficulty". Following this definition, abstract reasoning is a well-suited domain over which to measure aspects of the learning and generalisation abilities of a system. To this end, the Abstract Reasoning Challenge (ARC) has been proposed as a benchmark for artificial systems \citep{DBLP:journals/corr/abs-1911-01547}. A handful of works have proposed to measure abstract reasoning abilities in neural networks, but they focus on visual tasks \citep{DBLP:conf/icml/SantoroHBML18, DBLP:conf/cvpr/ZhangGJZZ19, DBLP:conf/cvpr/0017JEZZ21}. To the best of our knowledge, this paper is the first to present an extensive evaluation of abstract reasoning for Large Language Models. Other domains of study focus on problems similar to abstract reasoning. Notably, in program induction, DreamCoder is a system that learns to solve problems described by a small set of input-output pairs by writing programs \citep{DBLP:journals/corr/abs-2006-08381}. 
Abstract reasoning can also be related to causal representation learning, as finding abstract relations amounts to recovering the causal structure of a task and the Independent Causal Mechanisms (ICMs) linking the variables \citep{DBLP:journals/pieee/ScholkopfLBKKGB21, DBLP:journals/corr/abs-2302-00293}.

\section{Evaluation Method}

\subsection{Evaluation Data}
\label{sec:data}

To evaluate language models on a large variety of abstract reasoning tasks, we build a new framework that adapts text and vision datasets for abstract reasoning. 
We select the tasks based on their capacity to evaluate the ability of a system to find a general abstract rule from limited examples. The visual datasets are converted into text and symbolic versions to be used with language models. 
After formatting, the datasets can be divided into two categories: Open-Ended Question Answering (Open QA) and Multiple-Choice Question Answering (MCQA). Open QA datasets require the model to generate the correct answer, while MCQA requires it to choose the answer from a set of possible answers. We note that most of the evaluated models are built for general-purpose text generation. Therefore, even when choosing between several options, they must \textit{generate} the correct choice and may fail to do so (e.g. answering D when only options A, B, or C are available). For comparison, we also evaluate models built for question answering. We give more details in Section \ref{sec:models}. As shown in Figure \ref{fig:model_data_dep}, QA engines can only answer MCQA datasets, while text completion models can answer any type of question. Some MCQA datasets can also be converted to Open QA datasets by removing the choices. The datasets obtained are summarised in Table \ref{tab:datasets}.

\begin{figure}[ht]
  \centering
    
  \begin{minipage}[c]{0.48\linewidth}
  \begin{tikzpicture}[node distance=3.2cm,text width=1.1cm, font=\scriptsize, align=center]
    \node[draw, rectangle, fill=blue!25] (tc) {Text Completion};
    \node[draw, rectangle, fill=blue!25] (qa) [below of=tc] {QA Engine};
    \node[draw, circle, fill=green!25] (op) [right of=tc] {Open QA};
    \node[draw, circle, fill=green!25] (mc) [right of=qa] {MCQA};

    \draw[->, dashed] (tc) -- node[pos=0.5, above]{answers} (op);
    \draw[->, dashed] (tc) -- node[pos=0.5, sloped, above]{answers} (mc);
    \draw[->, dashed] (qa) -- node[pos=0.5, above]{answers} (mc);
    \draw[->] (mc) -- node[pos=0.5, sloped, above]{remove choices} (op);
  \end{tikzpicture}
  \caption{Different types of models and datasets considered in our experiments and their interactions. Dataset types are represented as green circles and model types are represented as blue rectangles. Text completion models can answer both types of datasets while QA engines can only answer MCQA datasets. 
  However, MCQA datasets can be altered to fit into the Open QA category.
  }
  \label{fig:model_data_dep}
  \end{minipage}
  \hfill
  \begin{minipage}[c]{0.48\linewidth}
    \centering
    \small
    \begin{tcolorbox}[title=Example Cases]
    {[9, 4]} $\rightarrow$ [9, 4]\\
    {[8, 2, 9, 4, 1, 7]} $\rightarrow$ [9, 4, 8, 2, 1, 7]\\
    {[5, 7, 3, 4, 9, 2, 0]} $\rightarrow$ [3, 4, 5, 7, 9, 2, 0]\\
    {[1, 5, 6, 4, 0, 3, 7]} $\rightarrow$ [6, 4, 1, 5, 0, 3, 7]
    \end{tcolorbox}
    \begin{tcolorbox}[title=Test Case]
    {[4, 3, 2, 8, 9, 6, 7, 1]} $\rightarrow$ \textit{\textcolor{red!50}{[2, 8, 4, 3, 9, 6, 7, 1]}}
    \end{tcolorbox}
    \caption{Example task in the BIG-Bench-F dataset. For this task, the system must return the input list with the first two elements switched with the following two if they exist. Pre-prompts are omitted from the input. In the test case, the target answer is indicated in \textit{\textcolor{red!50}{italics}}. }
    \label{fig:bench_ex}
  \end{minipage}
\end{figure}

\begin{table}[ht]
    \centering
    \begin{minipage}[t][][b]{.48\textwidth}
    \caption{Datasets considered. When not written, type is similar to the one above. Datasets can exist in text or symbolic versions. Text datasets built from an image dataset are indicated with the symbol $^T$. }
    \begin{tabular}{llccc}
        \hline
        \multirow{2}*{Dataset} & \multirow{2}*{Type} & \multicolumn{2}{c}{Versions} \\
        \cline{3-4}
         & & Text & Symb \\
        \hline
        ARC$^T$ & Open QA & & \checkmark \\
        BIG-Bench-F & & & \checkmark \\
        Evals-S & & & \checkmark \\
        PVR & & & \checkmark \\
        ACRE$^T$ & MCQA & \checkmark & \checkmark \\
        Evals-P & & & \checkmark \\
        RAVEN$^T$ & & \checkmark & \checkmark \\
        \hline
    \end{tabular}
    \label{tab:datasets}
    \end{minipage}
    \hfill
    \begin{minipage}[t][][b]{.48\textwidth}
    \caption{Models considered. When not written, type is similar to the one above. Models with the symbol $^*$ are introduced in this paper. "-AR" indicates that the model has been fine-tuned for abstract reasoning. }
    \begin{tabular}{ll}
        \hline
        Model & Type\\
        \hline
        GPT-2 & Text completion \\
        Text-Davinci-3 & \\
        GPT-3.5-Turbo \\
        GPT-4 &  \\
        LLaMA-7B & \\
        LLaMA2-7B & \\
        Alpaca & \\
        Alpaca-LoRA & \\
        Zephyr-7B-$\beta$ & \\
        LLaMA-7B-AR-LoRA$^*$ & \\
        LLaMA2-7B-AR-LoRA$^*$ & \\
        RoBERTa-AR$^*$ & QA Engine \\
        MERIt-AR$^*$ & \\
        \hline
    \end{tabular}
    \label{tab:models}
    \end{minipage}
\end{table}

\paragraph{Datasets}

We build a text-based version of the Abstract Causal Reasoning (ACRE) dataset \citep{DBLP:conf/cvpr/0017JEZZ21} that we name ACRE$^T$. ACRE is a Visual Question-Answering (VQA) dataset. Each sample in the data comprises six context images and four test cases. Each context image comprises a set of objects with various shapes, colours and textures, and a light. In the context images, the light can be on or off. The goal of a system is to determine from the context examples if the light is on, off, or if its state cannot be determined in the test cases. To solve this task, the model has to determine for each sample what objects are causally responsible for the activation of the light. We generate two versions of the dataset: in ACRE$^T$-Text, each image is replaced by a high-level textual description, and in ACRE$^T$-Symbolic, each image is replaced with a numerical vector representation.

The second dataset we build on is the Abstract Reasoning Challenge (ARC) dataset \citep{DBLP:journals/corr/abs-1911-01547}. The dataset is composed of tasks, each comprising three input and output grids. The goal of the system is to determine the algorithm that converts the input to the output and apply it to a test case. The grids have a variable size comprised between $8 \times 8$ and $30 \times 30$, and contain visual patterns (e.g. recognisable shapes, symmetries). We provide the raw grid to the model as a two-dimensional array of integers. We name this version ARC$^T$. The high dimensionality of the input makes it a challenging task for LLMs. The tasks themselves are also challenging as their transcription in natural language is often complex and supposedly impossible for 12\% of them \citep{DBLP:journals/corr/abs-2106-07824}.

We select a subset of the BIG-Bench dataset \citep{rule2020child, DBLP:journals/corr/abs-2206-04615} that we name BIG-Bench-F for \textit{Functions}. The subset comprises various tasks represented by a function taking a list as input and returning a new transformed list as output. For each task, several input-output samples are given. In BIG-Bench-F, we give four samples per task by default. The functions include typical list processing like replacing the value of one element, selecting a subset, or counting elements. An example is given in Figure \ref{fig:bench_ex}. The challenge in this task is to accurately recognise the function from a few samples.

We select a subset of the Evals dataset \citep{DBLP:journals/corr/abs-2303-08774} representing logic puzzles.
Evals-P is a set of tasks where a tuple containing a character and a list of characters is given as an input, and a single word from the set \{"foo", "bar"\} is generated from the input according to a logic hidden from the evaluated system. The task consists of finding the logic from a few samples and applying it to a test case. 
Evals-S is another set of tasks where a list of integers is given as an input, and an output list of words is generated. The task is the same as for Evals-P.

Pointer-Value Retrieval (PVR) tasks \citep{DBLP:journals/corr/abs-2107-12580} involve selecting one or several values in a list and applying a function on this subset. For each task, the system must recognise the retrieval and application functions and apply them to a test case. Samples are composed of a pointer-values pair and a label. The values are stored in an array, and the pointer is an integer pointing to an index in the array. The pointer indicates the subset of values to consider for the task. We generate a new dataset of PVR tasks following this methodology.

RAVEN \citep{DBLP:conf/cvpr/ZhangGJZZ19} is a VQA dataset composed of sequences of images to complete. The images contain Raven matrices \citep{raven1938raven}, i.e. geometric shapes (e.g. square, circle, pentagon) assembled together.
RAVEN is a dataset similar to Procedurally Generated Matrices (PGM) \citep{DBLP:conf/icml/SantoroHBML18} but also provides a tree structure describing the semantics of each image. We focus on a subset where a single shape appears in the image. The task is, given a sequence of eight images and eight possible choices, to pick the correct image that follows in the sequence. As RAVEN is a visual dataset like ACRE, we use the given semantic tree structure to generate a text description of each image we will feed to the evaluated models. We create two sets: RAVEN$^T$-Text contains natural language descriptions, and RAVEN$^T$-Symbolic contains symbolic descriptions. We also build another version of the dataset where choices are hidden. We name the former RAVEN$^T$-mcqa and the latter RAVEN$^T$-opqa.

\subsection{Models evaluated}
\label{sec:models}

We perform evaluations on the most recent and popular architectures for NLP tasks. Table \ref{tab:models} provides the list of models used in the experiments. More details are provided in the appendix.
We restrict our experiments to Large Language models (or \textit{Foundation Models} \citep{DBLP:journals/corr/abs-2108-07258}). We conduct experiments on the popular family of GPT architectures. We include three generations of GPT models: GPT-2 \citep{radford2019language}, a 1.5B parameter model;  aligned GPT-3 models with Text-Davinci-3, optimised for text completion, and GPT-3.5-Turbo, optimised for chat, two 175B models \citep{DBLP:conf/nips/BrownMRSKDNSSAA20, DBLP:conf/nips/Ouyang0JAWMZASR22}; and GPT-4, with unknown training and architectural details \citep{DBLP:journals/corr/abs-2303-08774}.
We also perform experiments on the popular open models LLaMA \citep{DBLP:journals/corr/abs-2302-13971} and LLaMA2 \citep{DBLP:journals/corr/abs-2307-09288}. Alpaca is a fine-tuned version of LLaMA to respond to instructions \citep{DBLP:journals/corr/abs-2212-10560, alpaca}, and Alpaca-LoRA is a LLaMA model instruction-tuned using Low-Rank Adaptation \citep{DBLP:conf/iclr/HuSWALWWC22}. We also fine-tune our own LLaMA and LLaMA2 models for abstract reasoning. For all models, we evaluate the 7B parameters versions by default. Finally, we evaluate the more recent Zephyr-7B-$\beta$ \citep{DBLP:journals/corr/abs-2310-16944, alignment_handbook2023}, a 7B parameters model fine-tuned from Mistral-7B \citep{DBLP:journals/corr/abs-2310-06825}.
We also compare these generic models on architecture fine-tuned for Multiple-Choice Question Answering. Unlike the text completion engines that produce text in the output, their task consists of discriminating the solution from a small set of options. This problem is more straightforward to solve than the problem of next token prediction tackled by the models described in the previous paragraph. We fine-tune two models for Multiple-Choice Question Answering: RoBERTa-large \citep{DBLP:journals/corr/abs-1907-11692}, a language model used for text comprehension, and MERIt \citep{DBLP:conf/acl/JiaoGSN22}, a model using contrastive pre-training on rules-based data to perform logical reasoning.

\section{Experiments}

\subsection{Open-Ended Question Answering}
\label{sec:opqa}

In this section, we detail our experiments on open-ended abstract reasoning. 
Depending on the dataset, the answer can be in natural language or a symbolic format. The model is asked to provide the answer directly. The accuracy for each model on every dataset is summarised in Table \ref{tab:open_qa_results}.

\begin{table}[ht]
    \centering
    \caption{Accuracy of Large Language Models on Open QA datasets. Datasets are represented in columns, and models in rows. The best result for each dataset is indicated in \textbf{bold}, and the second best is indicated in \textit{italics}. }
    \begin{tabular}{lccccccc}
        \hline
         & \multirow{2}*{ARC$^T$} & \multirow{2}*{BIG-Bench-F} & \multirow{2}*{Evals-S} & \multirow{2}*{PVR} & \multicolumn{2}{c}{RAVEN$^T$-opqa} \\
        \cline{6-7}
         & & & & & Text & Symb \\
        \hline
        Text-Davinci-3 & \textit{0.105} & \textit{0.404} & \textbf{0.314} & \textbf{0.228} & \textit{0.343} & \textit{0.234} \\
        GPT-3.5-Turbo & 0.033 & 0.153 & 0.186 & 0.124 & 0.226 & 0.161 \\
        GPT-4 & \textbf{0.119} & \textbf{0.514} & \textit{0.304} & 0.177 & \textbf{0.410} & \textbf{0.330} \\
        LLaMA-7B & 0.010 & 0.012 & 0.014 & 0.060 & 0.000 & 0.000 \\
        LLaMA2-7B & 0.005 & 0.108 & 0.000 & 0.000 & 0.000 & 0.001 \\
        Alpaca & 0.010 & 0.188 & 0.014 & 0.184 & 0.075 & 0.030 \\
        Alpaca-LoRA & 0.012 & 0.144 & 0.000 & 0.152 & 0.000 & 0.067 \\
        Zephyr-7B-$\beta$ & 0.015 & 0.292 & 0.043 & \textit{0.209} & 0.009 & 0.145 \\
        \hline
    \end{tabular}
    \label{tab:open_qa_results}
\end{table}

Our results indicate poor performance of language models on all the presented datasets, although the performance varies between datasets and models. In particular, Text-Davinci-3 and GPT-4 consistently achieve the best performance across the datasets. Zephyr-7B-$\beta$ has almost systematically the best accuracy among open models. 
On the other hand, LLaMA-7B has the worst performance of all models. LLaMA2-7B gets a similar accuracy except on BIG-Bench. Alpaca and Alpaca-LoRA present slight improvements on BIG-Bench-F, PVR and RAVEN$^T$. This improvement is explained by the instruction-tuning used to build Alpaca and Alpaca-LoRA. 
We provide several examples in the appendix that illustrate this difference. LLaMA-7B often does not attempt to solve the problem but completes the text by giving more examples. These examples do no match the abstract rule for the task. Alpaca and Alpaca-LoRA follow the instructions more faithfully but also fail to grasp the abstract patterns. Instruction-tuning seems to help the model understand the format of the answer and what it is asked to do but provides little help on how to solve the tasks. Moreover, the performance difference between Text-Davinci-3 and GPT-3.5-Turbo indicates that the type of instruction-tuning matters as Text-Davinci-3 performs systematically better than GPT-3.5-Turbo despite being based on the same model. 
Overall, GPT-4 performs noticeably better than all the other models. As the details of its architecture and training set are unavailable, we cannot provide satisfactory explanations for this difference. However, the increase in performance is highest on the RAVEN$^T$ dataset. Given that Raven matrices are a standard and long-existing test \citep{raven1938raven, carpenter1990one}, we can hypothesize that the training data of GPT-4 included some versions of the test. The same remark can be made for BIG-Bench-F as it includes traditional list processing algorithms.
Text-Davinci-3 and GPT-4 also achieve good performance on the ARC$^T$ dataset relative to other existing architectures challenged on the task, making them 11$^{th}$ and 14$^{th}$ on the Kaggle leaderboard\footnote{\url{https://www.kaggle.com/competitions/abstraction-and-reasoning-challenge/leaderboard}}. However, they still fail to answer a vast majority of the tasks correctly. All LLMs generally fail to answer most of the tasks in each dataset. Despite a performance increase compared to previous versions, the most recent language models do not perform open-ended abstract reasoning well.

\subsection{Multiple-Choice Question Answering}
\label{sec:mcqa}

As seen in Section \ref{sec:opqa}, open-ended abstract reasoning is a challenging problem for language models. We also perform a series of experiments on Multiple-Choice Question Answering tasks. For these tasks, the models are given a set of possible answers and must pick a single one from the set. This task is more accessible than Open-Ended QA, as the valid response is given as part of the input. Results are given in Table \ref{tab:mc_qa_results}.

% \begin{wraptable}{r}{9.2cm}
\begin{table}[ht]
    \centering
    \caption{Accuracy of Large Language Models for Multiple-Choice QA on the ACRE$^T$, Evals-P and RAVEN$^T$ datasets. 
    The last line indicates random performance. Completion models can perform worse than random if they do not reply with a valid answer. The best result for each dataset is indicated in \textbf{bold}, and the second best is indicated in \textit{italics}. }
    \begin{tabular}{lccccc}
        \hline
         & \multicolumn{2}{c}{ACRE$^T$} & \multirow{2}{*}{Evals-P} & \multicolumn{2}{c}{RAVEN$^T$-mcqa} \\
        \cline{2-3}\cline{5-6}
         & Text & Symb & & Text & Symb \\
        \hline
        GPT-2 & \textbf{0.371} & 0.00 & 0.496 & 0.00 & 0.126 \\
        Text-Davinci-3 & 0.098 & 0.427 & \textit{0.560} & \textit{0.461} & \textit{0.452} \\
        GPT-3.5-Turbo & 0.184 & 0.445 & 0.481 & 0.276 & 0.315 \\
        GPT-4 & \textit{0.272} & \textit{0.512} & \textbf{0.625} &\textbf{0.697} & \textbf{0.535} \\
        LLaMA-7B & 0.000 & 0.257 & 0.544 & 0.004 & 0.000 \\
        LLaMA2-7B & 0.014 & 0.003 & 0.500 & 0.026 & 0.149 \\
        Alpaca & 0.036 & 0.238 & 0.544 & 0.015 & 0.058 \\
        Alpaca-LoRA & 0.015 & 0.123 & 0.552 & 0.082 & 0.124 \\
        Zephyr-7B-$\beta$ & 0.106 & \textbf{0.516} & 0.504 & 0.000 & 0.022 \\
        \hline
        random & 0.33 & 0.33 & 0.5 & 0.125 & 0.125 \\
        \hline
    \end{tabular}
    \label{tab:mc_qa_results}
\end{table}
% \end{wraptable}

We first compare the results of RAVEN$^T$-mcqa and RAVEN$^T$-opqa from Table \ref{tab:open_qa_results}. RAVEN$^T$-opqa contains the same questions as RAVEN$^T$-mcqa, but the answer choices have been removed. Following intuition, giving multiple choices to LLMs helps systematically improve their performance. Only the performance of LLaMA remains the same, and the performance of Alpaca and Zephyr-7B-$\beta$ are slightly reduced. Given the low accuracy in both cases, it can be interpreted as noise.
MCQA models achieve slightly above random performance (see details in appendix), performing better than most LLMs. However, they have an advantage compared to completion engines as they have to select one answer among a list of possible choices, whereas completion models must generate the correct answer. Therefore, the latter may not return any valuable output (e.g. a nonsensical or empty answer), explaining how they can achieve worse than random performance.
The main takeaway from these experiments is that the performance of LLMs remains low even in discriminative settings. When given a set of possible answers, the models cannot recognise the proper solution among the other choices. This finding indicates that using LLMs as evaluators (as done in self-refinement techniques \citep{DBLP:journals/corr/abs-2303-17651}) is not suited for tasks requiring abstract reasoning.
We confirm this with additional experiments in the appendix using different refinement strategies.
Additionally, when comparing the results between natural language and symbolic tasks on ACRE$^T$, we observe that the results are better across all models when the input is symbolic. Inputs that use symbolic data are smaller and may convey only relevant information, while natural language could contain distracting information or biases harmful to task performance. The same observation can be made concerning RAVEN$^T$-mcqa, except for GPT-4.
In the open-ended version of RAVEN$^T$, models perform better with the natural language representation. Without the answer set available, inductive biases caused by language help performance.

\subsection{Chain-of-Thought Prompting}

We perform experiments on a subset of our framework using \textit{Chain-of-Thought} prompting \citep{DBLP:conf/nips/Wei0SBIXCLZ22}. The complete experiments are provided in the appendix (and include a side-by-side comparison for better readability). We perform experiments with GPT-3.5-turbo, GPT-4, and Alpaca-LoRA. Our experiments with \textit{Chain-of-Thought} have the suffix \textit{model}-cot. Our results are presented in Table \ref{tab:results_cot_concise}.
Overall, the results obtained using \textit{Chain-of-Thought} prompting are not higher than those obtained with the base models. On The BIG-Bench-F dataset, the \textit{Chain-of-Thought} versions achieve systematically lower performance than their base counterparts, although no no significant performance drop is observed. On PVR and RAVEN$^T$-opqa, while the accuracy for GPT-4 and Alpaca-LoRA remain unchanged or slightly reduced, the performance of GPT-3.5-Turbo is increased. On RAVEN$^T$-mcqa, the performance of all the models decreases. These experiments show that the quality of the prompt has little impact on the answer quality. It hints that the models can understand the task, but their failures are due to their ability to provide faithful reasoning. This limitation is further illustrated with examples in the appendix.

\begin{table}[ht]
    \centering
    \caption{Accuracy of Large Language Models on Open and Multiple-Choice QA datasets when prompted using \textit{Chain-of-Thought}. Datasets are represented in columns, and models in rows. The best result for each dataset is indicated in \textbf{bold}, and the second best is indicated in \textit{italics}. BBF stands for BIG-Bench-F. }
    \begin{minipage}[t][][b]{.61\textwidth}
    \begin{tabular}{lcccc}
        \hline
         & \multirow{2}*{BBF} & \multirow{2}*{PVR} & \multicolumn{2}{c}{RAVEN$^T$-opqa} \\
        \cline{4-5}
         & & & Text & Symb \\
        \hline
        GPT-3.5-Turbo-cot & \textit{0.097} & \textbf{0.210} & \textit{0.302} & \textit{0.211} \\
        GPT-4-cot & \textbf{0.476} & \textit{0.174} & \textbf{0.385} & \textbf{0.354} \\
        Alpaca-LoRA-cot & 0.084 & 0.152 & 0.000 & 0.069 \\
        \hline
    \end{tabular}
    \end{minipage}
    \hfill
    \begin{minipage}[t][][b]{.38\textwidth}
    \begin{tabular}{cccc}
        \hline
         \multicolumn{2}{c}{ACRE$^T$} & \multicolumn{2}{c}{RAVEN$^T$-mcqa} \\
        \cline{1-2}\cline{3-4}
         Text & Symb & Text & Symb \\
        \hline
        \textbf{0.255} & \textit{0.345} & \textit{0.257} &\textit{ 0.144} \\
        \textit{0.214} & \textbf{0.394} & \textbf{0.596} & \textbf{0.517} \\
        \textit{0.059} & 0.114 & 0.000 & 0.114  \\
        \hline
    \end{tabular}
    \end{minipage}
    \label{tab:results_cot_concise}
\end{table}

\subsection{Varying the Example Set Size}
\label{sec:var_ex}

We perform further experiments on the BIG-Bench-F and PVR datasets. For these two datasets, we alter the number of examples given to the system before the test case. By default, we give four examples to the model before asking it to answer.
The results are shown in Figures \ref{fig:res_incr_bench_size} and \ref{fig:res_incr_pvr_size}. In this section, we focus on the results of the base models (without the "-code" suffix).
We first observe that, for both datasets, there is no linear relationship linking performance and number of examples. For all but the Text-Davinci-3 and GPT-4 models, adding more examples has little or no effect on the accuracy. 
Text-Davinci-3 and GPT-4 perform similarly across all cases, and their performances consistently increase with the number of examples, achieving up to an accuracy of 0.6 when given 16 examples on the BIG-Bench-F dataset. However, on PVR, Text-Davinci-3 achieves only 0.26 when given 12 examples. GPT-4 follows a similar trend but performs slightly worse than its predecessor. In the Absence of technical details for GPT-4, we can only speculate on the reasons. As this effect is observed only on BIG-Bench-F and not PVR, we can assume that the models perform better because their training sets contain the list processing algorithms used by BIG-Bench-F.
We perform additional experiments in the appendix, where we provide solved instances into the prompt (input and solution program) to propel the model to reason correctly. No real improvements are observed.

\begin{figure}
    \centering
    \begin{subfigure}{0.48\linewidth}
    \begin{adjustbox}{width=0.95\linewidth}
    \begin{tikzpicture}
    \begin{axis}[
        xlabel={Number of example cases},
        ylabel={Accuracy},
        xmin=2, xmax=16,
        ymin=0, ymax=1,
        legend pos=north west,
        legend columns=2
        ymajorgrids=true,
        grid style=dashed,
    ]
    
    \addplot[
        color=green,
        mark=o,
        ]
        coordinates {
        (2,0.316)(4,0.404)(6,0.512)(8,0.504)(10,0.564)(12,0.576)(14,0.588)(16,0.604)
        };
        % \addlegendentry{text-davinci-3}
    
    \addplot[
        color=green,
        mark=o,
        dashed
        ]
        coordinates {
        (2,0.264)(4,0.272)(6,0.292)(8,0.252)(10,0.256)(12,0.268)(14,0.268)(16,0.268)
        };
        % \addlegendentry{text-davinci-3-code}
    
    \addplot[
        color=blue,
        mark=diamond,
        ]
        coordinates {
        (2,0.12)(4,0.153)(6,0.168)(8,0.175)(10,0.172)(12,0.161)(14,0.128)(16,0.169)
        };
        % \addlegendentry{gpt-3.5-turbo}
    
    \addplot[
        color=blue,
        mark=diamond,
        dashed
        ]
        coordinates {
        (2,0.124)(4,0.152)(6,0.108)(8,0.136)(10,0.156)(12,0.176)(14,0.196)(16,0.224)
        };
        % \addlegendentry{gpt-3.5-turbo-code}
    
    \addplot[
        color=cyan,
        mark=square,
        ]
        coordinates {
        (2,0.364)(4, 0.516)(6,0.563)(8,0.544)(10,0.596)(12,0.632)(14,0.628)(16,0.632)
        };
        % \addlegendentry{gpt-4}
    
    \addplot[
        color=cyan,
        mark=square,
        dashed
        ]
        coordinates {
        (2,0.325)(4,0.390)(6,0.398)(8,0.423)(10,0.422)(12,0.430)(14,0.428)(16,0.476)
        };
        % \addlegendentry{gpt-4-code}

    \addplot[
        color=orange,
        mark=halfcircle*,
        ]
        coordinates {
        (2,0.104)(4,0.188)(6,0.18)(8,0.168)(10,0.136)(12,0.164)(14,0.124)(16,0.164)
        };
        % \addlegendentry{alpaca}
    
    \addplot[
        color=orange,
        mark=halfcircle*,
        dashed
        ]
        coordinates {
        (2,0.004)(4,0.008)(4,0.004)(8,0.016)(10,0.008)(12,0.000)(14,0.008)(16,0.008)
        };
        % \addlegendentry{alpaca-code}
    
    \addplot[
        color=red,
        mark=x,
        ]
        coordinates {
        (2,0.152)(4,0.144)(6,0.116)(8,0.08)(10,0.108)(12,0.136)(14,0.128)(16,0.112)
        };
        % \addlegendentry{alpaca-lora}
    
    \addplot[
        color=red,
        mark=x,
        dashed
        ]
        coordinates {
        (2,0.02)(4,0.06)(6,0.02)(8,0.02)(10,0.032)(12,0.052)(14,0.04)(16,0.015)
        };
        % \addlegendentry{alpaca-lora-code}
        
    \end{axis}
    \end{tikzpicture}
    \end{adjustbox}
    \caption{BIG-Bench-F dataset. }
    \label{fig:res_incr_bench_size}
    \end{subfigure}
    \hfill
    \begin{subfigure}{0.48\linewidth}
    \centering
    \begin{adjustbox}{width=0.95\linewidth}
    \begin{tikzpicture}
    \begin{axis}[
        xlabel={Number of example cases},
        ylabel={Accuracy},
        xmin=4, xmax=16,
        ymin=0, ymax=1,
        legend pos=north west,
        legend columns=2
        ymajorgrids=true,
        grid style=dashed,
    ]
    
    \addplot[
        color=green,
        mark=o,
        ]
        coordinates {
        (4,0.228)(8,0.24)(12,0.272)(16,0.26)
        };
        \addlegendentry{text-davinci-3}
    
    \addplot[
        color=green,
        mark=o,
        dashed
        ]
        coordinates {
        (4,0.208)(8,0.184)(12,0.188)(16,0.168)
        };
        \addlegendentry{text-davinci-3-code}
    
    \addplot[
        color=blue,
        mark=diamond,
        ]
        coordinates {
        (4,0.124)(8,0.124)(12,0.112)(16,0.084)
        };
        \addlegendentry{gpt-3.5-turbo}
    
    \addplot[
        color=blue,
        mark=diamond,
        dashed
        ]
        coordinates {
        (4,0.204)(8,0.252)(12,0.22)(16,0.228)
        };
        \addlegendentry{gpt-3.5-turbo-code}
    
    \addplot[
        color=cyan,
        mark=square,
        ]
        coordinates {
        (4,0.177)(8,0.196)(12,0.235)(16,0.244)
        };
        \addlegendentry{gpt-4}
    
    \addplot[
        color=cyan,
        mark=square,
        dashed
        ]
        coordinates {
        (4,0.189)(8,0.185)(12,0.197)(16,0.197)
        };
        \addlegendentry{gpt-4-code}
    
    \addplot[
        color=orange,
        mark=halfcircle*,
        ]
        coordinates {
        (4,0.184)(8,0.184)(12,0.192)(16,0.208)
        };
        \addlegendentry{alpaca}
    
    \addplot[
        color=orange,
        mark=halfcircle*,
        dashed
        ]
        coordinates {
        (4,0.012)(8,0.008)(12,0.012)(16,0.008)
        };
        \addlegendentry{alpaca-code}
    
    \addplot[
        color=red,
        mark=x,
        ]
        coordinates {
        (4,0.152)(8,0.14)(12,0.124)(16,0.2)
        };
        \addlegendentry{alpaca-lora}
    
    \addplot[
        color=red,
        mark=x,
        dashed
        ]
        coordinates {
        (4,0.004)(8,0.008)(12,0.0)(16,0.0)
        };
        \addlegendentry{alpaca-lora-code}

    \end{axis}
    \end{tikzpicture}
    \end{adjustbox}
    \caption{PVR dataset. }
    \label{fig:res_incr_pvr_size}
    \end{subfigure}
    \caption{Evolution of the model performance as a function of the number of examples seen from the dataset. The legend is shared by both figures. Models with straight lines are used with default prompting, while models with dashed lines are prompted to produce code. }
    
\end{figure}

\subsection{Enabling Structure Discovery with Code}
\label{sec:code}

In the next experiments, we follow an idea similar to \textit{Progam-of-Thought} prompting \citep{DBLP:journals/corr/abs-2211-12588} and ask the model to generate the code of the function responsible for generating the output from the input. Then, we execute the produced code on the test case and evaluate the result. This method differs from a base prompt as we do not ask the model to produce the answer directly. This part is delegated to a code interpreter in Python. This method aims to verify the ability of LLMs to extract the correct structure behind each abstract reasoning task under code format. We test this method on the BIG-Bench-F and PVR datasets. The results of these models (with the "-code" suffix) can be compared with their original counterparts in Figures \ref{fig:res_incr_bench_size} and \ref{fig:res_incr_pvr_size}.
In general, we observe that the models prompted to produce code perform worse than those tasked to produce the answer directly. The only exception is GPT-3.5-Turbo. On the BIG-Bench-F dataset, the performance of GPT-3.5-Turbo-code increases steadily while that of GPT-3.5-Turbo stagnates, and on PVR, GPT-3.5-Turbo-code outperforms GPT-3.5-Turbo by a significant margin. Producing code solving the abstract problem is a more complicated task for an LLM as it requires the model to produce a rigorous code explanation for its answer. It is consistent with the results for most models, but we also observe in the case of GPT-3.5-Turbo-code that it can help the model better understand the task. On BIG-Bench-F, the code versions of Text-Davinci-3 and GPT-4 perform better than both base and code versions of the other models. %, showing a good capacity to understand the abstract structure of the tasks.  
As this behaviour is not observed with PVR, we infer that this performance is due to the functions being part of the training sets of the models.
The models can almost always generate code able to compile and produce an answer (details are in the appendix). We deduce that producing a program with a valid syntax is not a bottleneck for performance. The issue lies in the recovery of the correct reasoning process.

\subsection{Fine-tuning LLaMA2}

We now study the performance of LLaMA2 models after fine-tuning on RAVEN$^T$-mcqa. Experiments on more datasets are provided in the appendix. The training and test sets may share distribution-specific patterns that the model may learn during the fine-tuning phase. It may overfit on these patterns instead of learning the correct abstract patterns. To alleviate this pitfall, we generate out-of-distribution (o.o.d) splits. The \textit{-Four} split contains samples with four figures instead of one. The \textit{-In-Center} splits contains samples with two figures instead of one, a big and a small located within the former. The shape and colours of the figures all are observed in the training set. The two splits can be considered as compositional splits. The results on RAVEN$^T$-mcqa are shown in Table \ref{tab:fine-tuned-raven-short}. 
We observe a significant increase in the accuracy on the test set. LLaMA2 achieves close to perfect accuracy. The performance partially transfers to the alternative syntax task.
We now observe the performance on the o.o.d splits. The performance of the fine-tuned LLaMA2 significantly drops on the new tasks, showing a lack of generalisation. We can deduce that fine-tuning yields representations that are highly invariant to the syntax but does not transfer other abstract reasoning abilities. The rules required to solve the \textit{-Four} and \textit{-In-Center} splits manipulate several figures, they are compositions of rules used for single figures. LLMs can compose with unseen quantities (e.g. new syntax) but have more difficulty composing new abstract rules.

\begin{table}[t]
    \centering
    \caption{Accuracy of base and fine-tuned LLaMA2 on the RAVEN$^T$-mcqa dataset i.i.d and o.o.d splits. Rows represent the dataset on which the model is fine-tuned, and columns represent the dataset on which the model is evaluated. The best result for each dataset in indicated in \textbf{bold}. }
    \begin{tabular}{lllcccccc}
        \hline
        \multirow{2}{*}{Model} & \multicolumn{2}{c}{Test Set $\Rightarrow$} & \multicolumn{2}{c}{RAVEN$^T$-Eval} & \multicolumn{2}{c}{-Four} & \multicolumn{2}{c}{-In-Center} \\
        \cline{4-9}
         & \multicolumn{2}{c}{Tuning Set $\Downarrow$} & Text & Symb & Text & Symb & Text & Symb \\
        \hline
        LLaMA2-7B & & & 0.135 & 0.114 & 0.073 & 0.121 & 0.000 & 0.001 \\
        \multirow{2}*{~~-AR-LoRA$^*$} & \multirow{2}*{RAVEN$^T$-Train} & Text & \textbf{0.977} & 0.694 & \textbf{0.557} & \textbf{0.522} & 0.536 & \textbf{0.085} \\
        & & Symb & 0.965 & \textbf{0.938} & 0.498 & 0.442 & \textbf{0.767} & 0.064 \\
        \hline
    \end{tabular}
    \label{tab:fine-tuned-raven-short}
\end{table}

\subsection{A Perspective from Causal Induction}

We perform further analysis on ACRE$^T$. The dataset can be divided into four causal paths: Direct, Indirect, Backward-blocking, Screening-off \citep{DBLP:conf/cvpr/0017JEZZ21}. 
Direct path queries can be established using direct evidence. Indirect paths require the combination of multiple pieces of evidence. Backward-blocking paths cannot be determined because the true mechanisms cannot be discriminated from other possibilities based solely on the data. Screening-off paths are causal paths affected by spurious correlations. 
Figure \ref{fig:acre_causal_res_cot_short} shows the results for each type of query. We restrict our analysis to the \textit{Chain-of-Thought} models (see the appendix for the full analysis). 
Although accuracy scores are similar, the distribution of the results among the causal paths differs between models and input types. GPT models overfit to backward-blocking cases on the text ACRE$^T$ but not on the symbolic version. We can deduce that natural language contains distracting information or biases harmful to abstract reasoning performance. It is consistent with the higher score of the models on the symbolic tasks.

\begin{figure}
    \centering
    \begin{subfigure}{0.48\linewidth}
    \begin{adjustbox}{width=0.8\linewidth}
    \begin{tikzpicture}
    \begin{axis}[
        ybar,
        ymin=0,
        ymax=1,
        ylabel={Accuracy},
        enlarge x limits={abs=0.6},
        xtick=data,
        tick label style={rotate=15},
        xticklabels={GPT-3.5-Turbo-cot,GPT-4-cot,Alpaca-LoRA-cot}
        ]
    
    \addplot coordinates {(1,0.076) (2,0.124) (3,0.070)};
    \addplot coordinates {(1,0.009) (2,0.026) (3,0.172)};
    \addplot coordinates {(1,0.798) (2,0.530) (3,0.020)};
    \addplot coordinates {(1,0.094) (2,0.119) (3,0.029)};

    \legend {Direct, Indirect, Backward-blocking, Screening-off};
    \end{axis}
    \end{tikzpicture}
    \end{adjustbox}
    \caption{Text ACRE$^T$. }
    \end{subfigure}
    \hfill
    \begin{subfigure}{0.48\linewidth}
    \begin{adjustbox}{width=0.8\linewidth}
    \begin{tikzpicture}
    \begin{axis}[
        ybar,
        ymin=0,
        ymax=1,
        ylabel={Accuracy},
        enlarge x limits={abs=0.6},
        xtick=data,
        tick label style={rotate=15},
        xticklabels={GPT-3.5-Turbo-cot,GPT-4-cot,Alpaca-LoRA-cot}
        ]
    
    \addplot coordinates {(1,0.462) (2,0.475) (3,0.104)};
    \addplot coordinates {(1,0.422) (2,0.500) (3,0.164)};
    \addplot coordinates {(1,0.107) (2,0.154) (3,0.000)};
    \addplot coordinates {(1,0.366) (2,0.462) (3,0.231)};

    \legend {Direct, Indirect, Backward-blocking, Screening-off};
    \end{axis}
    \end{tikzpicture}
    \end{adjustbox}
    \caption{Symbolic ACRE$^T$. }
    \end{subfigure}
    \caption{Results of chain-of-thought models on ACRE$^T$ divided by causal paths. }
    \label{fig:acre_causal_res_cot_short}
\end{figure}

\section{Conclusion}

Understanding the potential reasoning capabilities of LLMs is crucial as they are starting to be widely adopted. Measuring the level of intelligence of a system is hard, but abstract reasoning provides a valuable framework for this task. In this paper, we present what is, to the best of our knowledge, the first extensive evaluation of Large Language Models for abstract reasoning. 
We show that LLMs do not perform well on all types of tasks, although not all models are equally poor. 
Prompting and refinement techniques that improve performance on NLP tasks do not work for abstract reasoning. 
Our experiments show that the bottleneck in the performance lies in the recognition of new unseen abstract patterns and not in a lack of understanding of the task or the prompt. These results hold in discriminative settings, where the models must find the correct answer within a small set of propositions.
A qualitative study of selected failure cases in the appendix further reveals that models tend to reason inconsistently and in a shallow way. 
We hypothesise that current self-supervised autoregressive LLMs lack fundamental properties for strong abstract reasoning tasks and human-like cognition. 
We posit that methods based on causal reasoning and program induction could help improve the reasoning abilities of neural networks.

\printbibliography

@article{DBLP:journals/corr/abs-2206-04615,
  author       = {Aarohi Srivastava and
                  Abhinav Rastogi and
                  Abhishek Rao and
                  Abu Awal Md Shoeb and
                  Abubakar Abid and
                  Adam Fisch and
                  Adam R. Brown and
                  Adam Santoro and
                  Aditya Gupta and
                  Adri{\`{a}} Garriga{-}Alonso and
                  Agnieszka Kluska and
                  Aitor Lewkowycz and
                  Akshat Agarwal and
                  Alethea Power and
                  Alex Ray and
                  Alex Warstadt and
                  Alexander W. Kocurek and
                  Ali Safaya and
                  Ali Tazarv and
                  Alice Xiang and
                  Alicia Parrish and
                  Allen Nie and
                  Aman Hussain and
                  Amanda Askell and
                  Amanda Dsouza and
                  Ameet Rahane and
                  Anantharaman S. Iyer and
                  Anders Andreassen and
                  Andrea Santilli and
                  Andreas Stuhlm{\"{u}}ller and
                  Andrew M. Dai and
                  Andrew La and
                  Andrew K. Lampinen and
                  Andy Zou and
                  Angela Jiang and
                  Angelica Chen and
                  Anh Vuong and
                  Animesh Gupta and
                  Anna Gottardi and
                  Antonio Norelli and
                  Anu Venkatesh and
                  Arash Gholamidavoodi and
                  Arfa Tabassum and
                  Arul Menezes and
                  Arun Kirubarajan and
                  Asher Mullokandov and
                  Ashish Sabharwal and
                  Austin Herrick and
                  Avia Efrat and
                  Aykut Erdem and
                  Ayla Karakas and
                  et al.},
  title        = {Beyond the Imitation Game: Quantifying and extrapolating the capabilities
                  of language models},
  journal      = {CoRR},
  volume       = {abs/2206.04615},
  year         = {2022},
  url          = {https://doi.org/10.48550/arXiv.2206.04615},
  doi          = {10.48550/arXiv.2206.04615},
  eprinttype    = {arXiv},
  eprint       = {2206.04615},
  bibsource    = {dblp computer science bibliography, https://dblp.org}
}

@phdthesis{rule2020child,
  title={The child as hacker: building more human-like models of learning},
  author={Rule, Joshua Stewart},
  year={2020},
  school={Massachusetts Institute of Technology}
}

@inproceedings{DBLP:conf/cvpr/0017JEZZ21,
  author       = {Chi Zhang and
                  Baoxiong Jia and
                  Mark Edmonds and
                  Song{-}Chun Zhu and
                  Yixin Zhu},
  title        = {{ACRE:} Abstract Causal REasoning Beyond Covariation},
  booktitle    = {{IEEE} Conference on Computer Vision and Pattern Recognition, {CVPR}
                  2021, virtual, June 19-25, 2021},
  pages        = {10643--10653},
  publisher    = {Computer Vision Foundation / {IEEE}},
  year         = {2021},
  url          = {https://openaccess.thecvf.com/content/CVPR2021/html/Zhang\_ACRE\_Abstract\_Causal\_REasoning\_Beyond\_Covariation\_CVPR\_2021\_paper.html},
  doi          = {10.1109/CVPR46437.2021.01050},
  bibsource    = {dblp computer science bibliography, https://dblp.org}
}

@article{DBLP:journals/corr/abs-2107-12580,
  author       = {Chiyuan Zhang and
                  Maithra Raghu and
                  Jon M. Kleinberg and
                  Samy Bengio},
  title        = {Pointer Value Retrieval: {A} new benchmark for understanding the limits
                  of neural network generalization},
  journal      = {CoRR},
  volume       = {abs/2107.12580},
  year         = {2021},
  url          = {https://arxiv.org/abs/2107.12580},
  eprinttype    = {arXiv},
  eprint       = {2107.12580},
  bibsource    = {dblp computer science bibliography, https://dblp.org}
}

@inproceedings{DBLP:conf/cvpr/ZhangGJZZ19,
  author       = {Chi Zhang and
                  Feng Gao and
                  Baoxiong Jia and
                  Yixin Zhu and
                  Song{-}Chun Zhu},
  title        = {{RAVEN:} {A} Dataset for Relational and Analogical Visual REasoNing},
  booktitle    = {{IEEE} Conference on Computer Vision and Pattern Recognition, {CVPR}
                  2019, Long Beach, CA, USA, June 16-20, 2019},
  pages        = {5317--5327},
  publisher    = {Computer Vision Foundation / {IEEE}},
  year         = {2019},
  url          = {http://openaccess.thecvf.com/content\_CVPR\_2019/html/Zhang\_RAVEN\_A\_Dataset\_for\_Relational\_and\_Analogical\_Visual\_REasoNing\_CVPR\_2019\_paper.html},
  doi          = {10.1109/CVPR.2019.00546},
  bibsource    = {dblp computer science bibliography, https://dblp.org}
}

@inproceedings{DBLP:conf/icml/SantoroHBML18,
  author       = {Adam Santoro and
                  Felix Hill and
                  David G. T. Barrett and
                  Ari S. Morcos and
                  Timothy P. Lillicrap},
  editor       = {Jennifer G. Dy and
                  Andreas Krause},
  title        = {Measuring abstract reasoning in neural networks},
  booktitle    = {Proceedings of the 35th International Conference on Machine Learning,
                  {ICML} 2018, Stockholmsm{\"{a}}ssan, Stockholm, Sweden, July
                  10-15, 2018},
  series       = {Proceedings of Machine Learning Research},
  volume       = {80},
  pages        = {4477--4486},
  publisher    = {{PMLR}},
  year         = {2018},
  url          = {http://proceedings.mlr.press/v80/santoro18a.html},
  bibsource    = {dblp computer science bibliography, https://dblp.org}
}

@article{DBLP:journals/corr/abs-2106-07824,
  author       = {Samuel Acquaviva and
                  Yewen Pu and
                  Marta Kryven and
                  Catherine Wong and
                  Gabrielle E. Ecanow and
                  Maxwell I. Nye and
                  Theodoros Sechopoulos and
                  Michael Henry Tessler and
                  Joshua B. Tenenbaum},
  title        = {Communicating Natural Programs to Humans and Machines},
  journal      = {CoRR},
  volume       = {abs/2106.07824},
  year         = {2021},
  url          = {https://arxiv.org/abs/2106.07824},
  eprinttype    = {arXiv},
  eprint       = {2106.07824},
  bibsource    = {dblp computer science bibliography, https://dblp.org}
}

@article{DBLP:journals/corr/abs-2108-07258,
  author       = {Rishi Bommasani and
                  Drew A. Hudson and
                  Ehsan Adeli and
                  Russ B. Altman and
                  Simran Arora and
                  Sydney von Arx and
                  Michael S. Bernstein and
                  Jeannette Bohg and
                  Antoine Bosselut and
                  Emma Brunskill and
                  Erik Brynjolfsson and
                  Shyamal Buch and
                  Dallas Card and
                  Rodrigo Castellon and
                  Niladri S. Chatterji and
                  Annie S. Chen and
                  Kathleen Creel and
                  Jared Quincy Davis and
                  Dorottya Demszky and
                  Chris Donahue and
                  Moussa Doumbouya and
                  Esin Durmus and
                  Stefano Ermon and
                  John Etchemendy and
                  Kawin Ethayarajh and
                  Li Fei{-}Fei and
                  Chelsea Finn and
                  Trevor Gale and
                  Lauren Gillespie and
                  Karan Goel and
                  Noah D. Goodman and
                  Shelby Grossman and
                  Neel Guha and
                  Tatsunori Hashimoto and
                  Peter Henderson and
                  John Hewitt and
                  Daniel E. Ho and
                  Jenny Hong and
                  Kyle Hsu and
                  Jing Huang and
                  Thomas Icard and
                  Saahil Jain and
                  Dan Jurafsky and
                  Pratyusha Kalluri and
                  Siddharth Karamcheti and
                  Geoff Keeling and
                  Fereshte Khani and
                  Omar Khattab and
                  Pang Wei Koh and
                  Mark S. Krass and
                  Ranjay Krishna and
                  Rohith Kuditipudi and
                  et al.},
  title        = {On the Opportunities and Risks of Foundation Models},
  journal      = {CoRR},
  volume       = {abs/2108.07258},
  year         = {2021},
  url          = {https://arxiv.org/abs/2108.07258},
  eprinttype    = {arXiv},
  eprint       = {2108.07258},
  bibsource    = {dblp computer science bibliography, https://dblp.org}
}

@article{DBLP:journals/corr/abs-2307-09288,
  author       = {Hugo Touvron and
                  Louis Martin and
                  Kevin Stone and
                  Peter Albert and
                  Amjad Almahairi and
                  Yasmine Babaei and
                  Nikolay Bashlykov and
                  Soumya Batra and
                  Prajjwal Bhargava and
                  Shruti Bhosale and
                  Dan Bikel and
                  Lukas Blecher and
                  Cristian Canton{-}Ferrer and
                  Moya Chen and
                  Guillem Cucurull and
                  David Esiobu and
                  Jude Fernandes and
                  Jeremy Fu and
                  Wenyin Fu and
                  Brian Fuller and
                  Cynthia Gao and
                  Vedanuj Goswami and
                  Naman Goyal and
                  Anthony Hartshorn and
                  Saghar Hosseini and
                  Rui Hou and
                  Hakan Inan and
                  Marcin Kardas and
                  Viktor Kerkez and
                  Madian Khabsa and
                  Isabel Kloumann and
                  Artem Korenev and
                  Punit Singh Koura and
                  Marie{-}Anne Lachaux and
                  Thibaut Lavril and
                  Jenya Lee and
                  Diana Liskovich and
                  Yinghai Lu and
                  Yuning Mao and
                  Xavier Martinet and
                  Todor Mihaylov and
                  Pushkar Mishra and
                  Igor Molybog and
                  Yixin Nie and
                  Andrew Poulton and
                  Jeremy Reizenstein and
                  Rashi Rungta and
                  Kalyan Saladi and
                  Alan Schelten and
                  Ruan Silva and
                  Eric Michael Smith and
                  Ranjan Subramanian and
                  Xiaoqing Ellen Tan and
                  Binh Tang and
                  Ross Taylor and
                  Adina Williams and
                  Jian Xiang Kuan and
                  Puxin Xu and
                  Zheng Yan and
                  Iliyan Zarov and
                  Yuchen Zhang and
                  Angela Fan and
                  Melanie Kambadur and
                  Sharan Narang and
                  Aur{\'{e}}lien Rodriguez and
                  Robert Stojnic and
                  Sergey Edunov and
                  Thomas Scialom},
  title        = {Llama 2: Open Foundation and Fine-Tuned Chat Models},
  journal      = {CoRR},
  volume       = {abs/2307.09288},
  year         = {2023},
  url          = {https://doi.org/10.48550/arXiv.2307.09288},
  doi          = {10.48550/ARXIV.2307.09288},
  eprinttype    = {arXiv},
  eprint       = {2307.09288},
  bibsource    = {dblp computer science bibliography, https://dblp.org}
}

@article{DBLP:journals/corr/abs-2310-06825,
  author       = {Albert Q. Jiang and
                  Alexandre Sablayrolles and
                  Arthur Mensch and
                  Chris Bamford and
                  Devendra Singh Chaplot and
                  Diego de Las Casas and
                  Florian Bressand and
                  Gianna Lengyel and
                  Guillaume Lample and
                  Lucile Saulnier and
                  L{\'{e}}lio Renard Lavaud and
                  Marie{-}Anne Lachaux and
                  Pierre Stock and
                  Teven Le Scao and
                  Thibaut Lavril and
                  Thomas Wang and
                  Timoth{\'{e}}e Lacroix and
                  William El Sayed},
  title        = {Mistral 7B},
  journal      = {CoRR},
  volume       = {abs/2310.06825},
  year         = {2023},
  url          = {https://doi.org/10.48550/arXiv.2310.06825},
  doi          = {10.48550/ARXIV.2310.06825},
  eprinttype    = {arXiv},
  eprint       = {2310.06825},
  bibsource    = {dblp computer science bibliography, https://dblp.org}
}

@article{DBLP:journals/corr/abs-2310-16944,
  author       = {Lewis Tunstall and
                  Edward Beeching and
                  Nathan Lambert and
                  Nazneen Rajani and
                  Kashif Rasul and
                  Younes Belkada and
                  Shengyi Huang and
                  Leandro von Werra and
                  Cl{\'{e}}mentine Fourrier and
                  Nathan Habib and
                  Nathan Sarrazin and
                  Omar Sanseviero and
                  Alexander M. Rush and
                  Thomas Wolf},
  title        = {Zephyr: Direct Distillation of {LM} Alignment},
  journal      = {CoRR},
  volume       = {abs/2310.16944},
  year         = {2023},
  url          = {https://doi.org/10.48550/arXiv.2310.16944},
  doi          = {10.48550/ARXIV.2310.16944},
  eprinttype    = {arXiv},
  eprint       = {2310.16944},
  bibsource    = {dblp computer science bibliography, https://dblp.org}
}

@misc{alignment_handbook2023,
  author = {Lewis Tunstall and Edward Beeching and Nathan Lambert and Nazneen Rajani and Alexander M. Rush and Thomas Wolf},
  title = {The Alignment Handbook},
  year = {2023},
  publisher = {GitHub},
  journal = {GitHub repository},
  howpublished = {\url{https://github.com/huggingface/alignment-handbook}}
}

@article{DBLP:journals/corr/abs-1911-01547,
  author       = {Fran{\c{c}}ois Chollet},
  title        = {On the Measure of Intelligence},
  journal      = {CoRR},
  volume       = {abs/1911.01547},
  year         = {2019},
  url          = {http://arxiv.org/abs/1911.01547},
  eprinttype    = {arXiv},
  eprint       = {1911.01547},
  bibsource    = {dblp computer science bibliography, https://dblp.org}
}

@article{DBLP:journals/corr/abs-2107-03374,
  author       = {Mark Chen and
                  Jerry Tworek and
                  Heewoo Jun and
                  Qiming Yuan and
                  Henrique Pond{\'{e}} de Oliveira Pinto and
                  Jared Kaplan and
                  Harrison Edwards and
                  Yuri Burda and
                  Nicholas Joseph and
                  Greg Brockman and
                  Alex Ray and
                  Raul Puri and
                  Gretchen Krueger and
                  Michael Petrov and
                  Heidy Khlaaf and
                  Girish Sastry and
                  Pamela Mishkin and
                  Brooke Chan and
                  Scott Gray and
                  Nick Ryder and
                  Mikhail Pavlov and
                  Alethea Power and
                  Lukasz Kaiser and
                  Mohammad Bavarian and
                  Clemens Winter and
                  Philippe Tillet and
                  Felipe Petroski Such and
                  Dave Cummings and
                  Matthias Plappert and
                  Fotios Chantzis and
                  Elizabeth Barnes and
                  Ariel Herbert{-}Voss and
                  William Hebgen Guss and
                  Alex Nichol and
                  Alex Paino and
                  Nikolas Tezak and
                  Jie Tang and
                  Igor Babuschkin and
                  Suchir Balaji and
                  Shantanu Jain and
                  William Saunders and
                  Christopher Hesse and
                  Andrew N. Carr and
                  Jan Leike and
                  Joshua Achiam and
                  Vedant Misra and
                  Evan Morikawa and
                  Alec Radford and
                  Matthew Knight and
                  Miles Brundage and
                  Mira Murati and
                  Katie Mayer and
                  Peter Welinder and
                  Bob McGrew and
                  Dario Amodei and
                  Sam McCandlish and
                  Ilya Sutskever and
                  Wojciech Zaremba},
  title        = {Evaluating Large Language Models Trained on Code},
  journal      = {CoRR},
  volume       = {abs/2107.03374},
  year         = {2021},
  url          = {https://arxiv.org/abs/2107.03374},
  eprinttype    = {arXiv},
  eprint       = {2107.03374},
  bibsource    = {dblp computer science bibliography, https://dblp.org}
}

@article{DBLP:journals/corr/abs-2211-12588,
  author       = {Wenhu Chen and
                  Xueguang Ma and
                  Xinyi Wang and
                  William W. Cohen},
  title        = {Program of Thoughts Prompting: Disentangling Computation from Reasoning
                  for Numerical Reasoning Tasks},
  journal      = {CoRR},
  volume       = {abs/2211.12588},
  year         = {2022},
  url          = {https://doi.org/10.48550/arXiv.2211.12588},
  doi          = {10.48550/arXiv.2211.12588},
  eprinttype    = {arXiv},
  eprint       = {2211.12588},
  bibsource    = {dblp computer science bibliography, https://dblp.org}
}

@article{DBLP:journals/corr/abs-2210-12023,
  author       = {Alessandro Stolfo and
                  Zhijing Jin and
                  Kumar Shridhar and
                  Bernhard Sch{\"{o}}lkopf and
                  Mrinmaya Sachan},
  title        = {A Causal Framework to Quantify the Robustness of Mathematical Reasoning
                  with Language Models},
  journal      = {CoRR},
  volume       = {abs/2210.12023},
  year         = {2022},
  url          = {https://doi.org/10.48550/arXiv.2210.12023},
  doi          = {10.48550/arXiv.2210.12023},
  eprinttype    = {arXiv},
  eprint       = {2210.12023},
  bibsource    = {dblp computer science bibliography, https://dblp.org}
}

@article{DBLP:journals/corr/abs-2011-15091,
  author       = {Anirudh Goyal and
                  Yoshua Bengio},
  title        = {Inductive Biases for Deep Learning of Higher-Level Cognition},
  journal      = {CoRR},
  volume       = {abs/2011.15091},
  year         = {2020},
  url          = {https://arxiv.org/abs/2011.15091},
  eprinttype    = {arXiv},
  eprint       = {2011.15091},
  bibsource    = {dblp computer science bibliography, https://dblp.org}
}

@inproceedings{DBLP:conf/nips/TalmorTCGB20,
  author       = {Alon Talmor and
                  Oyvind Tafjord and
                  Peter Clark and
                  Yoav Goldberg and
                  Jonathan Berant},
  editor       = {Hugo Larochelle and
                  Marc'Aurelio Ranzato and
                  Raia Hadsell and
                  Maria{-}Florina Balcan and
                  Hsuan{-}Tien Lin},
  title        = {Leap-Of-Thought: Teaching Pre-Trained Models to Systematically Reason
                  Over Implicit Knowledge},
  booktitle    = {Advances in Neural Information Processing Systems 33: Annual Conference
                  on Neural Information Processing Systems 2020, NeurIPS 2020, December
                  6-12, 2020, virtual},
  year         = {2020},
  url          = {https://proceedings.neurips.cc/paper/2020/hash/e992111e4ab9985366e806733383bd8c-Abstract.html},
  bibsource    = {dblp computer science bibliography, https://dblp.org}
}

@article{DBLP:journals/jmlr/RaffelSRLNMZLL20,
  author       = {Colin Raffel and
                  Noam Shazeer and
                  Adam Roberts and
                  Katherine Lee and
                  Sharan Narang and
                  Michael Matena and
                  Yanqi Zhou and
                  Wei Li and
                  Peter J. Liu},
  title        = {Exploring the Limits of Transfer Learning with a Unified Text-to-Text
                  Transformer},
  journal      = {J. Mach. Learn. Res.},
  volume       = {21},
  pages        = {140:1--140:67},
  year         = {2020},
  url          = {http://jmlr.org/papers/v21/20-074.html},
  bibsource    = {dblp computer science bibliography, https://dblp.org}
}

@inproceedings{DBLP:conf/nips/TirumalaMZA22,
  author       = {Kushal Tirumala and
                  Aram H. Markosyan and
                  Luke Zettlemoyer and
                  Armen Aghajanyan},
  title        = {Memorization Without Overfitting: Analyzing the Training Dynamics
                  of Large Language Models},
  booktitle    = {NeurIPS},
  year         = {2022},
  url          = {http://papers.nips.cc/paper\_files/paper/2022/hash/fa0509f4dab6807e2cb465715bf2d249-Abstract-Conference.html},
  bibsource    = {dblp computer science bibliography, https://dblp.org}
}

@article{lake2015human,
  title={Human-level concept learning through probabilistic program induction},
  author={Lake, Brenden M and Salakhutdinov, Ruslan and Tenenbaum, Joshua B},
  journal={Science},
  volume={350},
  number={6266},
  pages={1332--1338},
  year={2015},
  publisher={American Association for the Advancement of Science}
}

@inproceedings{DBLP:conf/naacl/DevlinCLT19,
  author       = {Jacob Devlin and
                  Ming{-}Wei Chang and
                  Kenton Lee and
                  Kristina Toutanova},
  editor       = {Jill Burstein and
                  Christy Doran and
                  Thamar Solorio},
  title        = {{BERT:} Pre-training of Deep Bidirectional Transformers for Language
                  Understanding},
  booktitle    = {Proceedings of the 2019 Conference of the North American Chapter of
                  the Association for Computational Linguistics: Human Language Technologies,
                  {NAACL-HLT} 2019, Minneapolis, MN, USA, June 2-7, 2019, Volume 1 (Long
                  and Short Papers)},
  pages        = {4171--4186},
  publisher    = {Association for Computational Linguistics},
  year         = {2019},
  url          = {https://doi.org/10.18653/v1/n19-1423},
  doi          = {10.18653/v1/n19-1423},
  bibsource    = {dblp computer science bibliography, https://dblp.org}
}

@inproceedings{DBLP:conf/fat/BenderGMS21,
  author       = {Emily M. Bender and
                  Timnit Gebru and
                  Angelina McMillan{-}Major and
                  Shmargaret Shmitchell},
  editor       = {Madeleine Clare Elish and
                  William Isaac and
                  Richard S. Zemel},
  title        = {On the Dangers of Stochastic Parrots: Can Language Models Be Too Big?},
  booktitle    = {FAccT '21: 2021 {ACM} Conference on Fairness, Accountability, and
                  Transparency, Virtual Event / Toronto, Canada, March 3-10, 2021},
  pages        = {610--623},
  publisher    = {{ACM}},
  year         = {2021},
  url          = {https://doi.org/10.1145/3442188.3445922},
  doi          = {10.1145/3442188.3445922},
  bibsource    = {dblp computer science bibliography, https://dblp.org}
}

@article{DBLP:journals/corr/abs-2303-17651,
  author       = {Aman Madaan and
                  Niket Tandon and
                  Prakhar Gupta and
                  Skyler Hallinan and
                  Luyu Gao and
                  Sarah Wiegreffe and
                  Uri Alon and
                  Nouha Dziri and
                  Shrimai Prabhumoye and
                  Yiming Yang and
                  Sean Welleck and
                  Bodhisattwa Prasad Majumder and
                  Shashank Gupta and
                  Amir Yazdanbakhsh and
                  Peter Clark},
  title        = {Self-Refine: Iterative Refinement with Self-Feedback},
  journal      = {CoRR},
  volume       = {abs/2303.17651},
  year         = {2023},
  url          = {https://doi.org/10.48550/arXiv.2303.17651},
  doi          = {10.48550/arXiv.2303.17651},
  eprinttype    = {arXiv},
  eprint       = {2303.17651},
  bibsource    = {dblp computer science bibliography, https://dblp.org}
}

@article{radford2019language,
  title={Language models are unsupervised multitask learners},
  author={Radford, Alec and Wu, Jeffrey and Child, Rewon and Luan, David and Amodei, Dario and Sutskever, Ilya and others},
  journal={OpenAI blog},
  volume={1},
  number={8},
  pages={9},
  year={2019}
}

@inproceedings{DBLP:conf/nips/BrownMRSKDNSSAA20,
  author       = {Tom B. Brown and
                  Benjamin Mann and
                  Nick Ryder and
                  Melanie Subbiah and
                  Jared Kaplan and
                  Prafulla Dhariwal and
                  Arvind Neelakantan and
                  Pranav Shyam and
                  Girish Sastry and
                  Amanda Askell and
                  Sandhini Agarwal and
                  Ariel Herbert{-}Voss and
                  Gretchen Krueger and
                  Tom Henighan and
                  Rewon Child and
                  Aditya Ramesh and
                  Daniel M. Ziegler and
                  Jeffrey Wu and
                  Clemens Winter and
                  Christopher Hesse and
                  Mark Chen and
                  Eric Sigler and
                  Mateusz Litwin and
                  Scott Gray and
                  Benjamin Chess and
                  Jack Clark and
                  Christopher Berner and
                  Sam McCandlish and
                  Alec Radford and
                  Ilya Sutskever and
                  Dario Amodei},
  editor       = {Hugo Larochelle and
                  Marc'Aurelio Ranzato and
                  Raia Hadsell and
                  Maria{-}Florina Balcan and
                  Hsuan{-}Tien Lin},
  title        = {Language Models are Few-Shot Learners},
  booktitle    = {Advances in Neural Information Processing Systems 33: Annual Conference
                  on Neural Information Processing Systems 2020, NeurIPS 2020, December
                  6-12, 2020, virtual},
  year         = {2020},
  url          = {https://proceedings.neurips.cc/paper/2020/hash/1457c0d6bfcb4967418bfb8ac142f64a-Abstract.html},
  bibsource    = {dblp computer science bibliography, https://dblp.org}
}

@inproceedings{DBLP:conf/nips/Ouyang0JAWMZASR22,
  author       = {Long Ouyang and
                  Jeffrey Wu and
                  Xu Jiang and
                  Diogo Almeida and
                  Carroll L. Wainwright and
                  Pamela Mishkin and
                  Chong Zhang and
                  Sandhini Agarwal and
                  Katarina Slama and
                  Alex Ray and
                  John Schulman and
                  Jacob Hilton and
                  Fraser Kelton and
                  Luke Miller and
                  Maddie Simens and
                  Amanda Askell and
                  Peter Welinder and
                  Paul F. Christiano and
                  Jan Leike and
                  Ryan Lowe},
  title        = {Training language models to follow instructions with human feedback},
  booktitle    = {NeurIPS},
  year         = {2022},
  url          = {http://papers.nips.cc/paper\_files/paper/2022/hash/b1efde53be364a73914f58805a001731-Abstract-Conference.html},
  bibsource    = {dblp computer science bibliography, https://dblp.org}
}

@article{DBLP:journals/corr/abs-2303-08774,
  author       = {OpenAI},
  title        = {{GPT-4} Technical Report},
  journal      = {CoRR},
  volume       = {abs/2303.08774},
  year         = {2023},
  url          = {https://doi.org/10.48550/arXiv.2303.08774},
  doi          = {10.48550/arXiv.2303.08774},
  eprinttype    = {arXiv},
  eprint       = {2303.08774},
  bibsource    = {dblp computer science bibliography, https://dblp.org}
}

@article{DBLP:journals/corr/abs-2302-13971,
  author       = {Hugo Touvron and
                  Thibaut Lavril and
                  Gautier Izacard and
                  Xavier Martinet and
                  Marie{-}Anne Lachaux and
                  Timoth{\'{e}}e Lacroix and
                  Baptiste Rozi{\`{e}}re and
                  Naman Goyal and
                  Eric Hambro and
                  Faisal Azhar and
                  Aur{\'{e}}lien Rodriguez and
                  Armand Joulin and
                  Edouard Grave and
                  Guillaume Lample},
  title        = {LLaMA: Open and Efficient Foundation Language Models},
  journal      = {CoRR},
  volume       = {abs/2302.13971},
  year         = {2023},
  url          = {https://doi.org/10.48550/arXiv.2302.13971},
  doi          = {10.48550/arXiv.2302.13971},
  eprinttype    = {arXiv},
  eprint       = {2302.13971},
  bibsource    = {dblp computer science bibliography, https://dblp.org}
}

@article{DBLP:journals/corr/abs-2212-10560,
  author       = {Yizhong Wang and
                  Yeganeh Kordi and
                  Swaroop Mishra and
                  Alisa Liu and
                  Noah A. Smith and
                  Daniel Khashabi and
                  Hannaneh Hajishirzi},
  title        = {Self-Instruct: Aligning Language Model with Self Generated Instructions},
  journal      = {CoRR},
  volume       = {abs/2212.10560},
  year         = {2022},
  url          = {https://doi.org/10.48550/arXiv.2212.10560},
  doi          = {10.48550/arXiv.2212.10560},
  eprinttype    = {arXiv},
  eprint       = {2212.10560},
  bibsource    = {dblp computer science bibliography, https://dblp.org}
}

@misc{alpaca,
  author = {Rohan Taori and Ishaan Gulrajani and Tianyi Zhang and Yann Dubois and Xuechen Li and Carlos Guestrin and Percy Liang and Tatsunori B. Hashimoto },
  title = {Stanford Alpaca: An Instruction-following LLaMA model},
  year = {2023},
  publisher = {GitHub},
  journal = {GitHub repository},
  howpublished = {\url{https://github.com/tatsu-lab/stanford_alpaca}},
}

@inproceedings{DBLP:conf/iclr/HuSWALWWC22,
  author       = {Edward J. Hu and
                  Yelong Shen and
                  Phillip Wallis and
                  Zeyuan Allen{-}Zhu and
                  Yuanzhi Li and
                  Shean Wang and
                  Lu Wang and
                  Weizhu Chen},
  title        = {LoRA: Low-Rank Adaptation of Large Language Models},
  booktitle    = {The Tenth International Conference on Learning Representations, {ICLR}
                  2022, Virtual Event, April 25-29, 2022},
  publisher    = {OpenReview.net},
  year         = {2022},
  url          = {https://openreview.net/forum?id=nZeVKeeFYf9},
  bibsource    = {dblp computer science bibliography, https://dblp.org}
}

@inproceedings{DBLP:conf/acl/JiaoGSN22,
  author       = {Fangkai Jiao and
                  Yangyang Guo and
                  Xuemeng Song and
                  Liqiang Nie},
  editor       = {Smaranda Muresan and
                  Preslav Nakov and
                  Aline Villavicencio},
  title        = {MERIt: Meta-Path Guided Contrastive Learning for Logical Reasoning},
  booktitle    = {Findings of the Association for Computational Linguistics: {ACL} 2022,
                  Dublin, Ireland, May 22-27, 2022},
  pages        = {3496--3509},
  publisher    = {Association for Computational Linguistics},
  year         = {2022},
  url          = {https://doi.org/10.18653/v1/2022.findings-acl.276},
  doi          = {10.18653/v1/2022.findings-acl.276},
  bibsource    = {dblp computer science bibliography, https://dblp.org}
}

@article{DBLP:journals/corr/abs-1907-11692,
  author       = {Yinhan Liu and
                  Myle Ott and
                  Naman Goyal and
                  Jingfei Du and
                  Mandar Joshi and
                  Danqi Chen and
                  Omer Levy and
                  Mike Lewis and
                  Luke Zettlemoyer and
                  Veselin Stoyanov},
  title        = {RoBERTa: {A} Robustly Optimized {BERT} Pretraining Approach},
  journal      = {CoRR},
  volume       = {abs/1907.11692},
  year         = {2019},
  url          = {http://arxiv.org/abs/1907.11692},
  eprinttype    = {arXiv},
  eprint       = {1907.11692},
  bibsource    = {dblp computer science bibliography, https://dblp.org}
}

@article{DBLP:journals/corr/abs-2206-02336,
  author       = {Yifei Li and
                  Zeqi Lin and
                  Shizhuo Zhang and
                  Qiang Fu and
                  Bei Chen and
                  Jian{-}Guang Lou and
                  Weizhu Chen},
  title        = {On the Advance of Making Language Models Better Reasoners},
  journal      = {CoRR},
  volume       = {abs/2206.02336},
  year         = {2022},
  url          = {https://doi.org/10.48550/arXiv.2206.02336},
  doi          = {10.48550/arXiv.2206.02336},
  eprinttype    = {arXiv},
  eprint       = {2206.02336},
  bibsource    = {dblp computer science bibliography, https://dblp.org}
}

@article{DBLP:journals/corr/abs-2206-10591,
  author       = {Moritz Willig and
                  Matej Zecevic and
                  Devendra Singh Dhami and
                  Kristian Kersting},
  title        = {Can Foundation Models Talk Causality?},
  journal      = {CoRR},
  volume       = {abs/2206.10591},
  year         = {2022},
  url          = {https://doi.org/10.48550/arXiv.2206.10591},
  doi          = {10.48550/arXiv.2206.10591},
  eprinttype    = {arXiv},
  eprint       = {2206.10591},
  bibsource    = {dblp computer science bibliography, https://dblp.org}
}

@article{DBLP:journals/corr/abs-2303-12712,
  author       = {S{\'{e}}bastien Bubeck and
                  Varun Chandrasekaran and
                  Ronen Eldan and
                  Johannes Gehrke and
                  Eric Horvitz and
                  Ece Kamar and
                  Peter Lee and
                  Yin Tat Lee and
                  Yuanzhi Li and
                  Scott M. Lundberg and
                  Harsha Nori and
                  Hamid Palangi and
                  Marco T{\'{u}}lio Ribeiro and
                  Yi Zhang},
  title        = {Sparks of Artificial General Intelligence: Early experiments with
                  {GPT-4}},
  journal      = {CoRR},
  volume       = {abs/2303.12712},
  year         = {2023},
  url          = {https://doi.org/10.48550/arXiv.2303.12712},
  doi          = {10.48550/arXiv.2303.12712},
  eprinttype    = {arXiv},
  eprint       = {2303.12712},
  bibsource    = {dblp computer science bibliography, https://dblp.org}
}

@article{raven1938raven,
  title={Raven standard progressive matrices},
  author={Raven, John C},
  journal={Journal of Cognition and Development},
  year={1938}
}

@article{carpenter1990one,
  title={What one intelligence test measures: a theoretical account of the processing in the Raven Progressive Matrices Test.},
  author={Carpenter, Patricia A and Just, Marcel A and Shell, Peter},
  journal={Psychological review},
  volume={97},
  number={3},
  pages={404},
  year={1990},
  publisher={American Psychological Association}
}

@article{DBLP:journals/corr/abs-2305-00050,
  author       = {Emre Kiciman and
                  Robert Ness and
                  Amit Sharma and
                  Chenhao Tan},
  title        = {Causal Reasoning and Large Language Models: Opening a New Frontier
                  for Causality},
  journal      = {CoRR},
  volume       = {abs/2305.00050},
  year         = {2023},
  url          = {https://doi.org/10.48550/arXiv.2305.00050},
  doi          = {10.48550/arXiv.2305.00050},
  eprinttype    = {arXiv},
  eprint       = {2305.00050},
  bibsource    = {dblp computer science bibliography, https://dblp.org}
}

@inproceedings{DBLP:conf/icml/0001ZSR22,
  author       = {Jiayao Zhang and
                  Hongming Zhang and
                  Weijie J. Su and
                  Dan Roth},
  editor       = {Kamalika Chaudhuri and
                  Stefanie Jegelka and
                  Le Song and
                  Csaba Szepesv{\'{a}}ri and
                  Gang Niu and
                  Sivan Sabato},
  title        = {{ROCK:} Causal Inference Principles for Reasoning about Commonsense
                  Causality},
  booktitle    = {International Conference on Machine Learning, {ICML} 2022, 17-23 July
                  2022, Baltimore, Maryland, {USA}},
  series       = {Proceedings of Machine Learning Research},
  volume       = {162},
  pages        = {26750--26771},
  publisher    = {{PMLR}},
  year         = {2022},
  url          = {https://proceedings.mlr.press/v162/zhang22am.html},
  bibsource    = {dblp computer science bibliography, https://dblp.org}
}

@inproceedings{DBLP:conf/semeval/GordonKR12,
  author       = {Andrew S. Gordon and
                  Zornitsa Kozareva and
                  Melissa Roemmele},
  editor       = {Eneko Agirre and
                  Johan Bos and
                  Mona T. Diab},
  title        = {SemEval-2012 Task 7: Choice of Plausible Alternatives: An Evaluation
                  of Commonsense Causal Reasoning},
  booktitle    = {Proceedings of the 6th International Workshop on Semantic Evaluation,
                  SemEval@NAACL-HLT 2012, Montr{\'{e}}al, Canada, June 7-8, 2012},
  pages        = {394--398},
  publisher    = {The Association for Computer Linguistics},
  year         = {2012},
  url          = {https://aclanthology.org/S12-1052/},
  bibsource    = {dblp computer science bibliography, https://dblp.org}
}

@article{DBLP:journals/corr/abs-2204-02311,
  author       = {Aakanksha Chowdhery and
                  Sharan Narang and
                  Jacob Devlin and
                  Maarten Bosma and
                  Gaurav Mishra and
                  Adam Roberts and
                  Paul Barham and
                  Hyung Won Chung and
                  Charles Sutton and
                  Sebastian Gehrmann and
                  Parker Schuh and
                  Kensen Shi and
                  Sasha Tsvyashchenko and
                  Joshua Maynez and
                  Abhishek Rao and
                  Parker Barnes and
                  Yi Tay and
                  Noam Shazeer and
                  Vinodkumar Prabhakaran and
                  Emily Reif and
                  Nan Du and
                  Ben Hutchinson and
                  Reiner Pope and
                  James Bradbury and
                  Jacob Austin and
                  Michael Isard and
                  Guy Gur{-}Ari and
                  Pengcheng Yin and
                  Toju Duke and
                  Anselm Levskaya and
                  Sanjay Ghemawat and
                  Sunipa Dev and
                  Henryk Michalewski and
                  Xavier Garcia and
                  Vedant Misra and
                  Kevin Robinson and
                  Liam Fedus and
                  Denny Zhou and
                  Daphne Ippolito and
                  David Luan and
                  Hyeontaek Lim and
                  Barret Zoph and
                  Alexander Spiridonov and
                  Ryan Sepassi and
                  David Dohan and
                  Shivani Agrawal and
                  Mark Omernick and
                  Andrew M. Dai and
                  Thanumalayan Sankaranarayana Pillai and
                  Marie Pellat and
                  Aitor Lewkowycz and
                  Erica Moreira and
                  Rewon Child and
                  Oleksandr Polozov and
                  Katherine Lee and
                  Zongwei Zhou and
                  Xuezhi Wang and
                  Brennan Saeta and
                  Mark Diaz and
                  Orhan Firat and
                  Michele Catasta and
                  Jason Wei and
                  Kathy Meier{-}Hellstern and
                  Douglas Eck and
                  Jeff Dean and
                  Slav Petrov and
                  Noah Fiedel},
  title        = {PaLM: Scaling Language Modeling with Pathways},
  journal      = {CoRR},
  volume       = {abs/2204.02311},
  year         = {2022},
  url          = {https://doi.org/10.48550/arXiv.2204.02311},
  doi          = {10.48550/arXiv.2204.02311},
  eprinttype    = {arXiv},
  eprint       = {2204.02311},
  bibsource    = {dblp computer science bibliography, https://dblp.org}
}

@inproceedings{DBLP:conf/ijcai/LiuCLHWZ20,
  author       = {Jian Liu and
                  Leyang Cui and
                  Hanmeng Liu and
                  Dandan Huang and
                  Yile Wang and
                  Yue Zhang},
  editor       = {Christian Bessiere},
  title        = {LogiQA: {A} Challenge Dataset for Machine Reading Comprehension with
                  Logical Reasoning},
  booktitle    = {Proceedings of the Twenty-Ninth International Joint Conference on
                  Artificial Intelligence, {IJCAI} 2020},
  pages        = {3622--3628},
  publisher    = {ijcai.org},
  year         = {2020},
  url          = {https://doi.org/10.24963/ijcai.2020/501},
  doi          = {10.24963/ijcai.2020/501},
  bibsource    = {dblp computer science bibliography, https://dblp.org}
}

@inproceedings{DBLP:conf/icml/VelickovicBBPBD22,
  author       = {Petar Velickovic and
                  Adri{\`{a}} Puigdom{\`{e}}nech Badia and
                  David Budden and
                  Razvan Pascanu and
                  Andrea Banino and
                  Misha Dashevskiy and
                  Raia Hadsell and
                  Charles Blundell},
  editor       = {Kamalika Chaudhuri and
                  Stefanie Jegelka and
                  Le Song and
                  Csaba Szepesv{\'{a}}ri and
                  Gang Niu and
                  Sivan Sabato},
  title        = {The {CLRS} Algorithmic Reasoning Benchmark},
  booktitle    = {International Conference on Machine Learning, {ICML} 2022, 17-23 July
                  2022, Baltimore, Maryland, {USA}},
  series       = {Proceedings of Machine Learning Research},
  volume       = {162},
  pages        = {22084--22102},
  publisher    = {{PMLR}},
  year         = {2022},
  url          = {https://proceedings.mlr.press/v162/velickovic22a.html},
  bibsource    = {dblp computer science bibliography, https://dblp.org}
}

@inproceedings{DBLP:conf/aaai/BoochFHKLLLMMRS21,
  author       = {Grady Booch and
                  Francesco Fabiano and
                  Lior Horesh and
                  Kiran Kate and
                  Jonathan Lenchner and
                  Nick Linck and
                  Andrea Loreggia and
                  Keerthiram Murugesan and
                  Nicholas Mattei and
                  Francesca Rossi and
                  Biplav Srivastava},
  title        = {Thinking Fast and Slow in {AI}},
  booktitle    = {Thirty-Fifth {AAAI} Conference on Artificial Intelligence, {AAAI}
                  2021, Thirty-Third Conference on Innovative Applications of Artificial
                  Intelligence, {IAAI} 2021, The Eleventh Symposium on Educational Advances
                  in Artificial Intelligence, {EAAI} 2021, Virtual Event, February 2-9,
                  2021},
  pages        = {15042--15046},
  publisher    = {{AAAI} Press},
  year         = {2021},
  url          = {https://ojs.aaai.org/index.php/AAAI/article/view/17765},
  bibsource    = {dblp computer science bibliography, https://dblp.org}
}

@article{DBLP:journals/corr/abs-2006-08381,
  author       = {Kevin Ellis and
                  Catherine Wong and
                  Maxwell I. Nye and
                  Mathias Sabl{\'{e}}{-}Meyer and
                  Luc Cary and
                  Lucas Morales and
                  Luke B. Hewitt and
                  Armando Solar{-}Lezama and
                  Joshua B. Tenenbaum},
  title        = {DreamCoder: Growing generalizable, interpretable knowledge with wake-sleep
                  Bayesian program learning},
  journal      = {CoRR},
  volume       = {abs/2006.08381},
  year         = {2020},
  url          = {https://arxiv.org/abs/2006.08381},
  eprinttype    = {arXiv},
  eprint       = {2006.08381},
  bibsource    = {dblp computer science bibliography, https://dblp.org}
}

@article{DBLP:journals/pieee/ScholkopfLBKKGB21,
  author       = {Bernhard Sch{\"{o}}lkopf and
                  Francesco Locatello and
                  Stefan Bauer and
                  Nan Rosemary Ke and
                  Nal Kalchbrenner and
                  Anirudh Goyal and
                  Yoshua Bengio},
  title        = {Toward Causal Representation Learning},
  journal      = {Proc. {IEEE}},
  volume       = {109},
  number       = {5},
  pages        = {612--634},
  year         = {2021},
  url          = {https://doi.org/10.1109/JPROC.2021.3058954},
  doi          = {10.1109/JPROC.2021.3058954},
  bibsource    = {dblp computer science bibliography, https://dblp.org}
}

@inproceedings{DBLP:conf/ijcai/ClarkTR20,
  author       = {Peter Clark and
                  Oyvind Tafjord and
                  Kyle Richardson},
  editor       = {Christian Bessiere},
  title        = {Transformers as Soft Reasoners over Language},
  booktitle    = {Proceedings of the Twenty-Ninth International Joint Conference on
                  Artificial Intelligence, {IJCAI} 2020},
  pages        = {3882--3890},
  publisher    = {ijcai.org},
  year         = {2020},
  url          = {https://doi.org/10.24963/ijcai.2020/537},
  doi          = {10.24963/ijcai.2020/537},
  bibsource    = {dblp computer science bibliography, https://dblp.org}
}

@inproceedings{DBLP:conf/nesy/BaoPHTDWL22,
  author       = {Qiming Bao and
                  Alex Yuxuan Peng and
                  Tim Hartill and
                  Neset Tan and
                  Zhenyun Deng and
                  Michael Witbrock and
                  Jiamou Liu},
  editor       = {Artur S. d'Avila Garcez and
                  Ernesto Jim{\'{e}}nez{-}Ruiz},
  title        = {Multi-Step Deductive Reasoning Over Natural Language: An Empirical
                  Study on Out-of-Distribution Generalisation},
  booktitle    = {Proceedings of the 16th International Workshop on Neural-Symbolic
                  Learning and Reasoning as part of the 2nd International Joint Conference
                  on Learning {\&} Reasoning {(IJCLR} 2022), Cumberland Lodge, Windsor
                  Great Park, UK, September 28-30, 2022},
  series       = {{CEUR} Workshop Proceedings},
  volume       = {3212},
  pages        = {202--217},
  publisher    = {CEUR-WS.org},
  year         = {2022},
  url          = {https://ceur-ws.org/Vol-3212/paper15.pdf},
  bibsource    = {dblp computer science bibliography, https://dblp.org}
}

@article{DBLP:journals/corr/abs-2212-10923,
  author       = {Zonglin Yang and
                  Li Dong and
                  Xinya Du and
                  Hao Cheng and
                  Erik Cambria and
                  Xiaodong Liu and
                  Jianfeng Gao and
                  Furu Wei},
  title        = {Language Models as Inductive Reasoners},
  journal      = {CoRR},
  volume       = {abs/2212.10923},
  year         = {2022},
  url          = {https://doi.org/10.48550/arXiv.2212.10923},
  doi          = {10.48550/arXiv.2212.10923},
  eprinttype    = {arXiv},
  eprint       = {2212.10923},
  bibsource    = {dblp computer science bibliography, https://dblp.org}
}

@inproceedings{DBLP:conf/iclr/YuJDF20,
  author       = {Weihao Yu and
                  Zihang Jiang and
                  Yanfei Dong and
                  Jiashi Feng},
  title        = {ReClor: {A} Reading Comprehension Dataset Requiring Logical Reasoning},
  booktitle    = {8th International Conference on Learning Representations, {ICLR} 2020,
                  Addis Ababa, Ethiopia, April 26-30, 2020},
  publisher    = {OpenReview.net},
  year         = {2020},
  url          = {https://openreview.net/forum?id=HJgJtT4tvB},
  bibsource    = {dblp computer science bibliography, https://dblp.org}
}

@inproceedings{DBLP:conf/acl/WangZTWFJ0D22,
  author       = {Siyuan Wang and
                  Wanjun Zhong and
                  Duyu Tang and
                  Zhongyu Wei and
                  Zhihao Fan and
                  Daxin Jiang and
                  Ming Zhou and
                  Nan Duan},
  editor       = {Smaranda Muresan and
                  Preslav Nakov and
                  Aline Villavicencio},
  title        = {Logic-Driven Context Extension and Data Augmentation for Logical Reasoning
                  of Text},
  booktitle    = {Findings of the Association for Computational Linguistics: {ACL} 2022,
                  Dublin, Ireland, May 22-27, 2022},
  pages        = {1619--1629},
  publisher    = {Association for Computational Linguistics},
  year         = {2022},
  url          = {https://doi.org/10.18653/v1/2022.findings-acl.127},
  doi          = {10.18653/v1/2022.findings-acl.127},
  bibsource    = {dblp computer science bibliography, https://dblp.org}
}

@inproceedings{DBLP:conf/acl/Young0BW22,
  author       = {Nathan Young and
                  Qiming Bao and
                  Joshua Bensemann and
                  Michael Witbrock},
  editor       = {Smaranda Muresan and
                  Preslav Nakov and
                  Aline Villavicencio},
  title        = {AbductionRules: Training Transformers to Explain Unexpected Inputs},
  booktitle    = {Findings of the Association for Computational Linguistics: {ACL} 2022,
                  Dublin, Ireland, May 22-27, 2022},
  pages        = {218--227},
  publisher    = {Association for Computational Linguistics},
  year         = {2022},
  url          = {https://doi.org/10.18653/v1/2022.findings-acl.19},
  doi          = {10.18653/v1/2022.findings-acl.19},
  bibsource    = {dblp computer science bibliography, https://dblp.org}
}

@article{DBLP:journals/corr/abs-2302-00293,
  author       = {Ga{\"{e}}l Gendron and
                  Michael Witbrock and
                  Gillian Dobbie},
  title        = {A Survey of Methods, Challenges and Perspectives in Causality},
  journal      = {CoRR},
  volume       = {abs/2302.00293},
  year         = {2023},
  url          = {https://doi.org/10.48550/arXiv.2302.00293},
  doi          = {10.48550/arXiv.2302.00293},
  eprinttype    = {arXiv},
  eprint       = {2302.00293},
  bibsource    = {dblp computer science bibliography, https://dblp.org}
}

@inproceedings{DBLP:conf/nips/Wei0SBIXCLZ22,
  author       = {Jason Wei and
                  Xuezhi Wang and
                  Dale Schuurmans and
                  Maarten Bosma and
                  Brian Ichter and
                  Fei Xia and
                  Ed H. Chi and
                  Quoc V. Le and
                  Denny Zhou},
  title        = {Chain-of-Thought Prompting Elicits Reasoning in Large Language Models},
  booktitle    = {NeurIPS},
  year         = {2022},
  url          = {http://papers.nips.cc/paper\_files/paper/2022/hash/9d5609613524ecf4f15af0f7b31abca4-Abstract-Conference.html},
  bibsource    = {dblp computer science bibliography, https://dblp.org}
}

@article{DBLP:journals/corr/abs-2305-12599,
  author       = {Qiming Bao and
                  Alex Yuxuan Peng and
                  Zhenyun Deng and
                  Wanjun Zhong and
                  Neset Tan and
                  Nathan Young and
                  Yang Chen and
                  Yonghua Zhu and
                  Michael Witbrock and
                  Jiamou Liu},
  title        = {Contrastive Learning with Logic-driven Data Augmentation for Logical
                  Reasoning over Text},
  journal      = {CoRR},
  volume       = {abs/2305.12599},
  year         = {2023},
  url          = {https://doi.org/10.48550/arXiv.2305.12599},
  doi          = {10.48550/arXiv.2305.12599},
  eprinttype    = {arXiv},
  eprint       = {2305.12599},
  bibsource    = {dblp computer science bibliography, https://dblp.org}
}

@inproceedings{DBLP:conf/pldi/Xu0NH22,
  author       = {Frank F. Xu and
                  Uri Alon and
                  Graham Neubig and
                  Vincent Josua Hellendoorn},
  editor       = {Swarat Chaudhuri and
                  Charles Sutton},
  title        = {A systematic evaluation of large language models of code},
  booktitle    = {MAPS@PLDI 2022: 6th {ACM} {SIGPLAN} International Symposium on Machine
                  Programming, San Diego, CA, USA, 13 June 2022},
  pages        = {1--10},
  publisher    = {{ACM}},
  year         = {2022},
  url          = {https://doi.org/10.1145/3520312.3534862},
  doi          = {10.1145/3520312.3534862},
  bibsource    = {dblp computer science bibliography, https://dblp.org}
}

@article{DBLP:journals/corr/abs-2309-05660,
  author       = {Ruocheng Wang and
                  Eric Zelikman and
                  Gabriel Poesia and
                  Yewen Pu and
                  Nick Haber and
                  Noah D. Goodman},
  title        = {Hypothesis Search: Inductive Reasoning with Language Models},
  journal      = {CoRR},
  volume       = {abs/2309.05660},
  year         = {2023},
  url          = {https://doi.org/10.48550/arXiv.2309.05660},
  doi          = {10.48550/ARXIV.2309.05660},
  eprinttype    = {arXiv},
  eprint       = {2309.05660},
  bibsource    = {dblp computer science bibliography, https://dblp.org}
}

@article{DBLP:journals/corr/abs-2310-08559,
  author       = {Linlu Qiu and
                  Liwei Jiang and
                  Ximing Lu and
                  Melanie Sclar and
                  Valentina Pyatkin and
                  Chandra Bhagavatula and
                  Bailin Wang and
                  Yoon Kim and
                  Yejin Choi and
                  Nouha Dziri and
                  Xiang Ren},
  title        = {Phenomenal Yet Puzzling: Testing Inductive Reasoning Capabilities
                  of Language Models with Hypothesis Refinement},
  journal      = {CoRR},
  volume       = {abs/2310.08559},
  year         = {2023},
  url          = {https://doi.org/10.48550/arXiv.2310.08559},
  doi          = {10.48550/ARXIV.2310.08559},
  eprinttype    = {arXiv},
  eprint       = {2310.08559},
  bibsource    = {dblp computer science bibliography, https://dblp.org}
}

@inproceedings{DBLP:conf/iclr/LoshchilovH19,
  author       = {Ilya Loshchilov and
                  Frank Hutter},
  title        = {Decoupled Weight Decay Regularization},
  booktitle    = {7th International Conference on Learning Representations, {ICLR} 2019,
                  New Orleans, LA, USA, May 6-9, 2019},
  publisher    = {OpenReview.net},
  year         = {2019},
  url          = {https://openreview.net/forum?id=Bkg6RiCqY7},
  bibsource    = {dblp computer science bibliography, https://dblp.org}
}

\appendix

\clearpage

\section{Dataset Details}
\label{sec:details}

This section provides more details and examples of each dataset used in the experiments.

\paragraph{ACRE} We conduct experiments on the Abstract Causal Reasoning (ACRE) dataset \citep{DBLP:conf/cvpr/0017JEZZ21}. ACRE is a Visual Question-Answering (VQA) dataset. In our work, we use a transcription of the dataset into text. Each sample in the data comprises six context images and four test cases. Each context image comprises a set of objects with various shapes, colours and textures, and a light. In the context images, the light can be on or off. The goal of a system is to determine from the context examples if the light is on, off, or if its state cannot be determined in the test cases. To solve this task, the model has to determine for each sample what objects are causally responsible for the activation of the light. We generate two versions of the dataset: in ACRE-Text, each image is replaced by a textual description, and in ACRE-Symbolic, each image is replaced with a vector representation. An example of ACRE-Text is given in Figure \ref{fig:acre_text_ex} and an example of ACRE-Symbolic is given in Figure \ref{fig:acre_symb_ex}.

\begin{figure}[ht]
    \centering
    \small
    \begin{tcolorbox}[title=Pre-Prompt]
    Objects of various color, shape, and texture are displayed. Some objects may contain a device to turn a light on if displayed. From the observations, deduce if the light is on, off, or if the state cannot be determined. Your answer must contain a single word:\\
    on.\\
    off.\\
    undetermined.
    \end{tcolorbox}
    \begin{tcolorbox}[title=Example Cases]
    A cyan cylinder in rubber is visible. The light is on. \\
    A gray cube in rubber is visible. The light is off. \\
    A cyan cylinder in rubber is visible. A gray cube in rubber is visible. The light is on. \\
    A blue cube in metal is visible. The light is off. \\
    A gray cylinder in rubber is visible. A gray cube in metal is visible. The light is off. \\
    A red sphere in metal is visible. A yellow cube in rubber is visible. The light is on.
    \end{tcolorbox}
    \begin{tcolorbox}[title=Test Case]
     A red sphere in metal is visible. The light is \textit{\textcolor{red!50}{undetermined}}
    \end{tcolorbox}
    \caption{Sample from the ACRE-Text dataset. In the test case, the target answer is indicated in \textit{\textcolor{red!50}{italics}}. }
    \label{fig:acre_text_ex}
\end{figure}

\begin{figure}[ht]
    \centering
    \small
    \begin{tcolorbox}[title=Pre-Prompt]
    Figure out the pattern in the following examples and apply it to the test case. Your answer must follow the format of the examples. You can answer 1 if the solution cannot be determined. Your answer must be one of the following choices:\\
    0.\\
    1.\\
    2.
    \end{tcolorbox}
    \begin{tcolorbox}[title=Example Cases]
    {[28]} $\rightarrow$ 2 \\
    {[0]} $\rightarrow$ 0 \\
    {[28, 0]} $\rightarrow$ 2 \\
    {[5]} $\rightarrow$ 0 \\
    {[16, 1]}$\rightarrow$ 0 \\
    {[35, 14]} $\rightarrow$ 2
    \end{tcolorbox}
    \begin{tcolorbox}[title=Test Case]
    [35] $\rightarrow$ \textit{\textcolor{red!50}{1}}
    \end{tcolorbox}
    \caption{Sample from the ACRE-Symbolic dataset. In the test case, the target answer is indicated in \textit{\textcolor{red!50}{italics}}. }
    \label{fig:acre_symb_ex}
\end{figure}

\paragraph{ARC} The second dataset we use is the Abstract Reasoning Challenge (ARC) dataset \citep{DBLP:journals/corr/abs-1911-01547}. The dataset is composed of tasks, each comprising several input and output grids. The goal of the system is to determine the algorithm that converts the input to the output and apply it to a test case. The grids have a variable size comprised between $8 \times 8$ and $30 \times 30$, and contain visual patterns (e.g. recognisable shapes, symmetries). We provide the raw grid to the model as a two-dimensional array of integers. The high dimensionality of the input makes it a challenging task for LLMs. The tasks themselves are also challenging as their transcription in natural language is often complex and supposedly impossible for 12\% of them \citep{DBLP:journals/corr/abs-2106-07824}. An example from the original ARC is given in Figure \ref{fig:arc_ex}.

\begin{figure}
    \centering
    \begin{subfigure}{0.18\textwidth}
        \centering
        \begin{tikzpicture}
            
            \matrix [nodes={draw=gray,minimum size=1.4mm},ampersand replacement=\&] (i1) at (0, 0)
            {
            \node[fill=red!40] {}; \& \node[fill=red!40] {}; \& \node {}; \& \node[fill=red!40] {}; \\
            \node {}; \& \node {}; \& \node[fill=red!40] {}; \& \node[fill=red!40] {}; \\
            \node[fill=red!40] {}; \& \node {}; \& \node[fill=red!40] {}; \& \node[fill=red!40] {}; \\
            };
            
            \matrix [nodes={draw=gray,minimum size=1.4mm},ampersand replacement=\&] (o1)  at (0, -2.5)
            {
            \node[fill=red!40] {}; \& \node {}; \& \node[fill=red!40] {}; \& \node[fill=red!40] {}; \\
            \node {}; \& \node {}; \& \node[fill=red!40] {}; \& \node[fill=red!40] {}; \\
            \node[fill=red!40] {}; \& \node[fill=red!40] {}; \& \node {}; \& \node[fill=red!40] {}; \\
            \node[fill=red!40] {}; \& \node[fill=red!40] {}; \& \node {}; \& \node[fill=red!40] {}; \\
            \node {}; \& \node {}; \& \node[fill=red!40] {}; \& \node[fill=red!40] {}; \\
            \node[fill=red!40] {}; \& \node {}; \& \node[fill=red!40] {}; \& \node[fill=red!40] {}; \\
            };
            
            \draw[->] (i1) -- (o1);
            
        \end{tikzpicture}
        \caption{Example Case 1. }
    \end{subfigure}
    \hfill
    \begin{subfigure}{0.18\textwidth}
        \centering
        \begin{tikzpicture}
            
            \matrix [nodes={draw=gray,minimum size=1.4mm},ampersand replacement=\&] (i1) at (0, 0)
            {
            \node[fill=yellow!40] {}; \& \node[fill=blue!40] {}; \& \node[fill=blue!40] {}; \& \node[fill=yellow!40] {}; \\
            \node[fill=blue!40] {}; \& \node[fill=blue!40] {}; \& \node[fill=blue!40] {}; \& \node[fill=blue!40] {}; \\
            \node[fill=yellow!40] {}; \& \node[fill=yellow!40] {}; \& \node[fill=yellow!40] {}; \& \node[fill=blue!40] {}; \\
            };
            
            \matrix [nodes={draw=gray,minimum size=1.4mm},ampersand replacement=\&] (o1)  at (0, -2.5)
            {
            \node[fill=yellow!40] {}; \& \node[fill=yellow!40] {}; \& \node[fill=yellow!40] {}; \& \node[fill=blue!40] {}; \\
            \node[fill=blue!40] {}; \& \node[fill=blue!40] {}; \& \node[fill=blue!40] {}; \& \node[fill=blue!40] {}; \\
            \node[fill=yellow!40] {}; \& \node[fill=blue!40] {}; \& \node[fill=blue!40] {}; \& \node[fill=yellow!40] {}; \\
            \node[fill=yellow!40] {}; \& \node[fill=blue!40] {}; \& \node[fill=blue!40] {}; \& \node[fill=yellow!40] {}; \\
            \node[fill=blue!40] {}; \& \node[fill=blue!40] {}; \& \node[fill=blue!40] {}; \& \node[fill=blue!40] {}; \\
            \node[fill=yellow!40] {}; \& \node[fill=yellow!40] {}; \& \node[fill=yellow!40] {}; \& \node[fill=blue!40] {}; \\
            };
            
            \draw[->] (i1) -- (o1);
            
        \end{tikzpicture}
        \caption{Example Case 2. }
    \end{subfigure}
    \hfill
    \begin{subfigure}{0.18\textwidth}
        \centering
        \begin{tikzpicture}
            
            \matrix [nodes={draw=gray,minimum size=1.4mm},ampersand replacement=\&] (i1) at (0, 0)
            {
            \node[fill=red!40] {}; \& \node[fill=yellow!40] {}; \& \node[fill=red!40] {}; \& \node[fill=yellow!40] {}; \\
            \node[fill=red!40] {}; \& \node[fill=red!40] {}; \& \node[fill=yellow!40] {}; \& \node[fill=yellow!40] {}; \\
            \node[fill=yellow!40] {}; \& \node[fill=yellow!40] {}; \& \node[fill=yellow!40] {}; \& \node[fill=yellow!40] {}; \\
            };
            
            \matrix [nodes={draw=gray,minimum size=1.4mm},ampersand replacement=\&] (o1)  at (0, -2.5)
            {
            \node[fill=yellow!40] {}; \& \node[fill=yellow!40] {}; \& \node[fill=yellow!40] {}; \& \node[fill=yellow!40] {}; \\
            \node[fill=red!40] {}; \& \node[fill=red!40] {}; \& \node[fill=yellow!40] {}; \& \node[fill=yellow!40] {}; \\
            \node[fill=red!40] {}; \& \node[fill=yellow!40] {}; \& \node[fill=red!40] {}; \& \node[fill=yellow!40] {}; \\
            \node[fill=red!40] {}; \& \node[fill=yellow!40] {}; \& \node[fill=red!40] {}; \& \node[fill=yellow!40] {}; \\
            \node[fill=red!40] {}; \& \node[fill=red!40] {}; \& \node[fill=yellow!40] {}; \& \node[fill=yellow!40] {}; \\
            \node[fill=yellow!40] {}; \& \node[fill=yellow!40] {}; \& \node[fill=yellow!40] {}; \& \node[fill=yellow!40] {}; \\
            };
            
            \draw[->] (i1) -- (o1);
            
        \end{tikzpicture}
        \caption{Example Case 3. }
    \end{subfigure}
    \hfill
    \begin{subfigure}{0.18\textwidth}
        \centering
        \begin{tikzpicture}
            
            \matrix [nodes={draw=gray,minimum size=1.4mm},ampersand replacement=\&] (i1) at (0, 0)
            {
            \node[fill=green!40] {}; \& \node[fill=green!40] {}; \& \node[fill=gray!40] {}; \& \node[fill=gray!40] {}; \\
            \node[fill=green!40] {}; \& \node[fill=gray!40] {}; \& \node[fill=gray!40] {}; \& \node[fill=green!40] {}; \\
            \node[fill=gray!40] {}; \& \node[fill=gray!40] {}; \& \node[fill=green!40] {}; \& \node[fill=green!40] {}; \\
            };
            
            \matrix [nodes={draw=gray,minimum size=1.4mm},ampersand replacement=\&] (o1)  at (0, -2.5)
            {
            \node[fill=gray!40] {}; \& \node[fill=gray!40] {}; \& \node[fill=green!40] {}; \& \node[fill=green!40] {}; \\
            \node[fill=green!40] {}; \& \node[fill=gray!40] {}; \& \node[fill=gray!40] {}; \& \node[fill=green!40] {}; \\
            \node[fill=green!40] {}; \& \node[fill=green!40] {}; \& \node[fill=gray!40] {}; \& \node[fill=gray!40] {}; \\
            \node[fill=green!40] {}; \& \node[fill=green!40] {}; \& \node[fill=gray!40] {}; \& \node[fill=gray!40] {}; \\
            \node[fill=green!40] {}; \& \node[fill=gray!40] {}; \& \node[fill=gray!40] {}; \& \node[fill=green!40] {}; \\
            \node[fill=gray!40] {}; \& \node[fill=gray!40] {}; \& \node[fill=green!40] {}; \& \node[fill=green!40] {}; \\
            };
            
            \draw[->] (i1) -- (o1);
            
        \end{tikzpicture}
        \caption{Example Case 4. }
    \end{subfigure}
    \hfill
    \begin{subfigure}{0.18\textwidth}
        \centering
        \begin{tikzpicture}
            
            \matrix [nodes={draw=gray,minimum size=1.4mm},ampersand replacement=\&] (i1) at (0, 0)
            {
            \node[fill=yellow!40] {}; \& \node[fill=yellow!40] {}; \& \node[fill=red!40] {}; \& \node[fill=red!40] {}; \\
            \node[fill=yellow!40] {}; \& \node[fill=yellow!40] {}; \& \node[fill=yellow!40] {}; \& \node[fill=yellow!40] {}; \\
            \node[fill=yellow!40] {}; \& \node[fill=yellow!40] {}; \& \node[fill=red!40] {}; \& \node[fill=red!40] {}; \\
            };
            
            \matrix [nodes={draw=gray!40,minimum size=1.4mm},ampersand replacement=\&] (o1)  at (0, -2.5)
            {
            \node[fill=yellow!10] {}; \& \node[fill=yellow!10] {}; \& \node[fill=red!10] {}; \& \node[fill=red!10] {}; \\
            \node[fill=yellow!10] {}; \& \node[fill=yellow!10] {}; \& \node[fill=yellow!10] {}; \& \node[fill=yellow!10] {}; \\
            \node[fill=yellow!10] {}; \& \node[fill=yellow!10] {}; \& \node[fill=red!10] {}; \& \node[fill=red!10] {}; \\
            \node[fill=yellow!10] {}; \& \node[fill=yellow!10] {}; \& \node[fill=red!10] {}; \& \node[fill=red!10] {}; \\
            \node[fill=yellow!10] {}; \& \node[fill=yellow!10] {}; \& \node[fill=yellow!10] {}; \& \node[fill=yellow!10] {}; \\
            \node[fill=yellow!10] {}; \& \node[fill=yellow!10] {}; \& \node[fill=red!10] {}; \& \node[fill=red!10] {}; \\
            };
            
            \draw[->] (i1) -- (o1);
            
        \end{tikzpicture}
        \caption{Test Case. }
    \end{subfigure}
    \caption{Sample for the ARC dataset. In our work, each grid is given as a numeric array to the model. In this example, the task consists of generating the symmetric to the input grid and appending it to the input. In the test case, the expected output is lightly coloured. }
    \label{fig:arc_ex}
\end{figure}

\paragraph{BIG-Bench} We select a subset of the BIG-Bench dataset \citep{rule2020child, DBLP:journals/corr/abs-2206-04615} that we name BIG-Bench-F for \textit{Functions}. The subset comprises various tasks represented by a function taking a list as input and returning a new transformed list as output. For each task, several input-output samples are given. In BIG-Bench-F, we give four samples per task by default. The functions include typical list-processing like replacing one list element with another value, selecting a subset of the list, or counting elements. The difficulty in this task is to accurately recognise the function from a few samples. An example is given in Figure \ref{fig:bench_ex_2}.

\begin{figure}[ht]
    \centering
    \small
    \begin{tcolorbox}[sidebyside,sidebyside align = top seam, skin=bicolor,colback=blue!10,colbacklower=green!10, title=Pre-Prompt]
    Apply a function to the final input list to generate the output list. Use any preceding inputs and outputs as examples to find what is the function used. All example outputs have been generated using the same function.
    \tcblower
    Your task is to write down the python function responsible for the transformation of the list in the following examples. The format is [input]  $\rightarrow$ [output]:
    \end{tcolorbox}
    \begin{tcolorbox}[title=Example Cases]
    {[1, 0, 9, 7, 4, 2, 5, 3, 6, 8]} $\rightarrow$ [9, 0, 1, 4, 4, 5] \\
    {[3, 8, 4, 6, 1, 5, 7, 0]} $\rightarrow$ [4, 8, 3, 4, 1, 7] \\
    {[5, 4, 7, 2, 9, 3, 8, 1]} $\rightarrow$ [7, 4, 5, 4, 9, 8] \\
    {[3, 9, 2, 0, 6, 8, 5, 1, 7]} $\rightarrow$ [2, 9, 3, 4, 6, 5]
    \end{tcolorbox}
    \begin{tcolorbox}[sidebyside,sidebyside align = top seam,skin=bicolor,colback=blue!10,colbacklower=green!10,title=Test Case]
    [9, 2, 1, 3, 4, 7, 6, 8, 5, 0] $\rightarrow$ \textit{\textcolor{red!50}{[1, 2, 9, 4, 4, 6] }}
    \tcblower
    Write the function. Next, write a line to print the output of this function for the input [9, 2, 1, 3, 4, 7, 6, 8, 5, 0]
    \end{tcolorbox}
    \caption{Example task in the BIG-Bench-F dataset. For this task, the system must return specific elements of the input list, i.e. [inp[2], inp[1], inp[0], 4, inp[4], inp[6]]. In the test case, the target answer is indicated in \textit{\textcolor{red!50}{italics}}. Text exclusive to base models are indicated by a \textcolor{blue}{blue} background, and text exclusive to code models are indicated by a \textcolor{green}{green} background.  }
    \label{fig:bench_ex_2}
\end{figure}

\paragraph{} The functions used in BIG-Bench are classic list-processing functions. Such functions are likely to be in the training sets of Large Language Models trained on large corpora of data on the internet. Figures \ref{fig:bench_leak_gpt-4_c030} and \ref{fig:bench_leak_gpt-4_c219} illustrate it with two examples. These examples are discussions with GPT-4, where the model is prompted to generate a function solving a list-processing problem and create examples. They show that the model has prior knowledge of the functions needed for the tasks and could solve them by memorising examples from its training set where these functions are applied without the need to reason abstractly.

\begin{figure}[ht]
    \centering
    \small
    \begin{tcolorbox}[title=Pre-Prompt]
    You are GPT-4, a powerful large language model trained by OpenAI. Answer as concisely as possible.
    \end{tcolorbox}
    \begin{tcolorbox}[title=User Query]
    Generate a function that remove the last two elements of a list. Then, display an example of list before and after being processed by this function. Use the format [input\_list] $\rightarrow$ [output\_list]. % BIG-BENCH C030 Task
    \end{tcolorbox}
    \begin{tcolorbox}[colback=blue!10, title=Model Response]
    Here's a Python function that removes the last two elements of a list:
    \begin{verbatim}
```python
def remove_last_two_elements(input_list):
    return input_list[:-2]
```
    \end{verbatim}
    Example: \\
    Input list: [1, 2, 3, 4, 5] \\
    After processing: [1, 2, 3] $\rightarrow$ [1, 2, 3]
    \end{tcolorbox}
    \caption{Example of exchange with GPT-4. When asking to provide a list-processing algorithm, the model is able to provide a correct solution and generates an example with the BIG-Bench-F format, although incorrect. }
    \label{fig:bench_leak_gpt-4_c030}
\end{figure}

\begin{figure}[ht]
    \centering
    \small
    \begin{tcolorbox}[title=Pre-Prompt]
    You are GPT-4, a powerful large language model trained by OpenAI. Answer as concisely as possible.
    \end{tcolorbox}
    \begin{tcolorbox}[title=User Query]
    Generate a function that removes one element every two elements in the list, starting from the end. Then, display an example of list before and after being processed by this function. Use the format [input\_list] $\rightarrow$ [output\_list]. % BIG-BENCH C219 Task
    \end{tcolorbox}
    \begin{tcolorbox}[colback=blue!10, title=Model Response]
    Here's a Python function that removes one element every two elements in the list, starting from the end:
    \begin{verbatim}
```python
def remove_one_every_two_from_end(input_list):
    return input_list[:-1:2] + input_list[-1:] if len(input_list) % 2 == 0 \
        else input_list[::2]
```
    \end{verbatim}
    Example: \\
    Input list: [1, 2, 3, 4, 5, 6] \\
    After processing: [1, 2, 3, 4, 5, 6] → [1, 3, 5]
    \end{tcolorbox}
    \caption{Example of exchange with GPT-4. When asking to provide a list-processing algorithm, the model is able to provide a solution, although incorrect, and generates an example with the BIG-Bench-F format, although also incorrect. }
    \label{fig:bench_leak_gpt-4_c219}
\end{figure}

\paragraph{Evals} We select a subset of the Evals dataset \citep{DBLP:journals/corr/abs-2303-08774} representing logic puzzles.
Evals-P is composed of a set of tasks. For each task, a tuple containing a character and a list of characters is given as an input and a single word from the set \{"foo", "bar"\} is generated from the input according to a logic hidden from the evaluated system. The task consists of finding the logic from eight samples and applying it to a test case. An example is given in Figure \ref{fig:evals_p_ex_2}.
Evals-S is composed of another set of tasks. For each task, a list of integers is given as an input and an output list of words is generated from the input according to a logic hidden from the evaluated system. The task consists of finding the logic from three samples and applying it to a test case. An example is given in Figure \ref{fig:evals_s_ex}.

\begin{figure}[ht]
    \centering
    \small
    \begin{tcolorbox}[title=Pre-Prompt]
    You are a helpful assistant. \\
    Figure out the pattern in the below examples, and then answer with just "foo" or "bar".
    \end{tcolorbox}
    \begin{tcolorbox}[title=Example Cases]
    f, [o, z, a, n, g, e, j, f, i, c, l, u, b] $\rightarrow$ foo\\
    l, [v, u, f, b, m, y, j, h, n, c, d, a, p] $\rightarrow$ bar\\
    p, [c, e, s, h, q, o, a, t, k, d, n, l, z] $\rightarrow$ bar\\
    p, [c, h, m, z, d, v, k, l, j, e, x, p, n] $\rightarrow$ foo\\
    y, [e, u, v, z, c, q, s, a, f, o, l, p, d] $\rightarrow$ bar\\
    u, [x, e, n, t, v, o, g, c, d, y, r, j, l] $\rightarrow$ bar\\
    m, [l, n, k, e, h, i, c, v, r, j, a, y, o] $\rightarrow$ bar\\
    v, [j, g, q, t, x, y, m, z, b, h, p, u, r] $\rightarrow$ bar
    \end{tcolorbox}
    \begin{tcolorbox}[title=Test Case]
    u, [d, a, x, i, h, v, e, z, r, c, n, y, o] $\rightarrow$ \textit{\textcolor{red!50}{bar}}
    \end{tcolorbox}
    \caption{Example task in the Evals-P dataset. For this task, the system must return "foo" if the first character of the input is in the list or "bar" otherwise. In the test case, the target answer is indicated in \textit{\textcolor{red!50}{italics}}. }
    \label{fig:evals_p_ex_2}
\end{figure}

\begin{figure}
    \centering
    \small
    \begin{tcolorbox}[title=Pre-Prompt]
    You are a pattern recognition bot, figure out the pattern and reply with just the solution, ensure that your reply starts with your solution.
    \end{tcolorbox}
    \begin{tcolorbox}[title=Example Cases]
        13, 17, 1, 6 $\rightarrow$ Brown,White,Purple,Blue\\
        1, 9, 6, 11 $\rightarrow$ Purple,Brown,Blue,White\\
        13, 2, 17, 10 $\rightarrow$ Brown,Purple,White,Blue
    \end{tcolorbox}
    \begin{tcolorbox}[title=Test Case]
    5, 9, 2, 11 $\rightarrow$ \textit{\textcolor{red!50}{Blue,Brown,Purple,White}}
    \end{tcolorbox}
    \caption{Example of task in the Evals-S dataset. For this task, the system must sort the words according to the numbers in input (e.g. word "white" is located at the index of the highest integer and word "purple" is located at the index of the lowest integer). In the test case, the target answer is indicated in \textit{\textcolor{red!50}{italics}}. }
    \label{fig:evals_s_ex}
\end{figure}

\paragraph{PVR} The Pointer-Value Retrieval (PVR) dataset \citep{DBLP:journals/corr/abs-2107-12580} is a dataset for retrieval tasks. Tasks involve selecting one or several values in a list and applying a function on this subset. For each task, the system must recognise the retrieval and application functions and apply them to a test case. Samples in the datasets are composed of a pointer-values pair and a label. The values are stored in an array, and the pointer is an integer pointing to an index in the array. The pointer indicates the subset of values to consider for the task. An example is given in Figure \ref{fig:pvr_ex}.

\begin{figure}
    \centering
    \small
    \begin{tcolorbox}[sidebyside,sidebyside align = top seam, skin=bicolor,colback=blue!10,colbacklower=green!10, title=Pre-Prompt]
    Figure out the pattern in the following examples and apply it to the test case. Your answer must follow the format of the examples.
    \tcblower
    Your task is to write down the python function responsible for the computation of the output from the list in the following examples. Your answer must follow the format of the examples.
    \end{tcolorbox}
    \begin{tcolorbox}[title=Example Cases]
        {[5, 7, 4, 1, 8, 9, 8, 1, 9, 8, 4]} $\rightarrow$ 8 \\
        {[4, 0, 0, 7, 0, 1, 0, 5, 3, 0, 0]} $\rightarrow$ 1 \\
        {[0, 2, 8, 2, 5, 9, 4, 3, 8, 5, 4]} $\rightarrow$ 2 \\
        {[3, 3, 2, 6, 5, 7, 4, 6, 7, 4, 8]} $\rightarrow$ 5
    \end{tcolorbox}
    \begin{tcolorbox}[sidebyside,sidebyside align = top seam,skin=bicolor,colback=blue!10,colbacklower=green!10,title=Test Case]
    {[3, 4, 9, 7, 1, 8, 7, 1, 0, 3, 5]} $\rightarrow$ \textit{\textcolor{red!50}{1}}
    \tcblower
    Write the function. Next, write a line to print the output of this function for the input {[3, 4, 9, 7, 1, 8, 7, 1, 0, 3, 5]}
    \end{tcolorbox}
    \caption{Example of task in the PVR dataset. In the test case, the target answer is indicated in \textit{\textcolor{red!50}{italics}}. Text exclusive to base models are indicated by a \textcolor{blue}{blue} background, and text exclusive to code models are indicated by a \textcolor{green}{green} background. }
    \label{fig:pvr_ex}
\end{figure}

\paragraph{RAVEN} RAVEN \citep{DBLP:conf/cvpr/ZhangGJZZ19} is a VQA dataset composed of sequences of images to complete. The images contain Raven matrices \citep{raven1938raven}, i.e. geometric shapes (e.g. square, circle, pentagon) assembled in various ways (e.g. one shape inside another, four shapes in a $4 \times 4$ grid). RAVEN is a dataset similar to Procedurally Generated Matrices (PGM) \citep{DBLP:conf/icml/SantoroHBML18} but has the advantage of providing a tree structure describing the semantics of each matrix.  We focus on a subset where a single shape appears in the image. The task is, given a sequence of eight images and eight possible choices, to pick the correct image that follows in the sequence. As RAVEN is a visual dataset like ACRE, we generate a text description of each image from their semantic tree that we will feed into the evaluated models. We create two sets: RAVEN-Text contains descriptions in natural language, and RAVEN-Symbolic contains symbolic descriptions. We also build another version of the dataset where choices are hidden. We name the former RAVEN-mcqa and the latter RAVEN-opqa. Examples for each are given in Figures \ref{fig:raven_text_ex} and \ref{fig:raven_symb_ex}, respectively.

\begin{figure}[ht]
    \centering
    \small
    \begin{tcolorbox}[sidebyside,sidebyside align = top seam, skin=bicolor,colback=blue!10,colbacklower=green!10, title=Pre-Prompt]
    Find the pattern number 9 that completes the sequence. Write the correct pattern with the same format as in the examples. Patterns in the sequence are preceded by a number from 1 to 8.
    \tcblower
    Find the pattern number 9 that completes the sequence. Pick the letter in front of the correct pattern that logically follows in the sequence from the answer set. Patterns in the sequence are preceded by a number from 1 to 8. Patterns in the answer set are preceded by a letter from A to H. Only return the letter in front of the correct pattern.
    \end{tcolorbox}
    \begin{tcolorbox}[skin=bicolor,colbacklower=green!10, title=Example Cases]
    1. On an image, a large lime square rotated at 180 degrees. \\
    2. On an image, a medium lime square rotated at 180 degrees. \\
    3. On an image, a huge lime square rotated at 180 degrees. \\
    4. On an image, a huge yellow circle rotated at 0 degrees. \\
    5. On an image, a large yellow circle rotated at 0 degrees. \\
    6. On an image, a medium yellow circle rotated at 0 degrees. \\
    7. On an image, a medium white hexagon rotated at -90 degrees. \\
    8. On an image, a huge white hexagon rotated at -90 degrees. 
    \tcblower
    A. On an image, a tiny white hexagon rotated at -90 degrees. \\
    B. On an image, a giant white hexagon rotated at -90 degrees. \\
    C. On an image, a large red hexagon rotated at -90 degrees. \\
    D. On an image, a large orange hexagon rotated at -90 degrees. \\
    E. On an image, a large white hexagon rotated at -90 degrees. \\
    F. On an image, a large green hexagon rotated at -90 degrees. \\
    G. On an image, a large blue hexagon rotated at -90 degrees. \\
    H. On an image, a large yellow hexagon rotated at -90 degrees.
    \end{tcolorbox}
    \begin{tcolorbox}[sidebyside,sidebyside align = top seam, skin=bicolor,colback=blue!10,colbacklower=green!10, title=Test Case]
    The pattern that logically follows is:\\9. \textit{\textcolor{red!50}{On an image, a large white hexagon rotated at -90 degrees.}}
    \tcblower
    The answer is \textit{\textcolor{red!50}{E}}
    \end{tcolorbox}
    \caption{Sample from the RAVEN$^T$-Text dataset. In the test case, the target answer is indicated in \textit{\textcolor{red!50}{italics}}. Text exclusive to RAVEN$^T$-opqa has a \textcolor{blue}{blue} background, and text exclusive to RAVEN$^T$-mcqa has a \textcolor{green}{green} background. Shared text has a \textcolor{gray}{gray} background. }
    \label{fig:raven_text_ex}
\end{figure}

\begin{figure}[ht]
    \centering
    \small
    \begin{tcolorbox}[sidebyside,sidebyside align = top seam, skin=bicolor,colback=blue!10,colbacklower=green!10, title=Pre-Prompt]
    Find the pattern number 9 that completes the sequence. Write the correct pattern with the same format as in the examples. Patterns in the sequence are preceded by a number from 1 to 8.
    \tcblower
    Find the pattern number 9 that completes the sequence. Pick the letter in front of the correct pattern that logically follows in the sequence from the answer set. Patterns in the sequence are preceded by a number from 1 to 8. Patterns in the answer set are preceded by a letter from A to H. Only return the letter in front of the correct pattern.
    \end{tcolorbox}
    \begin{tcolorbox}[skin=bicolor,colbacklower=green!10, title=Example Cases]
    1. [(D, D, C, H,)] \\
    2. [(C, D, C, H,)] \\
    3. [(E, D, C, H,)] \\
    4. [(E, C, F, D,)] \\
    5. [(D, C, F, D,)] \\
    6. [(C, C, F, D,)] \\
    7. [(C, J, E, B,)] \\
    8. [(E, J, E, B,)] 
    \tcblower
    A. [(A, J, E, B,)] \\
    B. [(F, J, E, B,)] \\
    C. [(D, A, E, B,)] \\
    D. [(D, B, E, B,)] \\
    E. [(D, J, E, B,)] \\
    F. [(D, E, E, B,)] \\
    G. [(D, G, E, B,)] \\
    H. [(D, C, E, B,)]
    \end{tcolorbox}
    \begin{tcolorbox}[sidebyside,sidebyside align = top seam, skin=bicolor,colback=blue!10,colbacklower=green!10, title=Test Case]
    The pattern that logically follows is:\\9. \textit{\textcolor{red!50}{[(D, J, E, B,)]}}
    \tcblower
    The answer is \textit{\textcolor{red!50}{E}}
    \end{tcolorbox}
    \caption{Sample from the RAVEN$^T$-Symbolic dataset. In the test case, the target answer is indicated in \textit{\textcolor{red!50}{italics}}. Text exclusive to RAVEN$^T$-opqa has a \textcolor{blue}{blue} background, and text exclusive to RAVEN$^T$-mcqa has a \textcolor{green}{green} background. Shared text has a \textcolor{gray}{gray} background. }
    \label{fig:raven_symb_ex}
\end{figure}

\paragraph{} Raven matrices are a standard and long-existing test \citep{raven1938raven, carpenter1990one}, likely in the training sets of Large Language Models trained on large corpora of data on the internet. To figure it out, we directly prompt GPT-3.5-Turbo and GPT-4. The discussions are represented in Figures \ref{fig:raven_leak_gpt-3-5} and \ref{fig:raven_leak_gpt-4}. The responses of the model indicate knowledge of the RAVEN test, although GPT-4 generates a correct sample of a Raven test, whereas GPT-3.5-Turbo generates an example that does not have a valid logic. The main takeaway from these examples is that Raven test data has been leaked to the training sets of those models. However, whether exact examples from the test set are also part of the training data is unknown.

\begin{figure}[ht]
    \centering
    \small
    \begin{tcolorbox}[title=Pre-Prompt]
    You are GPT-4, a powerful large language model trained by OpenAI. Answer as concisely as possible.
    \end{tcolorbox}
    \begin{tcolorbox}[title=User Query]
    What are Raven matrices?
    \end{tcolorbox}
    \begin{tcolorbox}[colback=blue!10, title=Model Response]
    Raven matrices are a type of nonverbal intelligence test that require test-takers to identify patterns in a series of abstract designs.
    \end{tcolorbox}
    \begin{tcolorbox}[title=User Query]
    Give me an example of Raven test in plain text, give the visual patterns, the possible options and indicate the correct one. Describe each matrix with natural language. Use semantic descriptors, do not attempt to directly represent the matrices.
    \end{tcolorbox}
    \begin{tcolorbox}[colback=blue!10, title=Model Response]
    Sure, here's an example of a Raven test matrix: \\
    The matrix consists of a 3x3 grid of abstract shapes. In the top row, there are two shapes: a circle and a triangle, both filled with diagonal lines. In the middle row, there is a single shape: a square filled with vertical lines. In the bottom row, there are two shapes: a diamond and a pentagon, both filled with horizontal lines. \\
    There are 8 possible options to complete the matrix. Option A has a square filled with diagonal lines in the bottom right corner. Option B has a diamond filled with diagonal lines in the bottom right corner. Option C has a pentagon filled with diagonal lines in the bottom right corner. Option D has a square filled with horizontal lines in the bottom right corner. Option E has a diamond filled with horizontal lines in the bottom right corner. Option F has a pentagon filled with horizontal lines in the bottom right corner. Option G has a circle filled with horizontal lines in the bottom right corner. Option H has a triangle filled with horizontal lines in the bottom right corner. \\
    The correct answer is Option C, which completes the pattern of shapes in the matrix. The pattern is that the top row contains two shapes with diagonal lines, the middle row contains one shape with vertical lines, and the bottom row should contain two shapes with horizontal lines. The diamond and pentagon shapes in the bottom row both have five sides, making Option C the correct choice.
    \end{tcolorbox}
    \caption{Example of exchange with GPT-3.5-Turbo. When queried about Raven matrices, the model is able to provide a description of the test and provide an \textit{incorrect} example in natural language. }
    \label{fig:raven_leak_gpt-4}
\end{figure}

\begin{figure}[ht]
    \centering
    \small
    \begin{tcolorbox}[title=Pre-Prompt]
    You are GPT-4, a powerful large language model trained by OpenAI. Answer as concisely as possible.
    \end{tcolorbox}
    \begin{tcolorbox}[title=User Query]
    What are Raven matrices?
    \end{tcolorbox}
    \begin{tcolorbox}[colback=blue!10, title=Model Response]
    Raven matrices, or Raven's Progressive Matrices, are non-verbal intelligence tests designed to measure abstract reasoning and problem-solving skills. They consist of visual patterns with a missing piece, and the test-taker must select the correct option to complete the pattern logically.
    \end{tcolorbox}
    \begin{tcolorbox}[title=User Query]
    Give me an example of Raven test in plain text, give the visual patterns, the possible options and indicate the correct one. Describe each matrix with natural language. Use semantic descriptors, do not attempt to directly represent the matrices.
    \end{tcolorbox}
    \begin{tcolorbox}[colback=blue!10, title=Model Response]
    Matrix: A 3x3 grid with geometric shapes. \\
    Row 1: \\
    1. Square with a small circle inside \\
    2. Square with a small triangle inside \\
    3. Square with a small square inside \\
    Row 2: \\
    4. Circle with a small circle inside \\
    5. Circle with a small triangle inside \\
    6. Circle with a small square inside \\
    Row 3: \\
    7. Triangle with a small circle inside \\
    8. Triangle with a small triangle inside \\
    9. Missing piece \\
    Options: \\
    A. Triangle with a small square inside \\
    B. Triangle with a small hexagon inside \\
    C. Triangle with a small star inside \\
    Correct Option: A \\
    The pattern is that each row has a different primary shape (square, circle, triangle) and each column has a different secondary shape (circle, triangle, square) inside the primary shape. The missing piece should be a triangle with a small square inside to complete the pattern.
    \end{tcolorbox}
    \caption{Example of exchange with GPT-4. When queried about Raven matrices, the model is able to provide a description of the test and provide a \textit{correct} example in natural language. }
    \label{fig:raven_leak_gpt-3-5}
\end{figure}

\clearpage 
\section{Additional Experiments}

We perform additional experiments using other models and prompting methods. The settings are the same as in the main paper.

For Text-Davinci-3, GPT-3.5-Turbo, and GPT-4, we use the Open AI API to run all the evaluations. Text-Davinci is a text-completion model, so we convert our input context and question to a single string. GPT-3.5-Turbo and GPT-4 are chat completion models, so we provide the instructions in chat format. The pre-prompt and examples are given to the model by the system, and the supposed user gives the question. We use a temperature of 0.5 for the output generation and the default parameters of each model for the maximum number of generated tokens. Unless specified otherwise, the version of GPT-3.5-Turbo is gpt-3.5-turbo-0301 and the version of GPT-4 is gpt-4-0314.
For the open models, we use the weights provided on the Huggingface hub. RoBERTa-large and MERIt are used as MCQA models, while the others are used as causal language modelling models. We set the maximum number of generated tokens to 128 for the default models, 512 for \textit{chain-of-thought}-prompted models (see Appendix \ref{sec:cot}), and 256 for the code models (see Appendix \ref{sec:code_plus}). We evaluate each model with its default configuration. 
As the language models generate free-text answers, we need to extract the answers using regular expression patterns. We consider a model to provide a valid answer even if the format is incorrect (e.g. if they accompany their answer with additional text although we ask only for the answer). Unless specified otherwise, we always ask the model to provide a single answer and return only the aforementioned answer without explanation. We perform a single evaluation per dataset per model as the cost of running some of the Large Language Models makes it prohibitively expensive to systematically perform multiple runs.

\subsection{Chain-of-Thought Prompting}

We perform a series of experiments with \textit{Chain-of-Thought} prompting \citep{DBLP:conf/nips/Wei0SBIXCLZ22}. To elicit multi-step reasoning, we use the following pre-prompt: \textit{"Figure out the pattern in the following examples and apply it to the test case. Describe every step of your reasoning before proposing a solution. When giving the solution, start your sentence with 'ANSWER:' "}. Appendix \ref{sec:pvr_ex} gives several examples illustrating this principle. We perform experiments with GPT-3.5-turbo, GPT-4, and Alpaca-LoRA. Our experiments with \textit{Chain-of-Thought} have the suffix \textit{model}-cot. Our results on BIG-Bench-F, Evals-S, and PVR datasets are presented in Table \ref{tab:open_qa_results_cot}.

\begin{table}[ht]
    \centering
    \caption{Accuracy of Large Language Models on Open QA datasets when prompted using \textit{Chain-of-Thought}. Datasets are represented in columns, and models in rows. The best result for each dataset is indicated in \textbf{bold}, and the second best is indicated in \textit{italics}. }
    \begin{tabular}{lccccc}
        \hline
         & \multirow{2}*{BIG-Bench-F} & \multirow{2}*{Evals-S} & \multirow{2}*{PVR} & \multicolumn{2}{c}{RAVEN$^T$-opqa} \\
        \cline{5-6}
         & & & & Text & Symb \\
        \hline
        GPT-3.5-Turbo & \textit{0.153} & \textit{0.186} & 0.124 & \textit{0.226} & \textit{0.161} \\
        GPT-4 & \textbf{0.514} & \textbf{0.304} & \textbf{0.177} & \textbf{0.410} & \textbf{0.330} \\
        Alpaca-LoRA & 0.144 & 0.000 & \textit{0.152} & 0.000 & 0.067 \\
        \hline
        GPT-3.5-Turbo-cot & \textit{0.097} & \textit{0.130} & \textbf{0.210} & \textit{0.302} & \textit{0.211} \\
        GPT-4-cot & \textbf{0.476} & \textbf{0.148} & \textit{0.174} & \textbf{0.385} & \textbf{0.354} \\
        Alpaca-LoRA-cot & 0.084 & 0.029 & 0.152 & 0.000 & 0.069 \\
        \hline
    \end{tabular}
    \label{tab:open_qa_results_cot}
\end{table}

\begin{table}[ht]
    \centering
    \caption{Accuracy of Large Language Models on Multiple-Choice QA datasets when prompted using \textit{Chain-of-Thought}. Datasets are represented in columns, and models in rows. The best result for each dataset is indicated in \textbf{bold}, and the second best is indicated in \textit{italics}. }
    \begin{tabular}{lcccc}
        \hline
         & \multicolumn{2}{c}{ACRE$^T$} & \multicolumn{2}{c}{RAVEN$^T$-mcqa} \\
        \cline{2-3}\cline{4-5}
         & Text & Symb & Text & Symb \\
        \hline
        GPT-3.5-Turbo & \textit{0.184} & \textit{0.445} & \textit{0.276} & \textit{0.315} \\
        GPT-4 & \textbf{0.272} & \textbf{0.512} &\textbf{0.697} & \textbf{0.535} \\
        Alpaca-LoRA & 0.015 & 0.123 & 0.082 & 0.124 \\
        \hline
        GPT-3.5-Turbo-cot &  \textbf{0.255} & \textit{0.345} & \textit{0.257} &\textit{ 0.144} \\
        GPT-4-cot & \textit{0.214} & \textbf{0.394} & \textbf{0.596} & \textbf{0.517} \\
        Alpaca-LoRA-cot & \textit{0.059} & 0.114 & 0.000 & 0.114  \\
        \hline
        random & 0.33 & 0.33 & 0.125 & 0.125 \\
        \hline
    \end{tabular}
    \label{tab:mc_qa_results_cot}
\end{table}

Overall, the results obtained using \textit{Chain-of-Thought} prompting are not higher than those obtained with the base models. On The BIG-Bench-F dataset, the \textit{Chain-of-Thought} versions achieve systematically lower performance than their base counterparts, although no important drop of performance is observed. On Evals-S, the performances of GPT-3.5 and GPT-4 are also reduced. The accuracy of base GPT-4 is higher than base GPT-3.5 by a fair margin, but this margin is highly reduced in the \textit{Chain-of-Thought} version. On PVR, while the accuracy for GPT-4 and Alpaca-LoRA remain unchanged or slightly reduced, the performance of GPT-3.5-Turbo is increased.

\label{sec:cot}

\subsection{Refinement}

In this section, we investigate various refinement and filtering strategies that have been successful in improving LLM reasoning abilities and see if they can be used to improve abstract reasoning performance. We study two types of strategies: code\textit{-based} and \textit{self-based}. Code-based strategies ask the model to provide a code answer and an interpreter is used to evaluate the quality of the program. Self-based strategies ask the model to provide a plain-text answer and prompt a separate instance of the model to evaluate the quality of the response. 

\textit{Code-filtering} is a code-based strategy that consists in generating multiple code responses and filtering out the programs that cannot solve the example cases. \textit{Code-refinement} \citep{DBLP:journals/corr/abs-2309-05660, DBLP:journals/corr/abs-2310-08559} is an iterative process where the model generates a first program. The program is run on the context examples and, if not all answers are correct, the model is prompted to correct its answer based on the output of the interpreter. \textit{Self-filtering} and \textit{self-refinement} \citep{DBLP:journals/corr/abs-2310-08559, DBLP:journals/corr/abs-2303-17651} are similar self-based techniques. They ask the LLM to assess whether the given answer is correct rather than relying on an interpreter. We conduct experiments on BIG-Bench-F and PVR using GPT-3.5 and GPT-4. We use the latest versions of GPT-3.5-Turbo (gpt-3.5-turbo-0613) and GPT-4-Turbo (gpt-4-1106-preview).

Table \ref{tab:bench_refinement} shows the main results. Overall, the improvements brought by the refinement strategies are limited. In particular, self-refinement is detrimental to both GPT-3.5 and GPT-4. The bottleneck in the reasoning is the recognition of the abstract rule linking the context examples. Therefore, the LLM cannot be a good evaluator. This is consistent with the MCQA results observed in Section \ref{sec:mcqa} where the LLMs fail to discriminate the good answers. Unlike self-refinement, self-filtering generates multiple answers independently, not conditioned on the previous iterations. As the LLM performance as a discriminator is above chance, the filtering process can help improving the performance. Code-refinement provides slight improvements in the accuracy for BIG-Bench but decreases it for PVR. The LLMs struggle to accurately exploit the feedback from the interpreter. On BIG-Bench, code-filtering improves the performance the most. The reasons are similar to the self-filtering strategy although the code interpreter is a more rigorous discriminator.

\begin{table}[t]
    \centering
    \caption{Accuracy of refined Large Language Models on BIG-Bench-F and PVR datasets. The best result for each dataset is indicated in \textbf{bold}. Experiments are performed with the latest version of GPT-3.5 (\textit{gpt-3.5-turbo-0613}) and GPT-4 (\textit{gpt-4-1106-preview}). }
    \begin{tabular}{lcc}
        \hline
         & BIG-Bench-F & PVR \\
        \hline
        GPT-4-Turbo-code & 0.280 &\textbf{0.152} \\
        GPT-4-Turbo-code-filtering & \textbf{0.400} & \textbf{0.152} \\
        GPT-4-Turbo-code-refinement & 0.296 & 0.144 \\
        GPT-4-Turbo & 0.268 & 0.000 \\
        GPT-4-Turbo-self-filtering & 0.284 & 0.004 \\
        GPT-4-Turbo-self-refinement & 0.252 & 0.000 \\
        \hline
        GPT-3.5-Turbo-code & 0.316 & \textbf{0.200} \\
        GPT-3.5-Turbo-code-filtering & 0.352 & \textbf{0.200} \\
        GPT-3.5-Turbo-code-refinement & 0.336 & 0.188 \\
        GPT-3.5-Turbo & 0.416 & 0.116 \\
        GPT-3.5-Turbo-self-filtering & \textbf{0.444} & 0.124 \\
        GPT-3.5-Turbo-self-refinement & 0.323 & 0.084 \\
        \hline
    \end{tabular}
    \label{tab:bench_refinement}
\end{table}

We conduct additional experiments where we vary the number of refinement steps or answer generations. The results are shown in Table \ref{tab:bench_refinement_steps}. For the refinement strategies, we show the number of times the LLM is tasked to evaluate the answer and re-generate it. For the filtering strategies, we show the number of independent generations made by the LLM. For cost reasons, we perform our experiments with GPT-3.5 only. Self-refinement achieves its best performance with 2 steps, it then declines as the number of steps increases. As LLMs are not good discriminators, errors accumulate as steps increase. Filtering methods tend to get a higher performance as the number of generations increase. This is expected as the model gets more chances to find a suitable answer. The only exception is self-filtering on BIG-Bench, where the best performance is achieved with 4 steps. Finally, increasing the number of steps helps the code-refinement strategy. Nevertheless, the improvements brought are limited.

\begin{table}[t]
    \centering
    \caption{Variation of the accuracy of refined GPT-3.5-Turbo on BIG-Bench-F and PVR datasets when increasing the refinement steps. The step yielding the best result for each dataset and model is indicated in \textbf{bold}. Experiments performed with the latest version of GPT-3.5 (\textit{gpt-3.5-turbo-0613}). }
    \begin{tabular}{lccccccc}
        \hline
         & \multicolumn{3}{c}{BIG-Bench-F} & & \multicolumn{3}{c}{PVR} \\
        \cline{2-4}\cline{6-8}
         & 2 steps & 4 steps & 8 steps & & 2 steps & 4 steps & 8 steps \\
        \hline
        GPT-3.5-Turbo-code-filtering & 0.320 & 0.352 & \textbf{0.380} & & 0.200 & 0.200 & \textbf{0.208} \\
        GPT-3.5-Turbo-code-refinement & 0.320 &\textbf{ 0.336} & 0.335 & & 0.188 & 0.188 & \textbf{0.201} \\
        GPT-3.5-Turbo-self-filtering & 0.428 & \textbf{0.444} & 0.424 & & 0.132 & 0.124 & \textbf{0.148} \\
        GPT-3.5-Turbo-self-refinement & \textbf{0.364} & 0.323 & 0.307 & & \textbf{0.112} & 0.084 & 0.080 \\
        \hline
    \end{tabular}
    \label{tab:bench_refinement_steps}
\end{table}
\label{sec:refinement}

\subsection{Code Generation}

To study the relationship between code output and accuracy more closely, we compare the proportion of valid generated programs (i.e. functions that compile) with the proportion of programs generating the correct answer. We summarise the result for PVR in Figure \ref{fig:code_prop_valid}. We observe that models can almost systematically generate a code able to compile and produce an answer. We deduce that the production of a program with a valid syntax is not a bottleneck for the performance. The issue lies in the recovery of the correct reasoning process.

\begin{figure}
    \centering
    \begin{subfigure}[t]{0.19\linewidth}
        \begin{adjustbox}{width=\textwidth}
        \begin{tikzpicture}
        \begin{axis}[
            xmin=4, xmax=16,
            ymin=0, ymax=1.0,
            legend pos=north west,
            ymajorgrids=true,
            grid style=dashed,
        ]    
        \addplot[name path=B,draw=green!40,ultra thick] coordinates {(4,0.996)(8,0.996)(12,0.996)(16,0.996)};
        \addplot[name path=A,draw=green!60,ultra thick] coordinates {(4,0.208) (8,0.184) (12,0.188) (16,0.168)};
        \addplot[name path=D,draw=none] coordinates {(4,0) (16,0)};
            
        \addplot[green!10, area legend] fill between[of=B and D, soft clip={domain=0:16}];
        % \addlegendentry{Compilable Code}
            
        \addplot[green!30, area legend] fill between[of=A and D, soft clip={domain=0:16}];
        % \addlegendentry{Correct Result}
            
        \end{axis}
        \end{tikzpicture}
        \end{adjustbox}
        \caption{Text-Davinci-3-code }
        \label{fig:gpt_3_code_prop_valid}
    \end{subfigure}
    \hfill
    \begin{subfigure}[t]{0.194\linewidth}
        \begin{adjustbox}{width=0.98\textwidth} % trick to make caption fit in one line
            \begin{tikzpicture}
            \begin{axis}[
                xmin=4, xmax=16,
                ymin=0, ymax=1.0,
                legend pos=north west,
                ymajorgrids=true,
                grid style=dashed,
            ]    
            \addplot[name path=B,draw=blue!40,ultra thick] coordinates {(4,0.964)(8,0.964)(12,0.972)(16,0.992)};
            \addplot[name path=A,draw=blue!60,ultra thick] coordinates {(4,0.204) (8,0.252) (12,0.22) (16,0.228)};
            \addplot[name path=D,draw=none] coordinates {(4,0) (16,0)};
                
            \addplot[blue!10, area legend] fill between[of=B and D, soft clip={domain=0:16}];
            % \addlegendentry{Compilable Code}
                
            \addplot[blue!30, area legend] fill between[of=A and D, soft clip={domain=0:16}];
            % \addlegendentry{Correct Result}
                
            \end{axis}
            \end{tikzpicture}
        \end{adjustbox}
        \caption{GPT-3.5-Turbo-code }
        \label{fig:gpt_3_5_code_prop_valid}
    \end{subfigure}
    \hfill
    \begin{subfigure}[t]{0.19\linewidth}
        \begin{adjustbox}{width=\textwidth}
            \begin{tikzpicture}
            \begin{axis}[
                xmin=4, xmax=16,
                ymin=0, ymax=1.0,
                legend pos=north west,
                ymajorgrids=true,
                grid style=dashed,
            ]    
            \addplot[name path=B,draw=cyan!40,ultra thick] coordinates {(4,1.0)(8,1.0)(12,0.996)(16,0.991)};
            \addplot[name path=A,draw=cyan!60,ultra thick] coordinates {(4,0.189)(8,0.185)(12,0.197)(16,0.197)};
            \addplot[name path=D,draw=none] coordinates {(4,0) (16,0)};
                
            \addplot[cyan!10, area legend] fill between[of=B and D, soft clip={domain=0:16}];
            % \addlegendentry{Compilable Code}
                
            \addplot[cyan!30, area legend] fill between[of=A and D, soft clip={domain=0:16}];
            % \addlegendentry{Correct Result}
                
            \end{axis}
            \end{tikzpicture}
        \end{adjustbox}
        \caption{GPT-4-code }
        \label{fig:gpt_4_code_prop_valid}
    \end{subfigure}
    \hfill
    \begin{subfigure}[t]{0.19\linewidth}
        \begin{adjustbox}{width=\textwidth}
            \begin{tikzpicture}
            \begin{axis}[
                xmin=4, xmax=16,
                ymin=0, ymax=1.0,
                legend pos=north west,
                ymajorgrids=true,
                grid style=dashed,
            ]    
            \addplot[name path=B,draw=yellow!40,ultra thick] coordinates {(4,1.0)(8,1.0)(12,1.0)(16,1.0)};
            \addplot[name path=A,draw=yellow!60,ultra thick] coordinates {(4,0.012)(8,0.008)(12,0.012)(16,0.008)};
            \addplot[name path=D,draw=none] coordinates {(4,0) (16,0)};
                
            \addplot[yellow!10, area legend] fill between[of=B and D, soft clip={domain=0:16}];
            % \addlegendentry{Compilable Code}
                
            \addplot[yellow!30, area legend] fill between[of=A and D, soft clip={domain=0:16}];
            % \addlegendentry{Correct Result}
                
            \end{axis}
            \end{tikzpicture}
        \end{adjustbox}
        \caption{Alpaca-code }
        \label{fig:alpaca_code_prop_valid}
    \end{subfigure}
    \hfill
    \begin{subfigure}[t]{0.19\linewidth}
        \begin{adjustbox}{width=\textwidth}
            \begin{tikzpicture}
            \begin{axis}[
                xmin=4, xmax=16,
                ymin=0, ymax=1.0,
                legend pos=north west,
                ymajorgrids=true,
                grid style=dashed,
            ]    
            \addplot[name path=B,draw=red!40,ultra thick] coordinates {(4,0.972)(8,0.952)(12,0.976)(16,0.86)};
            \addplot[name path=A,draw=red!60,ultra thick] coordinates {(4,0.004)(8,0.008)(12,0.0)(16,0.0)};
            \addplot[name path=D,draw=none] coordinates {(4,0) (16,0)};
                
            \addplot[red!10, area legend] fill between[of=B and D, soft clip={domain=0:16}];
            % \addlegendentry{Compilable Code}
                
            \addplot[red!30, area legend] fill between[of=A and D, soft clip={domain=0:16}];
            % \addlegendentry{Correct Result}
                
            \end{axis}
            \end{tikzpicture}
        \end{adjustbox}
        \caption{Alpaca-LoRA-code }
        \label{fig:alpaca_lora_code_prop_valid}
    \end{subfigure}
    \caption{Evolution of the performance of code models on the PVR dataset as a function of the number of examples seen. The x-axis shows the number of examples and the y-axis shows the accuracy. Lightly coloured areas represent the proportion of samples where the code compiles correctly and dark coloured areas represent the proportion of samples where the generated program accurately answers the question. }
    \label{fig:code_prop_valid}
\end{figure}

\label{sec:code_plus}

\subsection{Varying the Model Size}

In this section, we compare the performance of models of various sizes. We divide our experiments into two parts. First, we evaluate fine-tuned RoBERTa-AR$^*$ and MERIt-AR$^*$ on an MCQA dataset. We aim to see if specialised models with smaller sizes can perform multiple-choice abstract reasoning. Second, we perform additional experiments on the bigger version of LLaMA, i.e. LLamA-13B and LLaMA-30B. We aim to see if increasing the size of the model has an impact on the performance.

\paragraph{MCQA Engines} MCQA models have an advantage over completion engines as they must select one answer from a list of possible choices, whereas completion models must generate the correct answer. Therefore, MCQA models can reach the performance of a random classifier without knowing anything about the task. We perform experiments on the ACRE$^T$-Text and ACRE$^T$-Symbolic datasets. 
The fine-tuned models are trained for 10 epochs with a batch size of 10, using AdamW optimizer \citep{DBLP:conf/iclr/LoshchilovH19} and a learning rate of $5 \times 10^{-4}$.
Results with RoBERTa-AR$^*$ and MERIt-AR$^*$ are shown in Table \ref{tab:qa_engines_mc_qa_results}. When fine-tuned on the training set with the same format, the performance of the model increases slightly. However, the overall performance remains close to random.

\begin{table}[ht]
    \centering
    \caption{Accuracy of the specified model for a Multiple-Choice QA task on the ACRE dataset. Rows represent the dataset on which the model is fine-tuned, and columns represent the dataset on which the model is evaluated. The best result for each dataset in indicated in \textbf{bold}. }
    \begin{minipage}[t][][b]{.48\textwidth}
    \begin{tabular}{llcc}
        \hline
        \multicolumn{2}{c}{\multirow{2}{*}{RoBERTa-AR$^*$}} & \multicolumn{2}{c}{ACRE$^T$-Eval} \\
        \cline{3-4}
         & & Text & Symb \\
        \hline
        \multirow{2}*{ACRE$^T$-Train}& Text & \textbf{0.370} & 0.361 \\
        & Symb & 0.262 & \textbf{0.371} \\
        \hline
    \end{tabular}
    \end{minipage}
    \hfill
    \begin{minipage}[t][][b]{.48\textwidth}
    \begin{tabular}{llcc}
        \hline
        \multicolumn{2}{c}{\multirow{2}{*}{MERIt-AR$^*$}} & \multicolumn{2}{c}{ACRE$^T$-Eval} \\
        \cline{3-4}
         & & Text & Symb \\
        \hline
        \multirow{2}*{ACRE$^T$-Train} & Text & \textbf{0.338} & 0.331 \\
        & Symb & 0.332 & \textbf{0.336} \\
        \hline
    \end{tabular}
    \end{minipage}
    \label{tab:qa_engines_mc_qa_results}
\end{table}

\paragraph{LLaMA Variations} The main results with the various versions of LLaMA on Open QA datasets are displayed in Table \ref{tab:llama_open_qa_results}. We observe a slight increase in accuracy with LLaMA-13B on ARC$^T$, Evals-S, and PVR datasets, but the accuracy then decreases with LLaMA-30B. Performance remains close to null on the RAVEN$^T$ datasets. However, on BIG-Bench-F, the accuracy increases with LLaMA-30B. The overall performance remains poor on every dataset.

\begin{table}[ht]
    \centering
    \caption{Main results of LLaMA versions for open QA. Datasets are represented in columns and models in rows. The best result for each dataset in indicated in \textbf{bold} and the second best is indicated in \textit{italics}. }
    \begin{tabular}{lccccccc}
        \hline
         & \multirow{2}*{ARC$^T$} & \multirow{2}*{BIG-Bench-F} & \multirow{2}*{Evals-S} & \multirow{2}*{PVR} & \multicolumn{2}{c}{RAVEN$^T$-opqa} \\
        \cline{6-7}
         & & & & & Text & Symb \\
        \hline
        LLaMA-7B & \textit{0.010} & \textit{0.012} & 0.014 & 0.060 & 0.000 & 0.000 \\
        LLaMA-13B &  \textbf{0.019} & 0.008 & \textbf{0.029} & \textbf{0.204} & 0.000 & 0.001 \\
        LLaMA-30B & 0.006 & \textbf{0.088} & \textit{0.016} & \textit{0.172} & 0.000 & 0.000 \\
        % LLaMA-65B & \\
        \hline
    \end{tabular}
    \label{tab:llama_open_qa_results}
\end{table}

\subsection{Fine-tuning LLaMA}

We now study the performance of LLaMA and LLaMA2 models after fine-tuning. We fine-tune the models using LoRA for 3 epochs using the AdamW optimizer \citep{DBLP:conf/iclr/LoshchilovH19} with a batch size of 64. As we aim to study the abstract reasoning abilities of LLMs, fine-tuned models' results must be analysed with care. Our goal is to investigate the abilities of the models to extract abstract patterns from a small set of examples. As seen with the example of GPT-4, this task can be bypassed if some samples are in the training data of the model. This problem is prevalent with fine-tuning. The training and test sets may share distribution-specific patterns that the model may learn during the fine-tuning phase and overfit on these patterns. Therefore, we generate out-of-distribution (o.o.d) splits for each dataset to alleviate this pitfall. We conduct our experiments on ARC$^T$, ACRE$^T$, RAVEN$^T$ and PVR datasets.

\paragraph{ARC$^T$}

The results on ARC$^T$ are shown in Table \ref{tab:fine-tuned-arc}. The accuracy almost doubles with the fine-tuned models but remains low and below the performance achieved by other models like GPT-4 (with an accuracy of 0.119). This result is expected. The ARC dataset is very challenging and the size of the training set is small ($\sim 400$ samples).

\begin{table}[t]
    \centering
    \caption{Accuracy of base and fine-tuned models on the ARC dataset. ARC$^T$-Eval is the test set used in the main experiments. }
    \begin{tabular}{llc}
        \hline
        \multirow{2}*{Model} & Test Set $\Rightarrow$ & \multirow{2}*{ARC$^T$-Eval} \\
        & Tuning Set $\Downarrow$ & \\
        \hline
        LLaMA & & 0.010 \\
        LLaMA2 & & 0.005 \\
        \hline
        LLaMA-7B-AR-LoRA* & ARC$^T$-Train & 0.018 \\
        LLaMA2-7B-AR-LoRA* & ARC$^T$-Train & 0.010 \\
        \hline
    \end{tabular}
    \label{tab:fine-tuned-arc}
\end{table}

\paragraph{ACRE$^T$}

The results on ACRE$^T$ are shown in Table \ref{tab:fine-tuned-acre}. The training set for ACRE$^T$ contains 24K samples. The fine-tuned LLaMA and LLaMA2 achieve very good performance on the i.i.d test set, with LLaMA2 reaching close to perfect accuracy. We also observe that fine-tuning one model on the \textit{Text} version of the task increases the performance on the \textit{Symbolic} task. The converse holds for LLaMA2: fine-tuning on the \textit{Symbolic} task increases performance on the \textit{Text} task. This effect is not observed with LLaMA.
This test provides evidence that fine-tuning increases performance and generalisation abilities. LLaMA2 can transfer to the alternative syntax with good accuracy without being trained on it. The results remain lower than for the same-syntax task. 
To further investigate if this observation holds in other settings, we perform experiments on additional splits, following the division made by \citet{DBLP:conf/cvpr/0017JEZZ21}. The \textit{compositional} split (-Comp) uses a different composition of objects than in the base split. E.g. "red cylinders in metal" than are never seen in the training set ("red", "cylinder", and "metal" all are in the training set but never combined together). The \textit{systematic} split (-Sys) changes the context distribution. For each sample in the training set, the context information shows 3 examples where the light is activated. In the \textit{systematic} split, four examples are shown. We find to significant performance changes on these o.o.d splits compared to the i.i.d split. The representations generated by the LLMs seem to be invariant to the sample compositions and to small presentation changes, and partially invariant to major syntax changes (\textit{Text} vs \textit{Symbolic}).

\begin{table}[t]
    \centering
    \caption{Accuracy of base and fine-tuned models on the ACRE dataset i.i.d and o.o.d splits. Rows represent the dataset on which the model is fine-tuned, and columns represent the dataset on which the model is evaluated. ACRE$^T$-Eval is the test set used in the main experiments. The LLaMA version is omitted from the fine-tuned model names for conciseness. The best result for each dataset in indicated in \textbf{bold}. }
    \begin{tabular}{lllcccccc}
        \hline
         \multirow{2}{*}{Model} & \multicolumn{2}{c}{Test Set $\Rightarrow$} & \multicolumn{2}{c}{ACRE$^T$-Eval} & \multicolumn{2}{c}{-Comp} & \multicolumn{2}{c}{-Sys} \\
        \cline{4-9}
         & \multicolumn{2}{c}{Tuning Set $\Downarrow$} & Text & Symb & Text & Symb & Text & Symb \\
        \hline
        LLaMA-7B & & & 0.000 & 0.257 & 0.000 & 0.033 & 0.000 & 0.021 \\
        \multirow{2}*{~~-AR-LoRA$^*$} & \multirow{2}*{ACRE$^T$-Train} & Text & \textbf{0.755} & 0.614 & \textbf{0.741} & 0.606 & \textbf{0.727} & 0.550 \\
        & & Symb & 0.081 & \textbf{1.000} & 0.102 & \textbf{0.999} & 0.095 & \textbf{0.999} \\
        \hline
        LLaMA2-7B & & & 0.246 & 0.003 & 0.244 & 0.001 & 0.288 & 0.001 \\
        \multirow{2}*{~~-AR-LoRA$^*$} & \multirow{2}*{ACRE$^T$-Train} & Text & \textbf{0.997} & 0.662 & \textbf{1.000} & 0.651 & \textbf{0.994} &  0.626 \\
        & & Symb & 0.568 & \textbf{1.000} & 0.579 & \textbf{1.000} & 0.539 & \textbf{0.999} \\
        \hline
    \end{tabular}
    \label{tab:fine-tuned-acre}
\end{table}

\paragraph{RAVEN$^T$-mcqa} 

The results on RAVEN$^T$-mcqa are shown in Table \ref{tab:fine-tuned-raven}. Given the low performance of the base LLaMA and LLaMA2 on the Multiple Choices Question Answering settings of RAVEN$^T$, we restrict our experiments to this settings. The training set for RAVEN$^T$ contains 9K samples. We observe a significant increase in the accuracy on the test set for both fine-tuned LLaMA and LLaMA2. As for ACRE$^T$, LLaMA2 achieves close to perfect accuracy. Again, similarly to ACRE$^T$, the performance partially transfers to the alternative syntax task. Notably, the LLaMA2 fine-tuned on the \textit{Symbolic} RAVEN$^T$-Train reaches an accuracy of 96.5\% on the \textit{Text} task.
We now observe the performance on additional o.o.d splits. The \textit{-Four} split contains samples with four figures instead of one. The \textit{-In-Center} splits contains samples with two figures instead of one, a big and a small located within the former. The shape and colours of the figures all are observed in the training set. The two splits can be considered as compositional splits. The performance of the fine-tuned models significantly drops on the new tasks, in particular the accuracy of LLaMA collapses. We can observe a ferw interesting fact with LLaMA2. First, on the \textit{-Four} split, fine-tuning on the \textit{Text} task yields better performance on both \textit{Text} and \textit{Symbolic} tests than when fine-tuning on the \textit{Symbolic} task. Curiously, for the \textit{-In-Center} split, the best performance on the \textit{Text} test is achieved by the model fine-tuned on the \textit{Symbolic} task. We can deduce that fine-tuning yields representations that are highly invariant to the syntax. However, it does not transfer most of the abstract reasoning abilities. The rules required to solve the \textit{-Four} and \textit{-In-Center} splits manipulate several figures, they are compositions of rules for single figures. In the ACRE$^T$ \textit{compositional} split, the rules to learn are the same but the objects to manipulate are compositions of seen objects. We can deduce that LLMs can compose with unseen quantities but have more difficulty composing new abstract rules.

\begin{table}[t]
    \centering
    \caption{Accuracy of base and fine-tuned models on the RAVEN$^T$-mcqa dataset i.i.d and o.o.d splits. Rows represent the dataset on which the model is fine-tuned, and columns represent the dataset on which the model is evaluated. RAVEN$^T$-Eval is the test set used in the main experiments. The LLaMA version is omitted from the fine-tuned model names for conciseness. The best result for each dataset in indicated in \textbf{bold}. }
    \begin{tabular}{lllcccccc}
        \hline
        \multirow{2}{*}{Model} & \multicolumn{2}{c}{Test Set $\Rightarrow$} & \multicolumn{2}{c}{RAVEN$^T$-Eval} & \multicolumn{2}{c}{-Four} & \multicolumn{2}{c}{-In-Center} \\
        \cline{4-9}
         & \multicolumn{2}{c}{Tuning Set $\Downarrow$} & Text & Symb & Text & Symb & Text & Symb \\
        \hline
        LLaMA-7B & & & 0.004 & 0.000 & 0.000 & 0.000 & 0.000 & 0.000 \\
        \multirow{2}*{~~-AR-LoRA$^*$} & \multirow{2}*{RAVEN$^T$-Train} & Text & \textbf{0.558} & 0.322 & \textbf{0.050} & 0.168 & 0.000 & 0.010 \\
        & & Symb & 0.232 & \textbf{0.460} & 0.014 & \textbf{0.287} & \textbf{0.002} & \textbf{0.016 }\\
        \hline
        LLaMA2-7B & & & 0.135 & 0.114 & 0.073 & 0.121 & 0.000 & 0.001 \\
        \multirow{2}*{~~-AR-LoRA$^*$} & \multirow{2}*{RAVEN$^T$-Train} & Text & \textbf{0.977} & 0.694 & \textbf{0.557} & \textbf{0.522} & 0.536 & \textbf{0.085} \\
        & & Symb & 0.965 & \textbf{0.938} & 0.498 & 0.442 & \textbf{0.767} & 0.064 \\
        \hline
    \end{tabular}
    \label{tab:fine-tuned-raven}
\end{table}

\paragraph{PVR}

The results on PVR are shown in Table \ref{tab:fine-tuned-pvr}. The training set for PVR contains 1K samples. The accuracies for the base LLaMA and LLaMA2 are 0.060 and 0.000, respectively. Fine-tuning significantly increases the performance of both models on the i.i.d test set. We construct multiple o.o.d splits. The \textit{compositional} (-Comp) split modifies the number of variables taken by the retrieval function. Composition-0 takes the variable pointed by the index while composition-N adds N extra variables (at location $index + n~\forall n \in [1\dots N]$) and sums them (modulo 10). The \textit{Holdout} split changes the distribution of the arrays. The holdout training set distribution is biased to force some values to do not appear at some given positions. The test set contains the complementary set. This split is used to verify if the model learns the PVR task or uses distribution-specific knowledge to solve the problem at hand. We can see that the fine-tuned models maintain their performance on the \textit{Holdout} split but fail to transfer to different function compositions. This observation is consistent with the results observed with RAVEN$^T$.

\begin{table}[t]
    \centering
    \caption{Accuracy of fine-tuned models on the PVR dataset i.i.d and o.o.d splits. Rows represent the dataset on which the model is fine-tuned, and columns represent the dataset on which the model is evaluated. PVR-Eval Comp-0 is the test set used in the main experiments. The best result for each dataset in indicated in \textbf{bold}. }
    \begin{tabular}{lccccccc}
        \hline
        \multirow{2}*{Model} & Test Set $\Rightarrow$ & \multicolumn{3}{c}{PVR-Eval} & \multicolumn{3}{c}{-Holdout} \\
        \cline{3-8}
         & Tuning Set $\Downarrow$ & Comp-0 & -1 & -2 & Comp-0 & -1 & -2 \\
        \hline
        LLaMA & \multirow{2}*{PVR-Train Comp0} & \multirow{2}*{0.496} & \multirow{2}*{0.110} & \multirow{2}*{0.100} & \multirow{2}*{0.483} & \multirow{2}*{0.107} & \multirow{2}*{0.118} \\
        ~~-AR-LoRA$^*$ \\
        LLaMA2 & \multirow{2}*{PVR-Train Comp0} & \multirow{2}*{0.728} & \multirow{2}*{0.098} & \multirow{2}*{0.100} & \multirow{2}*{0.708} & \multirow{2}*{0.116} & \multirow{2}*{0.122} \\
        ~~-AR-LoRA$^*$ \\
        \hline
    \end{tabular}
    \label{tab:fine-tuned-pvr}
\end{table}

\subsection{Varying the Model Temperature}
\label{sec:temperature}

We noticed in our experiments that the LLMs tend to repeat similar wrong reasoning patterns across samples or produce repeating sequences when they cannot identify the abstract pattern. Without fine-tuning, LlaMA is particularly susceptible to this issue. To reduce the number of occurrences of this problem, we set the temperature of the models in our experiments to a high value (temperature=0.5). Setting a high temperature increases the probability for the model to output different and non-repeating answers. For our experiments, it gives the opportunity for the models to explore a larger variety of reasoning paths. On the other hand, reducing the temperature reduces the uncertainty in the answer. A low temperature is usually associated with high fidelity answer while models with high temperature are more prone to hallucinations \citep{DBLP:conf/pldi/Xu0NH22}. 

We perform additional experiments where we vary the temperature of GPT-3.5-Turbo and GPT-4 to study the impact of this factor on performance. We use the base and code versions of these models to see if differences occur between models generating long answers and models generating short answers. We perform experiments with temperatures: $[0.0, 0.25, .05, 0.75, 1.0]$. The results on the BIG-Bench-F and PVR datasets are shown in Figure \ref{fig:temp}. We observe that there is no significant difference between code and base models. On both datasets, varying the temperature has little impact on the accuracy. On the PVR dataset, the accuracy remains similar for all models. On BIG-Bench-F, the accuracy drops when the temperature is equal to 1.0. The accuracy also drops for GPT-4 when the temperature is equal to 0.25 and 0.5 but increases when reaching 0.75. The standard deviation remains small (0.028). This phenomenon is not observed on the code model.

\begin{figure}
    \centering
    \begin{subfigure}{0.48\linewidth}
    \begin{adjustbox}{width=0.95\linewidth}
    \begin{tikzpicture}
    \begin{axis}[
        xlabel={Temperature},
        ylabel={Accuracy},
        xmin=0, xmax=1,
        ymin=0, ymax=1,
        legend pos=north west,
        legend columns=2
        ymajorgrids=true,
        grid style=dashed,
    ]
    
    \addplot[
        color=blue,
        mark=diamond,
        ]
        coordinates {
        (0,0.444)(0.25,0.444)(0.50,0.440)(0.75,0.431)(1.0,0.384)
        };
        \addlegendentry{gpt-3.5-turbo\footnotemark[1]}
    
    \addplot[
        color=blue,
        mark=diamond,
        dashed
        ]
        coordinates {
        (0,0.301)(0.25,0.309)(0.50,0.301)(0.75,0.304)(1.0,0.3)
        };
        \addlegendentry{gpt-3.5-turbo-code\footnotemark[1]}
    
    \addplot[
        color=cyan,
        mark=square,
        ]
        coordinates {
        (0,0.584)(0.25,0.52)(0.50,0.516)(0.75,0.576)(1.0,0.552)
        };
        \addlegendentry{gpt-4}
    
    \addplot[
        color=cyan,
        mark=square,
        dashed
        ]
        coordinates {
        (0,0.46)(0.25,0.456)(0.50,0.476)(0.75,0.468)(1.0,0.432)
        };
        \addlegendentry{gpt-4-code}
        
    \end{axis}
    \end{tikzpicture}
    \end{adjustbox}
    \caption{BIG-Bench-F dataset. }
    \label{fig:temp_bench}
    \end{subfigure}
    \hfill
    \begin{subfigure}{0.48\linewidth}
    \begin{adjustbox}{width=0.95\linewidth}
    \begin{tikzpicture}
    \begin{axis}[
        xlabel={Temperature},
        ylabel={Accuracy},
        xmin=0, xmax=1,
        ymin=0, ymax=1,
        legend pos=north west,
        legend columns=2
        ymajorgrids=true,
        grid style=dashed,
    ]
    
    \addplot[
        color=blue,
        mark=diamond,
        ]
        coordinates {
        (0,0.184)(0.25,0.222)(0.50,0.189)(0.75,0.177)(1.0,0.172)
        };
        \addlegendentry{gpt-3.5-turbo\footnotemark[1]}
    
    \addplot[
        color=blue,
        mark=diamond,
        dashed
        ]
        coordinates {
        (0,0.204)(0.25,0.216)(0.50,0.204)(0.75,0.18)(1.0,0.185)
        };
        \addlegendentry{gpt-3.5-turbo-code\footnotemark[1]}
    
    \addplot[
        color=cyan,
        mark=square,
        ]
        coordinates {
        (0,0.184)(0.25,0.184)(0.50,0.177)(0.75,0.192)(1.0,0.186)
        };
        \addlegendentry{gpt-4}
    
    \addplot[
        color=cyan,
        mark=square,
        dashed
        ]
        coordinates {
        (0,0.188)(0.25,0.22)(0.50,0.189)(0.75,0.214)(1.0,0.176)
        };
        \addlegendentry{gpt-4-code}

    \end{axis}
    \end{tikzpicture}
    \end{adjustbox}
    \caption{PVR dataset. }
    \label{fig:temp_pvr}
    \end{subfigure}
    \caption{Evolution of the performance of GPT models when varying temperature. }
    \label{fig:temp}
\end{figure}

\subsection{Providing Hints to the Model}

To disambiguate the source of the confusion in the LLMs in the failure cases, we study another prompt where we provide additional hints to the model. Each hint corresponds to a solved instance from the training dataset. It contains the context, the test case and its answer, and the ground truth reasoning path. This reasoning path is represented as a Python function. This choice avoids unwanted ambiguities from natural language and can be easily integrated with the code models. We run experiments on GPT-3.5-Turbo and GPT-4 on BIG-Bench-F. Zero-hints models correspond to the base code models.

Figure \ref{fig:hint_results} shows the results. We observe no significant variations on the performance of GPT-4. The accuracy of GPT-3.5-Turbo increases slightly when given one hint, increasing from 0.301 to 0.355, but does not increase more when given more hints. These experiments highlight that the failures of the models do not come from a misunderstanding of the task or the prompt but from the difficult nature of the task. This observation is confirmed when looking into the responses generated by the models (in Appendix \ref{sec:bench_ex}).

\begin{figure}
    \centering
    \begin{tikzpicture}
    \begin{axis}[
        xlabel={Number of hints},
        ylabel={Accuracy},
        xmin=0, xmax=4,
        ymin=0, ymax=1,
        legend pos=north west,
        legend columns=2
        ymajorgrids=true,
        grid style=dashed,
        width=0.5\linewidth
    ]
    
    \addplot[
        color=blue,
        mark=diamond,
        dashed
        ]
        coordinates {
        (0,0.301)(1,0.355)(2,0.345)(3,0.324)(4,0.321)
        };
        \addlegendentry{gpt-3.5-turbo-code\footnotemark[1]}
    
    \addplot[
        color=cyan,
        mark=square,
        dashed
        ]
        coordinates {
        (0,0.476)(1,0.459)(2,0.455)(3,0.439)(4,0.459)
        };
        \addlegendentry{gpt-4-code}

    \end{axis}
    \end{tikzpicture}
    \caption{Evolution of the accuracy of hinted GPT code models on the BIG-Bench-F dataset. Hints correspond to solved instances of the training set and are given as examples to the model as part of the pre-prompt. They contain the context examples, the answer to the test case, and the ground truth function that generates the output from the input. }
    \label{fig:hint_results}
\end{figure}

\footnotetext[1]{Please note that these experiments with GPT-3.5-Turbo have been performed at a later date than the other ones so the exact results may differ due model updates in the OpenAI API. The version used is gpt-3.5-turbo-0613. }

\subsection{Entropy as an Abstraction Measure}

We investigate further the experiments performed on the code models under the prism of Information theory. We modify the generation task into a classification task to measure the discriminative abilities of our studied models. We generate a new corrupted dataset from the test set by modifying the output of each sample so that it does not match the pattern. For each task, the program $P$ built by the language model is tasked to predict if the sample belongs to the original set or the corrupted set. We measure the resulting Information Gain (IG) or Mutual Information:

\begin{equation}
    IG(T,P) = H(T) - H(T|P)
\end{equation}

$T$ corresponds to the classification task. The entropy $H(T)$ is equal to 1 as the two outputs ("sample follows the pattern" and "sample does not follow the pattern" are balanced). The entropy $H(T|P)$ corresponds to the remaining entropy given the output of the program $P$. The Information Gain measures the amount of information regarding the class of the sample that has been captured by the program. The Information Gain should be high if the program captured the general pattern and low if it is grounded to particular instances or captured only sub-parts of the pattern.

\begin{figure}
    \centering
    \begin{subfigure}{0.48\linewidth}
    \begin{adjustbox}{width=0.95\linewidth}
    \begin{tikzpicture}
    \begin{axis}[
        xlabel={Number of example cases},
        ylabel={Information Gain},
        xmin=2, xmax=16,
        ymin=0, ymax=1,
        legend pos=north east,
        legend columns=1,
        ymajorgrids=true,
        grid style=dashed,
    ]
    
    \addplot[
        color=blue,
        mark=diamond,
        loosely dashdotdotted
        ]
        coordinates {
        (2,0.170)(4,0.204)(6,0.182)(8,0.205)(10,0.229)(12,0.268)(14,0.225)(16,0.265)
        };
        % \addlegendentry{gpt-3.5-turbo-code}
    
    \addplot[
        color=cyan,
        mark=square,
        dashed
        ]
        coordinates {
        (2,0.234)(4,0.311)(6,0.226)(8,0.027)(10,0.301)(12,0.176)(14,0.384)(16,0.030)
        };
        % \addlegendentry{gpt-4-code}
        
    \end{axis}
    \end{tikzpicture}
    \end{adjustbox}
    \caption{BIG-Bench-F dataset. }
    \label{fig:information_gain_results_bench}
    \end{subfigure}
    \hfill
    \begin{subfigure}{0.48\linewidth}
    \centering
    \begin{adjustbox}{width=0.95\linewidth}
    \begin{tikzpicture}
    \begin{axis}[
        xlabel={Number of example cases},
        ylabel={Information Gain},
        xmin=4, xmax=16,
        ymin=0, ymax=1,
        legend pos=north east,
        legend columns=1,
        ymajorgrids=true,
        grid style=dashed,
    ]
    
    \addplot[
        color=blue,
        mark=diamond,
        loosely dashdotdotted
        ]
        coordinates {
        (4,0.025)(8,0.024)(12,0.026)(16,0.024)
        };
        \addlegendentry{gpt-3.5-turbo-code}
    
    \addplot[
        color=cyan,
        mark=square,
        dashed
        ]
        coordinates {
        (4,0.015)(8,0.018)(12,0.017)(16,0.016)
        };
        \addlegendentry{gpt-4-code}

    \end{axis}
    \end{tikzpicture}
    \end{adjustbox}
    \caption{PVR dataset. }
    \label{fig:information_gain_results_pvr}
    \end{subfigure}
    \caption{Evolution of the Information Gain of GPT code models as a function of the number of examples seen. Information Gain measures the ability of the generated program to discriminate samples following the abstract pattern and samples not following it. The higher the better. The legend is shared by both figures.}
    \label{fig:information_gain_results}
\end{figure}

Figure \ref{fig:information_gain_results} shows the results on BIG-Bench-F and pVR for GPT-3.5-Turbo and GPT-4. The Information Gain remains low for both models. On the PVR dataset, IG is constantly low and close to zero, indicating that the programs have overfitted to specific instances. On BIG-Bench-F, the IG for GPT-3.5 remains constant but slightly increases as the number of context examples during training increases. Increasing the number of samples has a positive effect on generalisation. However, the IG varies significantly for GPT-4, IG has high variations, highlighting instability in the program generation, despite having the highest accuracy across all code models. This indicates that GPT-4 tend to unpredictably generate programs that overfit to the samples presented instead of grasping general rules. An example is given in Appendix \ref{sec:bench_ex}.

\clearpage
\section{Comparison Across Dataset Features}

This section presents an in-depth analysis of the dataset characteristics and of the results with respect to these characteristics, in particular relative to the types of causal queries.

\subsection{Features of Interest}

Table \ref{tab:features} shows the features of interest of each dataset. The \textit{Average Words per Context} column shows the average size of an instance prompt. The ARC$^T$ dataset has the largest context size by a great margin because of the high dimensionality of the grid input. Text inputs also have a greater size than their symbolic counterparts. 

The \textit{Task in Training Data} column estimates the chances of specific instances of the dataset to be in the training data of the studied models. As mentioned in the previous paragraph, PVR and ACRE have been created after the training of these models are cannot be in their training set. Evals-P and Evals-S are taken from datasets used to evaluate LLMs so it is unlikely they have been used for their training. RAVEN$^T$ is based on Raven Progressive Matrices \citep{raven1938raven}, a long-existing intelligence test. Substantial resources and instances can be found online so the chances that LLMs have been trained on instances of the test are very likely. Moreover, as shown in Appendix \ref{sec:details}, GPT-3.5-Turbo and GPT-4 know and can generate RAVEN matrices. The same is observed for BIG-Bench-F.

\begin{table}[ht]
    \centering
    \small
    \caption{Datasets considered and their features of interest. When not written, type is similar to the one above. Text datasets built from an image dataset are indicated with the symbol $^T$. Datasets can exist in text or symbolic versions. Text and symbolic splits can have different values for one feature of the same dataset. In those cases, both values are indicated, separated by a "/". }
    \begin{minipage}[t][][b]{\textwidth}
    \centering
    \begin{tabular}{llccccc}
        \hline
        \multirow{2}*{Dataset} & \multirow{2}*{Type} & \multirow{2}*{Eval Size} & \multicolumn{2}{c}{Versions} & Average Words & Task in  \\
        \cline{4-5}
         & & & Text & Symb & per Context & Training Data \\
        \hline
        ARC$^T$ & Open QA & 419 & & \checkmark & 1588.01 & No \\
        BIG-Bench-F & & 250 & & \checkmark & 88.97 & Likely \\
        Evals-S & & 70 & & \checkmark & 78.10 & Unlikely \\
        PVR & & 250 & & \checkmark & 83.0 & No \\
        ACRE$^T$ & MCQA & 1000 & \checkmark & \checkmark & 173.88 / 65.55 & No \\
        Evals-P & & 250 & & \checkmark & 155.00 & Unlikely \\
        RAVEN$^T$ & & 1000 & \checkmark & \checkmark & 198.50 / 114.50 & Very likely \\
        \hline
    \end{tabular}
    \end{minipage}
    \hfill
    \vspace{0.001cm}
    \begin{minipage}[t][][b]{\textwidth}
    \centering
    \begin{tabular}{lcccccc}
        \hline
        \multirow{2}*{Dataset} & \multicolumn{4}{c}{Causal Induction} \\
        \cline{2-5}
         & Direct & Indirect & backward-Blocking & Screening-Off \\
        \hline
        ARC$^T$ & \checkmark & \checkmark & & \checkmark \\
        BIG-Bench-F & \checkmark & & & \checkmark & \\
        Evals-S & \checkmark & \checkmark & & \\
        PVR & \checkmark & \checkmark & & \checkmark & \\
        ACRE$^T$ & \checkmark & \checkmark & \checkmark & \checkmark \\
        Evals-P & \checkmark & & & & \\
        RAVEN$^T$ & \checkmark & \checkmark & & \checkmark \\
        \hline
    \end{tabular}
    \end{minipage}
    \label{tab:features}
\end{table}

The \textit{Causal Induction} columns show the type of causal paths represented in the instances of the dataset. We use the same terminology as \citet{DBLP:conf/cvpr/0017JEZZ21}. 
Direct paths correspond to single-step inferences. They can be established using direct evidence. All datasets contain instances with direct paths.
Indirect paths require several steps of inference and need to combine multiple pieces of evidence. ARC$^T$, Evals-S, PVR, ACRE$^T$, and RAVEN$^T$ contain indirect paths.
Backward-blocking paths cannot be determined because the true mechanisms cannot be discriminated from other possible mechanisms based only on the data. We consider that only ACRE$^T$ contains such instances. We would like to raise the reader's awareness on the fact that some instances in the other datasets may still contain backward-blocking paths. This can happen when several mechanisms satisfy the constraints in the data. For instance, a key-value mapping between the inputs and the outputs will perfectly fit the data. However, we consider that the expected mechanism can be discriminated via other means, e.g. by favouring short and sparse causal paths or low-entropy methods.
Screening-off paths are causal paths affected by spurious correlations. For instances, parts of an instance may not be on the causal path (i.e. have no effect on the outcome) but can be correlated with a particular outcome. Screening-off tasks use a negatively correlated true outcome to verify if the model learned the true causal path or the correlation. ARC$^T$, BIG-Bench-F, PVR, ACRE$^T$, and RAVEN$^T$ contain screening-off paths.

\subsection{Causal Induction Results}

We study the accuracy of the language models for each type of causal path induction. We focus our analysis to the ACRE$^T$ dataset as it is the only one with instances matching the four types of causal paths. Figures \ref{fig:acre_txt_causal_res}, \ref{fig:acre_symb_causal_res} and \ref{fig:acre_causal_res_cot} present the results.

Figure \ref{fig:acre_txt_causal_res} shows the results of the base models on the text version of ACRE$^T$. GPT-2 and GPT-4 models tend to overfit to a single type of path. When looking at the generated answers, we observe that GPT-2 returns systematically the same answer, achieving close to random performance while GPT-4 very often states that it cannot answer the query. This response is classified as "undetermined". The results are very different on the symbolic version, shown in Figure \ref{fig:acre_symb_causal_res}. The accuracy is balanced across models and between the reasoning paths. This can be explained by the removal of spurious effects arising with language. The best accuracy is almost systematically achieved on the direct evidence queries. The first exception is Text-Davinci-3, which behaves similarly to GPT-4 on the text version. Models also tend to recognise screening-off cases more easily than indirect and backward-blocking paths. The performance remains poor overall, most models performing below chance.

\begin{figure}
    \centering
    \begin{tikzpicture}
    \begin{axis}[
        ybar,
        ymin=0,
        ymax=1,
        height=8cm,
        width=\linewidth,
        bar width=8pt,
        xtick=data,
        tick label style={rotate=45},
        xticklabels={GPT-2,Text-Davinci-3,GPT-3.5-Turbo,GPT-4,LLaMA-7B,Alpaca,Alpaca-LoRA}
        ]
    
    \addplot coordinates {(1,0.330) (2,0.028) (3,0.336) (4,0.046) (5,0.0) (6,0.044) (7,0.287)};
    \addplot coordinates {(1,0.000) (2,0.000) (3,0.342) (4,0.000) (5,0.0) (6,0.140) (7,0.013)};
    \addplot coordinates {(1,0.000) (2,0.304) (3,0.017) (4,0.988) (5,0.0) (6,0.000) (7,0.000)};
    \addplot coordinates {(1,1.000) (2,0.042) (3,0.064) (4,0.013) (5,0.0) (6,0.011) (7,0.013)};

    \legend {Direct, Indirect, Backward-blocking, Screening-off};
    \end{axis}
    \end{tikzpicture}
    \caption{Results of base models on the text version of ACRE$^T$. }
    \label{fig:acre_txt_causal_res}
\end{figure}

\begin{figure}
    \centering
    \begin{tikzpicture}
    \begin{axis}[
        ybar,
        ymin=0,
        ymax=1,
        height=8cm,
        width=\linewidth,
        bar width=8pt,
        xtick=data,
        tick label style={rotate=45},
        xticklabels={GPT-2,Text-Davinci-3,GPT-3.5-Turbo,GPT-4,LLaMA-7B,Alpaca,Alpaca-LoRA}
        ]
    
    \addplot coordinates {(1,0.0) (2,0.356) (3,0.458) (4,0.644) (5,0.357) (6,0.329) (7,0.139)};
    \addplot coordinates {(1,0.0) (2,0.026) (3,0.229) (4,0.395) (5,0.280) (6,0.349) (7,0.172)};
    \addplot coordinates {(1,0.0) (2,0.913) (3,0.336) (4,0.343) (5,0.010) (6,0.118) (7,0.000)};
    \addplot coordinates {(1,0.0) (2,0.220) (3,0.501) (4,0.530) (5,0.361) (6,0.177) (7,0.239)};

    \legend {Direct, Indirect, Backward-blocking, Screening-off};
    \end{axis}
    \end{tikzpicture}
    \caption{Results of base models on the symbolic version of ACRE$^T$. }
    \label{fig:acre_symb_causal_res}
\end{figure}

Figure \ref{fig:acre_text_causal_res_cot} shows the results of the chain-of-thought models on the text version of ACRE$^T$. Chain-of-thought prompts increase the accuracy of GPT-4 on the various causal paths. GPT-4 still often states that it cannot respond but provides more answers than with the base prompting. This is in opposition with what is observed on GPT-3.5-Turbo. The model answers less and instead returns "undetermined" more often. The performance of Alpaca-LoRA remains below chance so no conclusions can be drawn from the results. Similarly to what was observed in Figure \ref{fig:acre_symb_causal_res}, Figure \ref{fig:acre_symb_causal_res_cot} shows accuracy results more evenly distributed among the causal paths. The models do not achieve better than random performance but their answers are more diverse and less biased towards a single class.

\begin{figure}
    \centering
    \begin{subfigure}{0.48\linewidth}
    \begin{adjustbox}{width=0.95\linewidth}
    \begin{tikzpicture}
    \begin{axis}[
        ybar,
        ymin=0,
        ymax=1,
        enlarge x limits={abs=0.6},
        xtick=data,
        tick label style={rotate=45},
        xticklabels={GPT-3.5-Turbo-cot,GPT-4-cot,Alpaca-LoRA-cot}
        ]
    
    \addplot coordinates {(1,0.076) (2,0.124) (3,0.070)};
    \addplot coordinates {(1,0.009) (2,0.026) (3,0.172)};
    \addplot coordinates {(1,0.798) (2,0.530) (3,0.020)};
    \addplot coordinates {(1,0.094) (2,0.119) (3,0.029)};

    \legend {Direct, Indirect, Backward-blocking, Screening-off};
    \end{axis}
    \end{tikzpicture}
    \end{adjustbox}
    \caption{Text ACRE$^T$. }
    \label{fig:acre_text_causal_res_cot}
    \end{subfigure}
    \hfill
    \begin{subfigure}{0.48\linewidth}
    \begin{adjustbox}{width=0.95\linewidth}
    \begin{tikzpicture}
    \begin{axis}[
        ybar,
        ymin=0,
        ymax=1,
        enlarge x limits={abs=0.6},
        xtick=data,
        tick label style={rotate=45},
        xticklabels={GPT-3.5-Turbo-cot,GPT-4-cot,Alpaca-LoRA-cot}
        ]
    
    \addplot coordinates {(1,0.462) (2,0.475) (3,0.104)};
    \addplot coordinates {(1,0.422) (2,0.500) (3,0.164)};
    \addplot coordinates {(1,0.107) (2,0.154) (3,0.000)};
    \addplot coordinates {(1,0.366) (2,0.462) (3,0.231)};

    \legend {Direct, Indirect, Backward-blocking, Screening-off};
    \end{axis}
    \end{tikzpicture}
    \end{adjustbox}
    \caption{Symbolic ACRE$^T$. }
    \label{fig:acre_symb_causal_res_cot}
    \end{subfigure}
    \caption{Results of chain-of-thought models on ACRE$^T$. }
    \label{fig:acre_causal_res_cot}
\end{figure}

\clearpage 
\section{Dive Into Model Responses}

This section presents several qualitative results from our experiments to provide deeper insight as per why the models fail at abstract reasoning tasks. We format the answers before presenting them. In particular, we reduce multiple line returns to a single one and modify arrow symbols from "->" to "$\rightarrow$". We also present code blocks identified by the formatting "```" in verbatim. We provide examples from the PVR and RAVEN$^T$ datasets.

\subsection{PVR Responses}
\label{sec:pvr_ex}

\paragraph{Base Responses}

We give an example of responses from models on the PVR dataset in Figure \ref{fig:pvr-base-ex}. Most models return only the answer as prompted, except GPT-3.5-Turbo. The latter explains its answer. However the reasoning described is invalid. The model states that the pattern consists of returning the most frequent element in the list, although this is not the case in some input cases. Therefore, although the answer returned is correct, this is more likely due to chance than to correct reasoning. All the other models fail to answer correctly except Alpaca-LoRA. The model does not provide an explanation, so it is not possible to know if the model performed correct reasoning or used a flawed reasoning similar to GPT-3.5-Turbo. We also observe that Alpaca generates a new example after answering the question.

\begin{figure}[ht]
    \centering
    \small
    \begin{tcolorbox}[title=Input and Label]
    $[5, 7, 4, 1, 8, 9, 8, 1, 9, 8, 4] \rightarrow 8$ \\
    $[4, 0, 0, 7, 0, 1, 0, 5, 3, 0, 0] \rightarrow 1$ \\
    $[0, 2, 8, 2, 5, 9, 4, 3, 8, 5, 4] \rightarrow 2$ \\
    $[3, 3, 2, 6, 5, 7, 4, 6, 7, 4, 8] \rightarrow 5$ \\
    $[3, 4, 9, 7, 1, 8, 7, 1, 0, 3, 5] \rightarrow$ \textit{\textcolor{red!50}{1}} \\
    
    \end{tcolorbox}
    \begin{tcolorbox}[title=Text-Davinci-3 Output,colframe=blue!55,colback=blue!20]
    7
    \end{tcolorbox}
    \begin{tcolorbox}[title=GPT-3.5-Turbo Output,colframe=blue!55,colback=blue!20]
    The pattern is to find the most frequent element in the list. \\
    In the given test case, the most frequent element is 1, which occurs twice. The other elements occur only once or twice. Therefore, the answer is 1.
    \end{tcolorbox}
    \begin{tcolorbox}[title=GPT-4 Output,colframe=blue!55,colback=blue!20]
    8
    \end{tcolorbox}
    \begin{tcolorbox}[title=Alpaca Output,colframe=blue!55,colback=blue!20]
    8 \\
    $[0, 0, 0, 0, 0, 0, 0, 0, 0, 0, 0] \rightarrow 0$
    \end{tcolorbox}
    \begin{tcolorbox}[title=Alpaca-LoRA Output,colframe=blue!55,colback=blue!20]
    1
    \end{tcolorbox}
    \caption{Examples of response from base models on the PVR dataset. Pre-prompts are omitted from the input. The true answer (not visible to the model) is indicated in \textit{\textcolor{red!50}{italics}}. }
    \label{fig:pvr-base-ex}
\end{figure}

\paragraph{Code Responses} We give an example in Figure \ref{fig:pvr-code-ex}. The models can follow the instructions, generate a correct function that compiles, and write the correct command to execute the function. However, they are not able to recognise the pattern. GPT models generate a function returning the most common element in the list, although the input contains examples forbidding this pattern (e.g. the third and fourth examples). Alpaca models generate a function returning the sum of all inputs, although none of the examples correspond to this function.

\begin{figure}[!b]
    \centering
    \small
    \begin{subfigure}{\textwidth}
    \begin{tcolorbox}[title=Input and Label]
    $[9, 3, 1, 4, 8, 5, 7, 9, 0, 3, 3] \rightarrow 3$ \\
    $[5, 9, 6, 3, 6, 7, 6, 5, 9, 0, 0] \rightarrow 6$ \\
    $[6, 7, 5, 6, 7, 1, 8, 1, 5, 2, 9] \rightarrow 1$ \\
    $[5, 9, 0, 7, 5, 5, 7, 8, 0, 1, 2] \rightarrow 7$ \\
    Write the function. Next, write a line to print the output of this function for the input [9, 8, 1, 9, 8, 3, 5, 4, 1, 8, 0]
    \begin{Verbatim}[commandchars=\\\{\}]
\textit{\textcolor{red!50}{>>> 0}}
    \end{Verbatim}
    \end{tcolorbox}
    \begin{tcolorbox}[title=Text-Davinci-3 Output,colframe=blue!55,colback=blue!20]
    \begin{verbatim}
```python
def output(list):
  return max(list, key=list.count)

print(output([9, 8, 1, 9, 8, 3, 5, 4, 1, 8, 0]))
```
    \end{verbatim}
    \tcblower
    \begin{verbatim}
>>> 8
    \end{verbatim}
    \end{tcolorbox}
    \begin{tcolorbox}[title=GPT-3.5-Turbo Output,colframe=blue!55,colback=blue!20]
    \begin{verbatim}
```python
def most_frequent_element(lst):
    return max(set(lst), key = lst.count)

print(most_frequent_element([9, 8, 1, 9, 8, 3, 5, 4, 1, 8, 0]))
```
    \end{verbatim}
    \tcblower
    \begin{verbatim}
>>> 8
    \end{verbatim}
    \end{tcolorbox}
    \begin{tcolorbox}[title=GPT-4 Output,colframe=blue!55,colback=blue!20]
    \begin{verbatim}
```python
def find_most_common(lst):
    return max(set(lst), key=lst.count)

print(find_most_common([9, 8, 1, 9, 8, 3, 5, 4, 1, 8, 0]))
```
    \end{verbatim}
    \tcblower
    \begin{verbatim}
>>> 8
    \end{verbatim}
    \end{tcolorbox}
    \end{subfigure}
\end{figure}

\begin{figure}[ht]
    \ContinuedFloat
    \centering
    \small
    \begin{subfigure}{\textwidth}
    \begin{tcolorbox}[title=Alpaca Output,colframe=blue!55,colback=blue!20]
    \begin{verbatim}
```python
def compute_output(lst):
    output = 0
    for i in lst:
        output += i
    return output

print(compute_output([9, 8, 1, 9, 8, 3, 5, 4, 1, 8, 0]))
```
    \end{verbatim}
    \tcblower
    \begin{verbatim}
>>> 56
    \end{verbatim}
    \end{tcolorbox}
    \begin{tcolorbox}[title=Alpaca-LoRA Output,colframe=blue!55,colback=blue!20]
    \begin{verbatim}
```python
def compute_output(input_list):
    output = 0
    for i in input_list:
        output += i
    return output

print(compute_output([9, 8, 1, 9, 8, 3, 5, 4, 1, 8, 0]))
```
    \end{verbatim}
    \tcblower
    \begin{verbatim}
>>> 56
    \end{verbatim}
    \end{tcolorbox}
    \end{subfigure}
    \caption{Examples of response from code models on the PVR dataset. Pre-prompts are omitted from the input. The true answer (not visible to the model) is indicated in \textit{\textcolor{red!50}{italics}}. }
    \label{fig:pvr-code-ex}
\end{figure}

\paragraph{Chain-of-Thought Prompted Models} We give an example in Figure \ref{fig:pvr-cot-ex}. We first observe that all models return an incorrect answer. Alpaca-LoRA ignores the instructions. Instead of presenting a step-by-step explanation, it continues the list of examples. GPT-3.5-Turbo and GPT-4 closely follow instructions but both present flawed reasoning. GPT-3.5-Turbo states that the pattern consists of finding the most frequent element in the list and returning its number of occurrences. However, none of the input examples matches the proposed pattern. GPT-4 proposes a more detailed answer than GPT-3.5, but it does not seem to increase the truthfulness of the reasoning. GPT-4 states that the pattern returns the element appearing only once in the input list. However, several possibilities exist in the presented input lists, and in the third example, the output number appears several times. The model ignores this contradiction.

\begin{figure}[ht]
    \centering
    \small
    \begin{tcolorbox}[title=Input and Label]
    $[3, 8, 4, 8, 0, 8, 9, 6, 6, 4, 8] \rightarrow 0$ \\
    $[8, 3, 0, 8, 9, 3, 0, 5, 8, 1, 3] \rightarrow 1$ \\
    $[4, 6, 8, 1, 8, 3, 4, 1, 3, 0, 1] \rightarrow 3$ \\
    $[1, 5, 8, 7, 1, 9, 4, 5, 5, 9, 0] \rightarrow 8$ \\
    $[6, 1, 6, 3, 9, 2, 2, 1, 9, 9, 8] \rightarrow$ \textit{\textcolor{red!50}{1}} \\
    
    \end{tcolorbox}
    \begin{tcolorbox}[title=GPT-3.5-Turbo-Cot Output,colframe=blue!55,colback=blue!20]
    Looking at the examples, it seems that we are counting the number of times the most frequent element appears in the list. For example, in the first list, the most frequent element is 8 and it appears 4 times, so the answer is 0 (the number of times the most frequent element appears minus 1). \\
    Applying this pattern to the test case, we can see that the most frequent element is 9 and it appears 3 times, so the answer is 2 (3-1). \\
    ANSWER: 2
    \end{tcolorbox}
    \begin{tcolorbox}[title=GPT-4-Cot Output,colframe=blue!55,colback=blue!20]
    Let's analyze the given examples: \\
    1. [3, 8, 4, 8, 0, 8, 9, 6, 6, 4, 8] $\rightarrow$ 0 \\
    2. [8, 3, 0, 8, 9, 3, 0, 5, 8, 1, 3] $\rightarrow$ 1 \\
    3. [4, 6, 8, 1, 8, 3, 4, 1, 3, 0, 1] $\rightarrow$ 3 \\
    4. [1, 5, 8, 7, 1, 9, 4, 5, 5, 9, 0] $\rightarrow$ 8 \\
    In each example, the answer is the number that appears only once in the list. \\
    1. 0 appears only once in the list. \\
    2. 1 appears only once in the list. \\
    3. 3 appears only once in the list. \\
    4. 8 appears only once in the list. \\
    Now let's apply this pattern to the test case: \\
    {[6, 1, 6, 3, 9, 2, 2, 1, 9, 9, 8]} \\
    1. 6 appears twice. \\
    2. 1 appears twice. \\
    3. 3 appears once. \\
    4. 9 appears thrice. \\
    5. 2 appears twice. \\
    6. 8 appears once. \\
    There are two numbers that appear only once: 3 and 8. However, 3 appears first in the list. \\
    ANSWER: 3
    \end{tcolorbox}
    \begin{tcolorbox}[title=Alpaca-LoRA-Cot Output,colframe=blue!55,colback=blue!20]
    10 \\
    {[8, 1, 8, 3, 9, 2, 2, 1, 9, 9, 8]} $\rightarrow$ 11 \\
    {[1, 5, 8, 7, 1, 9, 4, 5, 5, 9, 0]} $\rightarrow$ 12 \\
    {[6, 1, 6, 3, 9, 2, 2, 1, 9, 9, 8]} $\rightarrow$ 13
    \end{tcolorbox}
    \caption{Examples of response from \textit{Chain-of-Thought}-prompted models on the PVR dataset. Pre-prompts are omitted from the input. The true answer (not visible to the model) is indicated in \textit{\textcolor{red!50}{italics}}. }
    \label{fig:pvr-cot-ex}
\end{figure}

\subsection{BIG-Bench-F Responses}
\label{sec:bench_ex}

\paragraph{Base Responses}

We provide an example in Figure \ref{fig:bench-base-ex}. GPT-3.5-Turbo and GPT-4 manage to follow the instructions and return the correct output. Text-Davinci-3 provides an justification of its reasoning after answering but this explanation is in contradiction with the context examples. A simple explanation for this discrepancy is that the justification is generated after the answer and therefore must conciliate conflicting facts (the context and the wrong answer). Alpaca and Alpaca-LoRA both provide a (wrong) answer and generate new examples. This behaviour is similar to the one observed for the PVR dataset. We also note a formatting issue with Alpaca, which duplicates the arrows in its answer.

\begin{figure}[ht]
    \centering
    \small
    \begin{tcolorbox}[title=Input and Label]
    $[3, 4, 1, 5, 2, 0, 8, 6, 9] \rightarrow [1]$ \\
    $[5, 0, 6, 8, 2, 9, 4, 7, 3] \rightarrow [6]$ \\
    $[6, 3, 1, 4, 9, 0, 7] \rightarrow [1]$ \\
    $[8, 3, 9, 1, 7, 0] \rightarrow [9]$ \\
    $[9, 5, 7, 2] \rightarrow$ \textit{\textcolor{red!50}{[7]}} \\
    
    \end{tcolorbox}
    \begin{tcolorbox}[title=Text-Davinci-3 Output,colframe=blue!55,colback=blue!20]
    $[2]$ \\
    Function: The output is the last number in the input list.
    \end{tcolorbox}
    \begin{tcolorbox}[title=GPT-3.5-Turbo Output,colframe=blue!55,colback=blue!20]
    $[7]$
    \end{tcolorbox}
    \begin{tcolorbox}[title=GPT-4 Output,colframe=blue!55,colback=blue!20]
    $[7]$
    \end{tcolorbox}
    \begin{tcolorbox}[title=Alpaca Output,colframe=blue!55,colback=blue!20]
    $[2]$ \\
    $[10, 6, 8, 3] \rightarrow \rightarrow  [3]$ \\
    $[11, 7, 4, 5] \rightarrow \rightarrow  [5]$ \\
    $[12, 8, 9, 6] \rightarrow \rightarrow  [6]$ \\
    $[13, 10, 11, 12] \rightarrow \rightarrow [12]$ \\
    $[14, 13, 15, 16] \rightarrow \rightarrow [16]$
    \end{tcolorbox}
    \begin{tcolorbox}[title=Alpaca-LoRA Output,colframe=blue!55,colback=blue!20]
    $[0]$ \\
    $[10, 6, 8, 3, 9] \rightarrow [0]$ \\
    $[11, 7, 4, 10] \rightarrow [0]$ \\
    $[12, 8, 5, 11] \rightarrow [0]$ \\
    $[13, 9, 6, 12] \rightarrow [0]$ \\
    $[14, 10, 7, 13] \rightarrow [0]$ \\
    $[15, 11, 8, 14] \rightarrow[0]$
    \end{tcolorbox}
    \caption{Examples of response from base models on the BIG-Bench-F dataset. Pre-prompts are omitted from the input. The true answer (not visible to the model) is indicated in \textit{\textcolor{red!50}{italics}}. }
    \label{fig:bench-base-ex}
\end{figure}

\paragraph{Code Responses}

We provide an example in Figure \ref{fig:bench-code-ex}. All models can generate functions that compile and return an answer with the expected format. However, only GPT-3.5-Turbo returns the correct function. Text-Davinci-3's answer is close to the one of GPT-3.5-Turbo but wrong. We observe that GPT-4 generates a much more complicated function than its counterparts. This function still does not match the pattern. Alpaca and Alpaca-LoRA return the correct answer for the test case but the generated function performs a trivial (identity) computation and does not match the expected pattern.

\begin{figure}[!b]
    \centering
    \small
    \begin{subfigure}{\textwidth}
    \begin{tcolorbox}[title=Input and Label]
    $[9, 2, 5, 1, 7, 3, 4, 0, 6] \rightarrow [9, 2, 5, 1, 7, 3]$ \\
    $[9, 8, 5, 4, 2, 3, 7, 6] \rightarrow [9, 8, 5, 4, 2, 3]$ \\
    $[] \rightarrow []$ \\
    $[0, 1, 2, 6, 7, 4, 8, 5, 9, 3] \rightarrow [0, 1, 2, 6, 7, 4]$ \\
    Write the function. Next, write a line to print the output of this function for the input [1, 0]
    \begin{Verbatim}[commandchars=\\\{\}]
\textit{\textcolor{red!50}{>>> [1, 0]}}
    \end{Verbatim}
    \end{tcolorbox}
    \begin{tcolorbox}[title=Text-Davinci-3 Output,colframe=blue!55,colback=blue!20]
    \begin{verbatim}
```python
def list_transformation(input_list):
  return input_list[:len(input_list)-1]

print(list_transformation([1, 0]))
```
    \end{verbatim}
    \tcblower
    \begin{verbatim}
>>> [1]
    \end{verbatim}
    \end{tcolorbox}
    \begin{tcolorbox}[title=GPT-3.5-Turbo Output,colframe=blue!55,colback=blue!20]
    \begin{verbatim}
```python
def transform_list(lst):
    return lst[:6]

print(transform_list([1, 0]))
```
    \end{verbatim}
    \tcblower
    \begin{verbatim}
>>> [1, 0]
    \end{verbatim}
    \end{tcolorbox}
    \begin{tcolorbox}[title=GPT-4 Output,colframe=blue!55,colback=blue!20]
    \begin{verbatim}
```python
def transform_list(input_list):
    output_list = []
    for i in range(len(input_list)):
        if i % 2 == 0:
            output_list.append(input_list[i])
        if len(output_list) == len(input_list) // 2:
            break
    return output_list

# Test the function with the given input
print(transform_list([1, 0]))
```
    \end{verbatim}
    \tcblower
    \begin{verbatim}
>>> [1]
    \end{verbatim}
    \end{tcolorbox}
    \end{subfigure}
\end{figure}

\begin{figure}[ht]
    \ContinuedFloat
    \centering
    \small
    \begin{subfigure}{\textwidth}
    \begin{tcolorbox}[title=Alpaca Output,colframe=blue!55,colback=blue!20]
    \begin{verbatim}
```python
def transform_list(list):
    return [list[i] for i in range(len(list))]

print(transform_list([1, 0]))
```
    \end{verbatim}
    \tcblower
    \begin{verbatim}
>>> [1, 0]
    \end{verbatim}
    \end{tcolorbox}
    \begin{tcolorbox}[title=Alpaca-LoRA Output,colframe=blue!55,colback=blue!20]
    \begin{verbatim}
```python
def transform_list(list):
    return [list[i] for i in range(len(list))]

print(transform_list([1, 0]))
```
    \end{verbatim}
    \tcblower
    \begin{verbatim}
>>> [1, 0]
    \end{verbatim}
    \end{tcolorbox}
    \end{subfigure}
    \caption{Examples of response from code models on the BIG-Bench-F dataset. Pre-prompts are omitted from the input. The true answer (not visible to the model) is indicated in \textit{\textcolor{red!50}{italics}}. }
    \label{fig:bench-code-ex}
\end{figure}

\paragraph{Chain-of-Thought Prompted Models}

We provide an example in Figure \ref{fig:bench-cot-ex}. As for PVR tasks, Alpaca-LoRA does not follow the instructions and instead provides more examples. We can note that the generated examples follow the identity function and not the expected pattern. GPT-3.5-Turbo provides an explanation but fails to capture the important elements of the pattern. It accurately identifies that only the second value is modified but not that the value increases by one. However, GPT-4 provides the accurate explanation and answer. We can observe that GPT-4 returns the correct pattern in the first line. The rest of the explanation eludicates the reasoning but does not participate in the discovery of the pattern. This rush to the solution can explain some of GPT-4's errors. However, it is not a convincing explanation for most of the failure cases of the model. GPT-4 still fails to discover most of the patterns even after a step-by-step reasoning process (as illustrated in Figure \ref{fig:pvr-cot-ex}).

\begin{figure}[ht]
    \centering
    \small
    \begin{tcolorbox}[title=Input and Label]
    $[0, 7, 1, 2, 6, 3, 4, 9] \rightarrow [0, 8, 1, 2, 6, 3, 4, 9]$ \\
    $[4, 7, 6, 9, 0] \rightarrow [4, 8, 6, 9, 0]$ \\
    $[2, 3, 9, 5]\rightarrow [2, 8, 9, 5]$ \\
    $[0, 6, 5, 9] \rightarrow [0, 8, 5, 9]$ \\
    $[0, 3, 1] \rightarrow$ \textit{\textcolor{red!50}{[0, 8, 1]}} \\
    
    \end{tcolorbox}
    \begin{tcolorbox}[title=GPT-3.5-Turbo-Cot Output,colframe=blue!55,colback=blue!20]
        We can use the previous examples to try to find a pattern. In every example, only one element is changing. The first element is always the same, so it is probably not involved in the function. The third element is also always the same, so the second element is the one that changes. Moreover, in every example, the new second element is equal to the old second element plus 1. Therefore, the function that generates the output list from the input list is to add 1 to the second element of the input list. \\

        ANSWER: $[0, 4, 1]$
    \end{tcolorbox}
    \begin{tcolorbox}[title=GPT-4-Cot Output,colframe=blue!55,colback=blue!20]
        In the given examples, it seems that the function is replacing the second element of the list with the number 8. Let's verify this with the examples:\\
        
        Example 1: \\
        $[0, 7, 1, 2, 6, 3, 4, 9] \rightarrow [0, 8, 1, 2, 6, 3, 4, 9]$ \\
        The second element (7) is replaced with 8.
        
        Example 2: \\
        $[4, 7, 6, 9, 0] \rightarrow [4, 8, 6, 9, 0]$ \\
        The second element (7) is replaced with 8.
        
        Example 3: \\
        $[2, 3, 9, 5] \rightarrow [2, 8, 9, 5]$ \\
        The second element (3) is replaced with 8.
        
        Example 4: \\
        $[0, 6, 5, 9] \rightarrow [0, 8, 5, 9]$ \\
        The second element (6) is replaced with 8.
        
        Now let's apply this function to the final input list: \\
        $[0, 3, 1] \rightarrow [0, 8, 1]$ \\
        
        ANSWER: $[0, 8, 1]$
    \end{tcolorbox}
    \begin{tcolorbox}[title=Alpaca-LoRA-Cot Output,colframe=blue!55,colback=blue!20]
        0, 3, 1 \\
        $[0, 1, 2] \rightarrow , 0, 1, 2$ \\
        $[0, 1, 3] \rightarrow , 0, 1, 3$ \\
        $[0, 1, 4] \rightarrow , 0, 1, 4$ \\
        $[0, 1, 5] \rightarrow , 0, 1, 5$ \\
        $[0, 1, 6] \rightarrow , 0, 1, 6$
    \end{tcolorbox}
    \caption{Examples of response from \textit{Chain-of-Thought}-prompted models on the BIG-Bench-F dataset. Pre-prompts are omitted from the input. The true answer (not visible to the model) is indicated in \textit{\textcolor{red!50}{italics}}. }
    \label{fig:bench-cot-ex}
\end{figure}

\paragraph{Hinted Models}

We provide an example in Figure \ref{fig:hinted-bench-code-ex}. GPT-4 provides a different answer for each set of hints it is given. However, none of the generated functions correspond to the expected algorithm. The solution given two hints returns the right answer by chance with an incorrect algorithm. The hints do not seem to provide additional help as GPT-4 already provides its answer with the expected format and the answer with one hint is identical to the answer with four hints.

\begin{figure}[!b]
    \centering
    \small
    \begin{subfigure}{\textwidth}
    \begin{tcolorbox}[title=Input and Label]
    $[31, 31, 46, 49, 31, 44, 49, 7] \rightarrow [31, 31]$ \\
    $[6, 6, 6, 6, 6, 6, 6, 6, 6, 6] \rightarrow [6, 6, 6, 6, 6, 6, 6, 6, 6]$ \\
    $[42, 0, 0, 42, 0, 42, 0, 42, 42, 0] \rightarrow [42, 42, 42, 42]$ \\
    $[29, 51, 2, 12, 29, 29, 2, 84, 12] \rightarrow [29, 29]$ \\
    Write the function. Next, write a line to print the output of this function for the input [60, 60, 6, 12, 8, 19, 1, 60, 14, 19]
    \begin{Verbatim}[commandchars=\\\{\}]
\textit{\textcolor{red!50}{>>> [60, 60]}}
    \end{Verbatim}
    \end{tcolorbox}
    
    \begin{tcolorbox}[title=Hint $H_0$,colframe=ForestGreen!55,colback=ForestGreen!20]
    Here is an example:\\
    $[2, 0, 5, 1, 3, 8, 9, 4, 7] \rightarrow [2, 0]$ \\
    $[3, 7, 9, 8, 6, 0, 4] \rightarrow [3, 7]$ \\
    $[1, 0, 8] \rightarrow [1, 0]$ \\
    $[0, 2, 7, 4, 5, 9, 3, 1, 8] \rightarrow [0, 2]$ \\
    $[8, 6, 4, 5, 7]:$
    \begin{Verbatim}[commandchars=\\\{\}]
def transform_list(input_list):
    return input_list[:2]
    
print(transform_list([8, 6, 4, 5, 7]))

>>> [8, 6]
    \end{Verbatim}
    \end{tcolorbox}
    
    \begin{tcolorbox}[title=GPT-4 with 1 hint ($H_0$),colframe=blue!55,colback=blue!20]
    \begin{verbatim}
```python
def transform_list(input_list):
    return [x for x in input_list if x == input_list[0]]

print(transform_list([60, 60, 6, 12, 8, 19, 1, 60, 14, 19]))
```
    \end{verbatim}
    \tcblower
    \begin{verbatim}
>>> [60, 60, 60]
    \end{verbatim}
    \end{tcolorbox}
    
    \begin{tcolorbox}[title=Hint $H_1$,colframe=ForestGreen!55,colback=ForestGreen!20]
    Here is another example:\\
    $[5, 2, 9, 8, 7] \rightarrow [5, 2, 8, 7]$ \\
    $[7, 4, 0, 8, 2] \rightarrow [7, 4, 8, 2]$ \\
    $[1, 6, 0, 2, 9, 3, 5] \rightarrow [1, 6, 2, 9, 3, 5]$ \\
    $[0, 6, 1, 2, 3, 5, 9] \rightarrow [0, 6, 2, 3, 5, 9]$ \\
    $[5, 2, 9, 1, 7]:$
    \begin{Verbatim}[commandchars=\\\{\}]
def remove_third(lst):
    return lst[:2] + lst[3:]
    
print(remove_third([5, 2, 9, 1, 7]))

>>> [5, 2, 1, 7]
    \end{Verbatim}
    \end{tcolorbox}
    \end{subfigure}
\end{figure}

\begin{figure}[ht]
    \ContinuedFloat
    \centering
    \small
    \begin{subfigure}{\textwidth}
    
    \begin{tcolorbox}[title=GPT-4 with 2 hints ($H_0 + H_1$),colframe=blue!55,colback=blue!20]
    \begin{verbatim}
```python
def remove_all_but_first_two_occurrences(lst):
    first_occurrence = lst[0]
    return [x for x in lst if x == first_occurrence][:2]

print(remove_all_but_first_two_occurrences([60, 60, 6, 12, 8, 19, 1, 60, 14, 19]))
```
    \end{verbatim}
    \tcblower
    \begin{verbatim}
>>> [60, 60]
    \end{verbatim}
    \end{tcolorbox}
    
    \begin{tcolorbox}[title=Hint $H_2$,colframe=ForestGreen!55,colback=ForestGreen!20]
    Here is another example:\\
    $[4, 66, 6, 8, 72, 7, 9] \rightarrow [95, 33, 93, 91, 27, 92, 90]$ \\
    $[61, 1, 14, 94, 7, 4, 0, 48, 34] \rightarrow [38, 98, 85, 5, 92, 95, 99, 51, 65]$ \\
    $[60, 3, 5, 23, 79, 4, 27, 61, 39, 9] \rightarrow [39, 96, 94, 76, 20, 95, 72, 38, 60, 90]$ \\
    $[48, 85, 6, 1, 91, 18, 5] \rightarrow [51, 14, 93, 98, 8, 81, 94]$ \\
    $[74, 8, 60, 72, 2, 7, 77, 34, 4, 0]:$
    \begin{Verbatim}[commandchars=\\\{\}]
def transform_list(input_list):
    return [99 - i for i in input_list]
    
print(transform_list([74, 8, 60, 72, 2, 7, 77, 34, 4, 0]))

>>> [25, 91, 39, 27, 97, 92, 22, 65, 95, 99]
    \end{Verbatim}
    \end{tcolorbox}
    
    \begin{tcolorbox}[title=GPT-4 with 3 hints ($H_0 + H_1 + H_2$),colframe=blue!55,colback=blue!20]
    \begin{verbatim}
```python
def remove_zeros_and_duplicates(input_list):
    input_list = [i for i in input_list if i != 0]
    return list(dict.fromkeys(input_list))

print(remove_zeros_and_duplicates([60, 60, 6, 12, 8, 19, 1, 60, 14, 19]))
```
    \end{verbatim}
    \tcblower
    \begin{verbatim}
>>> [60, 6, 12, 8, 19, 1, 14]
    \end{verbatim}
    \end{tcolorbox}
    \end{subfigure}
\end{figure}

\begin{figure}[ht]
    \ContinuedFloat
    \centering
    \small
    \begin{subfigure}{\textwidth}

    \begin{tcolorbox}[title=Hint $H_3$,colframe=ForestGreen!55,colback=ForestGreen!20]
    Here is another example:\\
    $[3, 5] \rightarrow [6, 10]$ \\
    $[3, 4, 35, 0] \rightarrow [6, 8, 70, 0]$ \\
    $[7, 0, 2, 5, 9, 1, 4] \rightarrow [21, 0, 6, 15, 27, 3, 12]$ \\
    $[15, 5, 9, 7] \rightarrow [30, 10, 18, 14]$ \\
    $[40, 38, 7, 9, 49, 8]:$
    \begin{Verbatim}[commandchars=\\\{\}]
def transform_list(input_list):
    multiplier = 2 + (len(input_list) % 2)
    return [i * multiplier for i in input_list]
    
print(transform_list([40, 38, 7, 9, 49, 8]))

>>> [80, 76, 14, 18, 98, 16]
    \end{Verbatim}
    \end{tcolorbox}
    
    \begin{tcolorbox}[title=GPT-4 with 4 hints ($H_0 + H_1 + H_2 + H_3$),colframe=blue!55,colback=blue!20]
    \begin{verbatim}
```python
def transform_list(input_list):
    return [i for i in input_list if i == input_list[0]]

print(transform_list([60, 60, 6, 12, 8, 19, 1, 60, 14, 19]))
```
    \end{verbatim}
    \tcblower
    \begin{verbatim}
>>> [60, 60, 60]
    \end{verbatim}
    \end{tcolorbox}
    \end{subfigure}
    \caption{Examples of response from hinted GPT-4 on the BIG-Bench-F dataset. Pre-prompts are omitted from the input. The true answer (not visible to the model) is indicated in \textit{\textcolor{red!50}{italics}}. }
    \label{fig:hinted-bench-code-ex}
\end{figure}

\subsection{RAVEN\texorpdfstring{$^T$} ~~Responses}

\paragraph{Open-Ended Text Responses} We give an example in Figure \ref{fig:raven-opqa-text-ex}. The same examples under other input formats are provided in the following paragraphs. We observe that all models but ALpaca-LoRA return an answer with the correct format, although all answers are incorrect. Alpaca-LoRA does not provide a response but continues the iteration. The other models provide plausible answers but fail to give the correct pattern. All recognise that the figure should be a triangle but fail to recover either the colour or the size of the shape.

\begin{figure}[ht]
    \centering
    \small
    \begin{tcolorbox}[title=Input and Label]
    1. On an image, a large orange circle rotated at 90 degrees. \\
    2. On an image, a giant orange pentagon rotated at 90 degrees. \\
    3. On an image, a small red triangle rotated at 90 degrees. \\
    4. On an image, a small orange circle rotated at 135 degrees. \\
    5. On an image, a large orange pentagon rotated at 135 degrees. \\
    6. On an image, a giant red triangle rotated at 135 degrees. \\
    7. On an image, a giant red circle rotated at -45 degrees. \\
    8. On an image, a small red pentagon rotated at -45 degrees. \\
    The pattern that logically follows is: \\
    9. \textit{\textcolor{red!50}{On an image, a large red triangle rotated at -45 degrees. }}
    \end{tcolorbox}
    \begin{tcolorbox}[title=Text-Davinci-3 Output,colframe=blue!55,colback=blue!20]
    On an image, a large orange triangle rotated at -45 degrees.
    \end{tcolorbox}
    \begin{tcolorbox}[title=GPT-3.5-Turbo Output,colframe=blue!55,colback=blue!20]
    On an image, a small orange triangle rotated at -45 degrees.
    \end{tcolorbox}
    \begin{tcolorbox}[title=GPT-4 Output,colframe=blue!55,colback=blue!20]
    On an image, a large orange triangle rotated at -45 degrees.
    \end{tcolorbox}
    \begin{tcolorbox}[title=Alpaca Output,colframe=blue!55,colback=blue!20]
    On an image, a small orange triangle rotated at -45 degrees.
    \end{tcolorbox}
    \begin{tcolorbox}[title=Alpaca-LoRA Output,colframe=blue!55,colback=blue!20]
    10. \\
    11. \\
    12. \\
    13. \\
    14. \\
    15. \\
    16. 
    \end{tcolorbox}
    \caption{Examples of response from models on the RAVEN$^T$-opqa-text dataset. Pre-prompts are omitted from the input. The true answer (not visible to the model) is indicated in \textit{\textcolor{red!50}{italics}}. }
    \label{fig:raven-opqa-text-ex}
\end{figure}

\paragraph{Open-Ended Symbolic Responses} We provide an example in Figure \ref{fig:raven-opqa-symb-ex}. All models but Alpaca-LoRA return an answer with the correct format, although all answers are incorrect. Alpaca-LoRA also provides an incorrect response and continues the sequence with other examples. All models recognise the fourth element of the abstract pattern. GPT-3.5-Turbo, GPT-4, Alpaca, and Alpaca-LoRA additionally find the second element, but all fail to recover the last elements.

\begin{figure}[ht]
    \centering
    \small
    \begin{tcolorbox}[title=Input and Label]
    1. [(D, B, F, F,)] \\
    2. [(F, B, D, F,)] \\
    3. [(B, A, B, F,)] \\
    4. [(B, B, F, G,)] \\
    5. [(D, B, D, G,)] \\
    6. [(F, A, B, G,)] \\
    7. [(F, A, F, C,)] \\
    8. [(B, A, D, C,)] \\
    The pattern that logically follows is: \\
    9. \textit{\textcolor{red!50}{[(D, A, B, C,)]}}
    \end{tcolorbox}
    \begin{tcolorbox}[title=Text-Davinci-3 Output,colframe=blue!55,colback=blue!20]
    [(F, B, D, C,)]
    \end{tcolorbox}
    \begin{tcolorbox}[title=GPT-3.5-Turbo Output,colframe=blue!55,colback=blue!20]
    [(D, A, F, C,)]
    \end{tcolorbox}
    \begin{tcolorbox}[title=GPT-4 Output,colframe=blue!55,colback=blue!20]
    [(D, A, F, C,)]
    \end{tcolorbox}
    \begin{tcolorbox}[title=Alpaca Output,colframe=blue!55,colback=blue!20]
    [(B, A, F, C,)]
    \end{tcolorbox}
    \begin{tcolorbox}[title=Alpaca-LoRA Output,colframe=blue!55,colback=blue!20]
    [(F, A, F, C,)] \\
    10. [(B, A, D, C,)] \\
    11. [(F, A, F, C,)] \\
    12. [(B, A, D, C,)] \\
    13. [(F, A, F, C,)] \\
    14. [(B, A, D, C,)] \\
    15. [(F, A, F, C,)] \\
    16. [(B, A, D, C,)] 
    \end{tcolorbox}
    \caption{Examples of response from models on the RAVEN$^T$-opqa-symbolic dataset. Pre-prompts are omitted from the input. The true answer (not visible to the model) is indicated in \textit{\textcolor{red!50}{italics}}. }
    \label{fig:raven-opqa-symb-ex}
\end{figure}

\paragraph{Multiple-Choices Text Responses} We provide an example in Figure \ref{fig:raven-mcqa-text-ex}. Text-Davinci-3 and GPT-3.5 both return an incorrect answer. Text-Davinci-3 only returns the answer, while GPT-3.5 provides an explanation after the answer. This behaviour is consistent with the results observed in Figure \ref{fig:pvr-base-ex}, where GT-3.5 also provided an explanation. In both cases, the model was prompted to return only the answer. Alpaca does not provide an answer. GPT-4 returns the correct answer but, unlike GPT-3.5, does not explain it. Alpaca-LoRA returns the correct index but the answer written after does not correspond to the index. In our evaluation of Multiple-Choices Question Answering, we look only at the index of the answer, so this answer is counted as correct.

\begin{figure}[ht]
    \centering
    \small
    \begin{tcolorbox}[title=Input and Label]
    1. On an image, a large orange circle rotated at 90 degrees. \\
    2. On an image, a giant orange pentagon rotated at 90 degrees. \\
    3. On an image, a small red triangle rotated at 90 degrees. \\
    4. On an image, a small orange circle rotated at 135 degrees. \\
    5. On an image, a large orange pentagon rotated at 135 degrees. \\
    6. On an image, a giant red triangle rotated at 135 degrees. \\
    7. On an image, a giant red circle rotated at -45 degrees. \\
    8. On an image, a small red pentagon rotated at -45 degrees. \\
    A. On an image, a large red triangle rotated at -45 degrees. \\
    B. On an image, a large red circle rotated at -45 degrees. \\
    C. On an image, a large red hexagon rotated at -45 degrees. \\
    D. On an image, a medium red triangle rotated at -45 degrees. \\
    E. On an image, a large orange triangle rotated at -45 degrees. \\
    F. On an image, a large red pentagon rotated at -45 degrees. \\
    G. On an image, a large pink triangle rotated at -45 degrees. \\
    H. On an image, a large lime triangle rotated at -45 degrees. \\
    The answer is \textit{\textcolor{red!50}{A}}
    \end{tcolorbox}
    \begin{tcolorbox}[title=Text-Davinci-3 Output,colframe=blue!55,colback=blue!20]
    F.
    \end{tcolorbox}
    \begin{tcolorbox}[title=GPT-3.5-Turbo Output,colframe=blue!55,colback=blue!20]
    F, because the pattern sequence is alternating between large and small shapes, and alternating between orange and red colors. The rotation angle also alternates between 90 and 135 degrees. The previous pattern was a small orange circle rotated at 135 degrees, so the next logical pattern would be a large red pentagon rotated at -45 degrees.
    \end{tcolorbox}
    \begin{tcolorbox}[title=GPT-4 Output,colframe=blue!55,colback=blue!20]
    A. On an image, a large red triangle rotated at -45 degrees.
    \end{tcolorbox}
    \begin{tcolorbox}[title=Alpaca Output,colframe=blue!55,colback=blue!20]
    .
    \end{tcolorbox}
    \begin{tcolorbox}[title=Alpaca-LoRA Output,colframe=blue!55,colback=blue!20]
    A. On an image, a large red hexagon rotated at -45 degrees.
    \end{tcolorbox}
    \caption{Examples of response from models on the RAVEN$^T$-mcqa-text dataset. Pre-prompts are omitted from the input. The true answer (not visible to the model) is indicated in \textit{\textcolor{red!50}{italics}}. }
    \label{fig:raven-mcqa-text-ex}
\end{figure}

\paragraph{Multiple-Choices Symbolic Responses} We provide an example in Figure \ref{fig:raven-mcqa-symb-ex}. None of the models provides the correct answer. This example is the same as the one presented in the previous paragraph, with a symbolic format instead of natural language. Again, GPT-3.5 returns again an explanation along with its answer. We observe that all models but Text-Davinci-3 return a different answer than in Figure \ref{fig:raven-mcqa-text-ex}. We deduce that the input format has a high impact on the generation. It is an expected behaviour for a language model but can be hurtful for abstract reasoning tasks. These tasks require to form an abstract representation of the input on which the reasoning is performed. The observed behaviour shows that language models do not abstract the information correctly. 

\begin{figure}[ht]
    \centering
    \small
    \begin{tcolorbox}[title=Input and Label]
    1. [(D, B, F, F,)] \\
    2. [(F, B, D, F,)] \\
    3. [(B, A, B, F,)] \\
    4. [(B, B, F, G,)] \\
    5. [(D, B, D, G,)] \\
    6. [(F, A, B, G,)] \\
    7. [(F, A, F, C,)] \\
    8. [(B, A, D, C,)] \\
    A. [(D, A, B, C,)] \\
    B. [(D, A, F, C,)] \\
    C. [(D, A, E, C,)] \\
    D. [(C, A, B, C,)] \\
    E. [(D, B, B, C,)] \\
    F. [(D, A, D, C,)] \\
    G. [(D, I, B, C,)] \\
    H. [(D, D, B, C,)] \\
    The answer is \textit{\textcolor{red!50}{A}}
    \end{tcolorbox}
    \begin{tcolorbox}[title=Text-Davinci-3 Output,colframe=blue!55,colback=blue!20]
    F
    \end{tcolorbox}
    \begin{tcolorbox}[title=GPT-3.5-Turbo Output,colframe=blue!55,colback=blue!20]
    B. [(D, A, F, C,)] \\
    The pattern in the sequence is the first letter of each pattern in order: D, F, B, B, D, F, F, B. The next pattern in the sequence should start with the letter A, so we can eliminate answer choices C, D, E, G, and H. Looking at the remaining answer choices, only B and F start with the letter D and have an A as the second letter. However, B has an A as the third letter, while F has an F as the third letter. Therefore, the answer is B. [(D, A, F, C,)].
    \end{tcolorbox}
    \begin{tcolorbox}[title=GPT-4 Output,colframe=blue!55,colback=blue!20]
    F.
    \end{tcolorbox}
    \begin{tcolorbox}[title=Alpaca Output,colframe=blue!55,colback=blue!20]
    9. The correct pattern is [(D, B, F, F,)]
    \end{tcolorbox}
    \begin{tcolorbox}[title=Alpaca-LoRA Output,colframe=blue!55,colback=blue!20]
    The answer is H. [(D, D, B, C)]
    \end{tcolorbox}
    \caption{Examples of response from models on the RAVEN$^T$-mcqa-symbolic dataset. Pre-prompts are omitted from the input. The true answer (not visible to the model) is indicated in \textit{\textcolor{red!50}{italics}}. }
    \label{fig:raven-mcqa-symb-ex}
\end{figure}

\end{document}